\documentclass{article}

\PassOptionsToPackage{numbers, compress, sort}{natbib}

\usepackage[font={footnotesize}]{caption}

    \usepackage[final]{neurips_2023}

\usepackage[utf8]{inputenc} 
\usepackage[T1]{fontenc}    
\usepackage{url}            
\usepackage{booktabs}       
\usepackage{amsfonts}       
\usepackage{nicefrac}       
\usepackage{microtype}      
\usepackage[dvipsnames]{xcolor}
\usepackage{pifont}
\usepackage[most]{tcolorbox}
\usepackage{adjustbox}
\usepackage{wrapfig}
\usepackage{multirow,mathtools } 
\usepackage{algorithm,algpseudocode}
\usepackage{rotating}
\usepackage{lscape}
\usepackage{longtable}
\usepackage{subcaption}
\usepackage[color=gray!20,textsize=scriptsize]{todonotes}
\usepackage[pagebackref,breaklinks,colorlinks]{hyperref}

\usepackage{algorithm,algpseudocode}
\algnewcommand{\algorithmicforeach}{\textbf{for each}}
\algdef{SE}[FOR]{ForEach}{EndForEach}[1]
  {\algorithmicforeach\ #1\ \algorithmicdo}
  {\algorithmicend\ \algorithmicforeach}

\usepackage{pifont}

\usepackage{color, colortbl}
\definecolor{Gray}{gray}{0.93}
\definecolor{Orange}{rgb}{1,0.5,0}
\definecolor{DGray}{gray}{0.83}
\definecolor{LightCyan}{rgb}{0.88,1,1}

\definecolor{natural}{rgb}{0.7137,0.3333,0.3333}
\definecolor{specialized}{rgb}{0.4118,0.6431,0.4314}
\definecolor{structured}{rgb}{0.3254,0.4431,0.6666}
\definecolor{all}{rgb}{0.7529,0.4902,0.6471}
\usepackage{wrapfig}

\usepackage{multirow,mathtools } \usepackage{algorithm,algpseudocode}

\usepackage{pifont}
\usepackage{color, colortbl}

\usepackage{blindtext}
\usepackage{lipsum}

\usepackage{multirow}
\usepackage{graphicx}
\usepackage{listings}

\usepackage{bbm}

\usepackage[most]{tcolorbox}

\usepackage{rotating}
\usepackage{lscape}
\usepackage{longtable}
\usepackage{subcaption}

\usepackage{threeparttable}
\usepackage{tabularx}
\usepackage{colortbl}
\usepackage{float}
\usepackage{tikz}
\usepackage{sidecap}
\sidecaptionvpos{figure}{t}

\usepackage{amsmath,amsfonts,bm}

\def\eqref#1{(\ref{#1})}

\def\1{\bm{1}}

\DeclareMathAlphabet{\mathsfit}{\encodingdefault}{\sfdefault}{m}{sl}
\SetMathAlphabet{\mathsfit}{bold}{\encodingdefault}{\sfdefault}{bx}{n}

\def\gC{{\mathcal{C}}}
\def\gD{{\mathcal{D}}}

\def\gS{{\mathcal{S}}}
\def\gT{{\mathcal{T}}}

\def\gX{{\mathcal{X}}}

\def\gZ{{\mathcal{Z}}}

\DeclareMathOperator*{\argmin}{arg\,min}

\newcommand{\bx}{\mathbf{x}}

\newcommand{\bz}{\mathbf{z}}

\newcommand{\random}{\textsc{Random}}
\newcommand{\gradnorm}{\textsc{GraNd}}
\newcommand{\eln}{\textsc{EL2N}}

\newcommand{\forget}{\textsc{Forget}}
\newcommand{\infmax}{\textsc{InfMax}}
\newcommand{\infsum}{\textsc{InfSum}}
\newcommand{\moderate}{\textsc{Moderate}}
\newcommand{\spr}{\textsc{SP}}
\newcommand{\ssp}{\textsc{SSP}}

\newcommand{\revision}[1]{{#1}}

\title{Selectivity Drives Productivity: Efficient Dataset Pruning for Enhanced Transfer Learning}

\author{%
  Yihua Zhang\textsuperscript{1,*} 
  \And 
  Yimeng Zhang\textsuperscript{1,*} 
  \And
  Aochuan Chen\textsuperscript{1,*} 
  \And
  Jinghan Jia\textsuperscript{1}
  \And
  Jiancheng Liu\textsuperscript{1} \\
  \And
  Gaowen Liu\textsuperscript{2} 
  \And
  Mingyi Hong\textsuperscript{3} 
  \And
  Shiyu Chang\textsuperscript{4} 
  \And
  Sijia Liu\textsuperscript{1}
  \And \vspace*{-6mm}\\
  \textsuperscript{1} Michigan State University, \textsuperscript{2} Cisco Research, \\
  \textsuperscript{3} University of Minnesota, Twin City, \textsuperscript{4} UC Santa Barbara \\
  \textsuperscript{*} Equal Contribution
}

\begin{document}

\maketitle

\begin{abstract}
Massive data is often considered essential for  deep learning applications, but it also incurs significant computational and infrastructural costs.  Therefore, dataset pruning (DP) has emerged as an effective way to improve data efficiency by identifying and removing redundant training samples without sacrificing performance. In this work, we aim to address the problem of DP for transfer learning, \textit{i.e.}, \textit{how to prune a source dataset for improved pretraining efficiency and lossless finetuning accuracy on downstream target tasks}. To our best knowledge,  the problem of DP for transfer learning remains open, as previous studies have primarily addressed DP and transfer learning as separate problems. 
By contrast, we establish a unified viewpoint to integrate DP with transfer learning and find that existing DP methods are not suitable for the transfer learning paradigm. We then propose two new DP methods, label mapping and feature mapping, for supervised and self-supervised pretraining settings respectively, by revisiting the DP problem through the lens of source-target domain mapping. Furthermore, we demonstrate the effectiveness of our approach on numerous transfer learning tasks. We show that source data classes can be pruned by up to $40\%\sim 80\%$ without sacrificing downstream performance, resulting in a significant $2\sim 5\times$ speed-up during the pretraining stage. Besides, our proposal exhibits broad applicability and can improve other computationally intensive transfer learning techniques, such as adversarial pretraining. Codes are available at \url{https://github.com/OPTML-Group/DP4TL}.

\end{abstract}

\section{Introduction}
\label{sec: intro}
The abundance of the training data has long been regarded as the key driver of the contemporary machine learning (ML) algorithms \cite{zhao2020dataset, zhao2021dataset, moderate2023xia}. 
However, the untrimmed, ever-growing training dataset could not only introduce training biases that compromise the model performance \cite{geirhos2018imagenet, ilyas2019adversarial, gu2017badnets}, but also poses an almost insurmountable obstacle to high training efficiency \cite{zhao2020dataset, zhao2021dataset}.
Therefore, understanding the impact of the training data and selecting the most critical samples has emerged as an important goal, collectively referred to as the problem of \textit{dataset pruning} (\textbf{DP}). 

Although the feasibility and the promise of DP have been unveiled in numerous applications, such as noisy data cleansing \cite{wei2020combating, jiang2018mentornet, guo2022deepcore}, continue learning \cite{borsos2020coresets, toneva2018empirical, castro2018end, aljundi2019gradient} and active learning \cite{sener2017active}, greater emphasis is placed on the \textit{in-domain} training setting, \textit{i.e.}, the pruned training set share the similar distribution as the evaluation set.
Examples of   methods to condense the   training dataset with lossless generalization (on the  in-distribution 
testing dataset) include  
data influence functions \cite{chatterjee1986influential, cook1986assessment, thomas1990assessing, wei1998generalized, koh2017understanding, schioppa2022scaling, guo2020fastif, yang2022dataset, borsos2020coresets, kong2022resolving},
model training dynamics \cite{pleiss2020identifying, welling2009herding, paul2021deep, aljundi2019gradient, castro2018end, toneva2018empirical, pruthi2020estimating, rebuffi2017icarl, sener2017active},
and coreset selection \cite{feldman2011unified, ju2022extending, huggins2016coresets, campbell2019automated, nguyen2017variational, farquhar2018towards, kim2023coreset, yang2023towards}. 
In contrast, our work investigates the problem of {DP} in the \textbf{transfer learning} paradigm \cite{pan2010survey, torrey2010transfer, yang2013theory}, which has emerged as a popular way to leverage the knowledge of a \textit{foundation model} learned on a \textit{source} dataset (referring to `\textit{pretraining}' stage) to further enhance the performance on a cross-domain \textit{target} dataset  (referring to `\textit{finetuning}' stage).
Recent evidence \cite{sorscher2022beyond,jain2022data,sattigeri2022fair} has shown that some source data points could make  a \textit{harmful influence} in the downstream performance.
In particular, the previous study \cite{jain2022data} showed that removing specific source data classes can improve the transfer learning accuracy of a pretrained model.

Yet, \textit{efficiently} identifying those \textit{harmful} source data classes for improved transfer learning is a highly non-trivial task. 
\underline{Firstly}, the unique pretrain-finetune paradigm complicates the analysis of the influence of source data on downstream performance, making it an indirect and very challenging process. Consequently, existing gradient and data influence-based in-domain DP methods \cite{koh2017understanding, schioppa2022scaling, guo2020fastif, yang2022dataset, borsos2020coresets, kong2022resolving} cannot be na\"ively adapted to the transfer setting. 
\underline{Secondly}, transfer learning encompasses a wide range of source training methods other than supervised learning, such as self-supervised learning (\textbf{SSL}) \cite{chen2020simple, he2020momentum, chen2020improved, chen2020big, wu2018unsupervised, tian2020makes, grill2020bootstrap, asano2019self, caron2020unsupervised, li2020prototypical}. Therefore, a generic DP framework is desired given    various pretraining scenarios.
\underline{Thirdly}, from an efficiency standpoint, the design of  efficient DP algorithms is non-trivial.
Even in the non-transfer learning paradigm, a majority of current in-domain DP methods \cite{paul2021deep, toneva2018empirical, welling2009herding} introduce substantial computational overheads. For instance, influence function-based DP methods \cite{chatterjee1986influential, cook1986assessment, thomas1990assessing, wei1998generalized, koh2017understanding, schioppa2022scaling, guo2020fastif, yang2022dataset, borsos2020coresets} necessitate the calculation of the inverse Hessian matrix of model parameters, which is a highly demanding and computationally intensive process. Moreover, training dynamics-based methods \cite{feldman2020neural, pleiss2020identifying, welling2009herding, paul2021deep, pruthi2020estimating, toneva2018empirical, rebuffi2017icarl, castro2018end, sener2017active, aljundi2019gradient}, such as GraNd-Score \cite{paul2021deep} and Forgetting-Score \cite{toneva2018empirical}, necessitate model training on the entire dataset multiple times. 
As a result, the development of an  efficient  and effective DP framework tailored for transfer learning remains a significant challenge in the field.

\begin{figure}
    \centering
    \includegraphics[width=\linewidth]{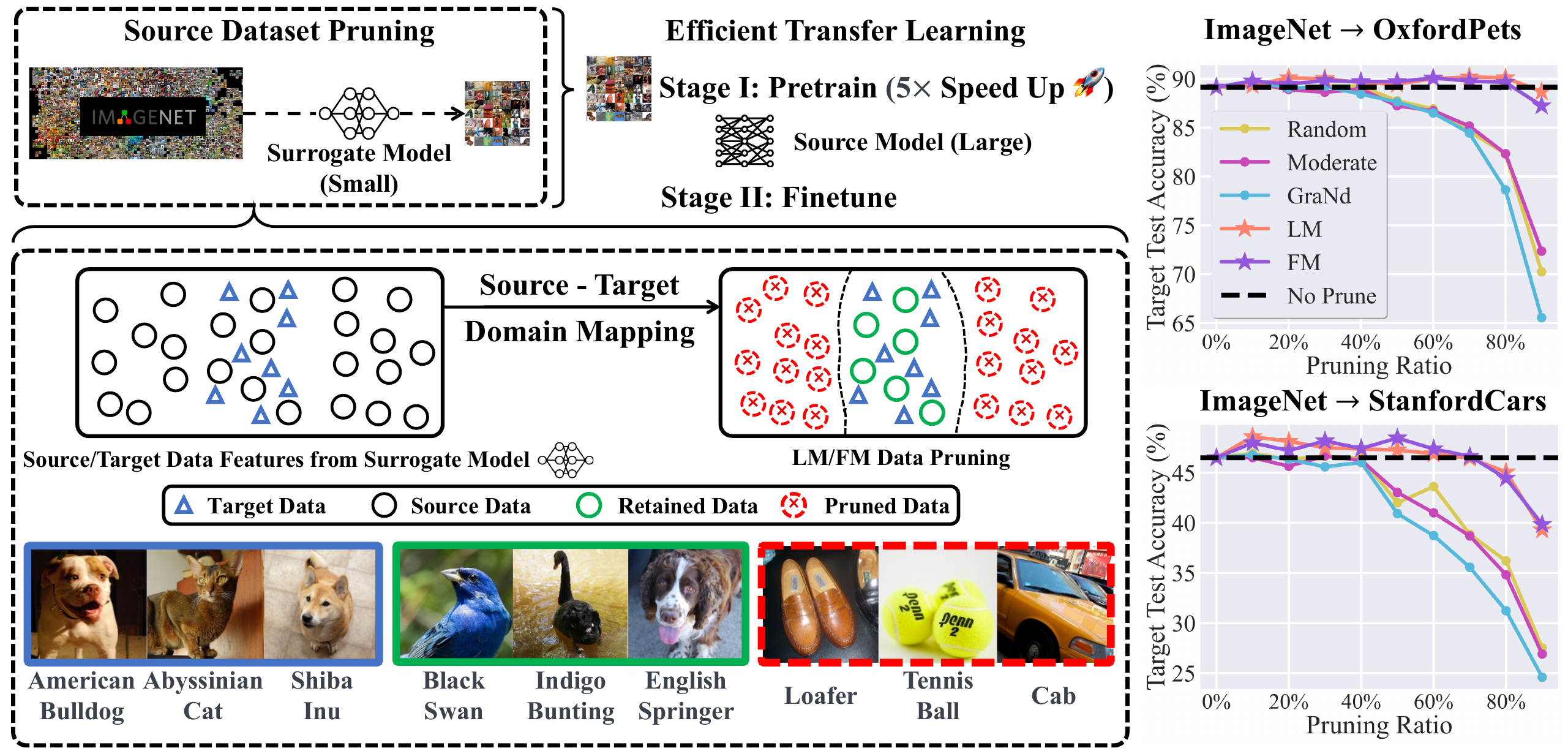}
    \caption{
    \textbf{Left}: An illustration of the proposed dataset pruning  methods (LM and FM) and their performance overview. Large scale source dataset is pruned by LM and FM through a small surrogate model (ResNet-18). Large foundation models can achieve up to $5\times$ speed-up on pretraining without no downstream performance drop. \textbf{Right}: Downstream performance overview at different pruning ratios when ResNet-101 \cite{he2016deep} is used as the source model (trained on the pruned source dataset) with ImageNet \cite{russakovsky2015imagenet} as the source dataset transferring to OxfordPets \cite{parkhi2012cats} and StanfordCars \cite{krause20133d}. The feature extractor is fixed during finetuning.}
    \label{fig: overview}
    \vspace*{-5mm}
\end{figure}

The most relevant work to ours is \cite{jain2022data}, which proposed a brute-force method to evaluate the influence of every source class in the downstream task following a leave-one-out principle. While this method effectively selects the most influential source classes, it is prohibitively slow, as the data selection process demands significantly higher computational resources than performing transfer learning itself. This brings us to the central question addressed in this paper:
\begin{tcolorbox}[before skip=0.2cm, after skip=0.2cm, boxsep=0.0cm, middle=0.1cm, top=0.1cm, bottom=0.1cm]
\begin{center}
    \textit{\textbf{(Q)} How can we extend DP to transfer learning with minimal computation overhead, 
    broad  applicability, and improved target performance?}
\end{center}
\end{tcolorbox}

To address \textbf{(Q)}, we   formally define the task of {DP (dataset pruning) for transfer learning}. We start by uncovering the limitations of conventional in-domain DP methods when applied to transfer learning tasks, and establish an intrinsic connection between 
source dataset pruning for transfer learning and source-target domain mapping.
In the supervised pretraining paradigm, 
we propose an effective and scalable label mapping (\textbf{LM})-based framework capable of pinpointing the most beneficial source data labels for a cross-domain target dataset. Furthermore, we extend the idea of LM to feature mapping (\textbf{FM}) for the SSL pretraining protocol, where data labels are unavailable. 
To achieve a greater practical impact, we demonstrate that our proposed methods (LM and FM) can facilitate effective DP over a small, simple surrogate source model, and further translate the positive impact of the pruned source dataset to other larger, more complex source models.
We provide a schematic overview and result highlights in \textbf{Fig.\,\ref{fig: overview}}.
Our contributions can be summarized as follows:

\ding{182} We connect the concept of DP (dataset pruning) to transfer learning for the first time and 
formalize the problem of DP for transfer learning.

\ding{183} We develop a highly efficient and effective framework for DP in transfer learning, leveraging the source-target domain mapping with customized LM (label mapping) and FM (feature mapping) for both supervised and SSL (self-supervised learning) settings. Our approach is principled and achieves significant speedups compared to the state-of-the-art methods.

\ding{184} We empirically show  the effectiveness of our proposals (LP and FP) on 8 downstream tasks. We find that the source dataset (ImageNet)   can be pruned up to $40\%\sim80\%$  without sacrificing the downstream performance, together with   a $2\times\sim5\times$ speed-up in the pretraining stage.
Our proposal also unlocks a door to prune a dataset using a simple surrogate source model (\textit{e.g.}, ResNet-18) and then reuse the pruned dataset to improve transfer learning on a larger source model (\textit{e.g.}, ResNet-101).

\ding{185} Lastly, we show that our proposed DP framework can benefit other  computationally-intensive transfer learning techniques. For example, DP for adversarial pretraining \cite{salman2020adversarially} leads to a $1\%\sim 2\%$ improvement in downstream performance with time consumption similar to standard pretraining.

\section{Related Work}
\label{sec: related_work}

\paragraph{Dataset pruning.}
DP (dataset pruning) is an emerging technique to improve the data efficiency of model training by selecting the most representative training samples or removing the less influential ones \cite{chatterjee1986influential, cook1986assessment, thomas1990assessing, wei1998generalized, koh2017understanding, schioppa2022scaling, guo2020fastif, yang2022dataset, borsos2020coresets, kong2022resolving, pleiss2020identifying, welling2009herding, paul2021deep, aljundi2019gradient, castro2018end, toneva2018empirical, pruthi2020estimating, rebuffi2017icarl, sener2017active}. 
Thus, the problem of coreset selection \cite{feldman2011unified, ju2022extending, huggins2016coresets, campbell2019automated, nguyen2017variational, farquhar2018towards, kim2023coreset, yang2023towards} can also be considered a form of DP.
Prior studies on DP range from clustering-based methods \cite{agarwal2004approximating, har2004coresets, feldman2020turning} to the more recent score-based pruning methods \cite{chatterjee1986influential, cook1986assessment, thomas1990assessing, wei1998generalized, koh2017understanding, schioppa2022scaling, guo2020fastif, yang2022dataset, borsos2020coresets, kong2022resolving, welling2009herding, paul2021deep, pruthi2020estimating, toneva2018empirical, rebuffi2017icarl, castro2018end, sener2017active, aljundi2019gradient, pleiss2020identifying}. 
In the latter approach, an importance score is assigned to each training data point   to quantify its influence on a particular permanence metric during model learning.
Specifically, score-based DP methods  can be broadly categorized into two main groups: influence function-based approaches  \cite{chatterjee1986influential, cook1986assessment, thomas1990assessing, wei1998generalized, koh2017understanding, schioppa2022scaling, guo2020fastif, yang2022dataset, borsos2020coresets, kong2022resolving} and training dynamics-based approaches \cite{feldman2020neural, pleiss2020identifying, welling2009herding, paul2021deep, pruthi2020estimating, toneva2018empirical, rebuffi2017icarl, castro2018end, sener2017active, aljundi2019gradient}.
The first category measures data influence by   examining the effect of data removal on the learning algorithm used during training \cite{chatterjee1986influential, cook1986assessment, thomas1990assessing, wei1998generalized} and  the model's prediction \cite{koh2017understanding, schioppa2022scaling, guo2020fastif, yang2022dataset, borsos2020coresets, kong2022resolving}. However, influence function-based approaches  typically require high computational costs due to the need for high-order derivatives and complex optimization methods, such as bi-level optimization, as used in  \cite{yang2022dataset}. Although an approximate influence score can be obtained efficiently, it may   result in a large estimation error  \cite{basu2020influence}. 
In the second category,  training dynamics-based approaches  find   statistical indicators for pruned data from the training trajectory. Examples of DP metrics include data loss/error \cite{welling2009herding, paul2021deep, pruthi2020estimating}, prediction confidence \cite{pleiss2020identifying}, model gradient norm \cite{paul2021deep}, forgetting event \cite{toneva2018empirical}, and compactness \cite{rebuffi2017icarl, castro2018end, sener2017active, aljundi2019gradient}. However,  these methods typically require repeated training  to ensure the representativeness of the collected statistics \cite{paul2021deep, toneva2018empirical, welling2009herding}. 
In addition to improving data efficiency, DP has been applied in a variety of contexts, such as   noisy label cleansing \cite{wei2020combating, jiang2018mentornet, guo2022deepcore},  continue learning \cite{borsos2020coresets, toneva2018empirical, castro2018end, aljundi2019gradient,jia2023robustness}, active learning \cite{sener2017active}, and reducing annotation cost of a finetuner \cite{xie2023active}.

\paragraph{Data valuation \& attribution.} 
Data valuation \cite{koh2017understanding, jain2022data, ghorbani2019data, jia2019towards, jia2019efficient, lin2022measuring, guu2023simfluence, hammoudeh2022identifying} and   attribution \cite{park2023trak, just2023lava, ilyas2022datamodels} are research streams related to dataset pruning that aim to quantify the influence of   training data points on model's performance. Unlike dataset pruning, these approaches do not focus primarily on training efficiency but instead aim to enhance  interpretability \cite{yeh2018representer}, adversarial robustness \cite{hammoudeh2022identifying, lin2022measuring}, generative model design \cite{song2019generative}, and data acquisition quality \cite{just2023lava}.
Representative methods include Shapley value-based methods \cite{ghorbani2019data, jia2019towards}, datamodels \cite{ilyas2022datamodels}, sampling-based methods \cite{ghorbani2019data, lin2022measuring}, and proxy-based methods \cite{jia2019efficient, just2023lava}. 
\revision{Recently, \citet{kim2023coreset} proposed to select data samples from a large public dataset (Open-Set) for self-supervised learning given a specific target dataset through distribution mismatch. However, it fails to make a general framework in both supervised and self-supervised scenarios, leaving the former under-explored.}
The most relevant work to ours is \cite{jain2022data}, which leverages a leave-one-out analysis to quantify the influence of source data on downstream tasks in the context of transfer learning.
With this inspiration, we propose to connect DP to transfer learning in this work.

\paragraph{Recent advancements in transfer learning.} Transfer learning has been a prominent area  over the past decade \cite{pan2010survey, torrey2010transfer, yang2013theory}. Significant strides have been made in understanding, analyzing, and improving various aspects of this technique \cite{deng2021adversarial, salman2020adversarially, liu2022same, liu2019towards, rosenstein2005transfer, duan2012exploiting, ge2014handling, yosinski2014transferable, cao2018partial, wang2019characterizing}.
Recent studies \cite{liu2022same, liu2019towards} have rigorously analyzed the relationship between source model performance and downstream task effectiveness, arguing that a narrow focus on minimizing source training loss may not lead to improved transfer learning results.
To improve transfer learning,  adversarial training has been shown to benefit the transferability of   source pretraining on target downstream tasks  \cite{deng2021adversarial, salman2020adversarially}.
There also exist studies  to identify and understand the failure cases in transfer learning \cite{rosenstein2005transfer, duan2012exploiting, ge2014handling, yosinski2014transferable, cao2018partial, wang2019characterizing}.
Other research \cite{liu2019towards, liu2022same} has examined model transferability from the perspective of model sharpness and argues that a good pretrained model should be situated in a flat basin in the downstream loss landscape. Last but not the least, transfer learning has progressed in several other directions, such as self-supervised learning \cite{chen2020simple, he2020momentum, chen2020improved, chen2020big, wu2018unsupervised, tian2020makes, grill2020bootstrap, asano2019self, caron2020unsupervised, li2020prototypical}, model weight pruning \cite{guo2020parameter,liu2021transtailor, chen2021lottery, chen2021chasing, myung2022pac}, and visual prompting \cite{bahng2022visual, elsayed2018adversarial, chen2022model, zheng2021adversarial, neekhara2018adversarial, neekhara2022cross, chen2021adversarial, chen2022understanding, chen2023visual, zhang2022fairness}.

\section{Problem Formulation}
\label{sec: prob}

In this section, we introduce some essential preliminaries on transfer learning and DP (dataset pruning), and  elucidate the design challenge of DP for transfer learning.

\paragraph{Preliminaries on transfer learning and connection to DP.}
Let $g \circ f$ denote a deep neural network that consists of a feature extractor $f $ and a classification head $g $, where $\circ$ denotes the function composition.  Given   \underline{s}ource and \underline{t}arget datasets $\gD_\gS$ and $\gD_\gT$,  we study transfer learning   in   the ``\textit{pretrain-finetune}'' paradigm. 
The primary goal of \textit{pretraining} is to obtain a high-quality feature extractor $f_\gS: \gX \rightarrow \gZ$, which 
draws a mapping from the input space ($\gX$) to the deep representation space ($\gZ$) in a data-rich source domain ($\gD_\gS$). 
Popular pertaining recipes include {supervised learning} (\textbf{SL}) \cite{wang2019characterizing, zhang2021quantifying} and {self-supervised learning} (\textbf{SSL}) \cite{he2020momentum, chen2020simple, chen2020big, chen2020improved}
depending on whether the source labels ($y_\gS$) are available or not in $\gD_\gS$. 
In the \textit{finetuning} stage, the pretrained model is further trained on a specific downstream task under the  target dataset $\gD_\gT$. Transfer learning expects  improved downstream performance over  training  on $\gD_\gT$ from scratch. 
In this work, we consider two finetuning protocols,    linear probe (\textbf{LP}) and full-finetune (\textbf{FF}), with $\gD_\gT$ being a labeled dataset.   
LP finetunes the linear classification head $g$ with a fixed feature extractor $f_\gS$, acquired from pretraining. In contrast, FF finetunes the entire  model $g\circ f$ from the initialization $f_\gS$.   FF typically yields a better transfer learning accuracy than LP, but the former takes a higher computation cost. 
 
Our motivation for connecting transfer learning with DP comes from a recent data-based perspective on transfer learning \cite{jain2022data}. The study shows that removing certain \textit{source data classes} from $\gD_\gS$ can potentially improve the accuracy of a finetuned model on $\gD_\gT$. 
However, the task of evaluating the transfer effects of source data and removing their influence from a pre-trained source model has not been addressed efficiently. The approach developed in \cite{jain2022data} involves a leave-one-out analysis to estimate the influence of a source class $c$ on a target example $t$, which is computed as the prediction discrepancy of the finetuned source model at $t$ when the class $c$ is either included or excluded from $\gD_\gS$. 
During this process,  one must train multiple source models (over $7000$ models on ImageNet in \cite{jain2022data}) from scratch over different subsets of $\gD_\gS$ for a given target task. 
This approach becomes  computationally unaffordable when dealing with large source datasets like ImageNet given limited computing resources.
To address this challenge,  we propose a   DP perspective on transfer learning.

\paragraph{Problem of interest: DP for transfer learning.}
Next, we introduce the background of DP and the problem we focus on. Let $\gD = \{\bz_1, \bz_2, \dots, \bz_N\}$ denote a dataset consisting of $n$ samples, where each sample $z_i$ is represented as a pair $(\bx_i, y_i)$, with $\bx_i$ denoting the input feature vector and $y_i$ denoting the corresponding label.  DP aims to generate a pruned dataset $\hat{\gD} = \{\hat{\bz}_1, \hat{\bz}_2, \dots, \hat{\bz}_M \} \subset \gD$ with $M < N$, which can reduce the training cost without a significant decrease in model generalization performance when trained on $\hat{\gD}$. In the context of \cite{jain2022data}, instead of individual source data sample $\{ \hat {\bz}_i \}$, the entire source classes   are evaluated and selected in terms of transfer influences.
Based on the above, we  define the problem of our interest below. 
\begin{center}
\vspace*{-1.1mm}
	\setlength\fboxrule{0.1pt}
	\noindent\fcolorbox{black}[rgb]{0.85,0.9,0.95}{\begin{minipage}{0.96\columnwidth}
	\textbf{(DP for transfer learning)}	
How to \textit{prune} source data classes to obtain $\hat{\gD}_\gS$ (a subset of ${\gD}_\gS$), with lossless or improved   transfer learning accuracy   of the source model ($f_\gS$) on a   target task $\gD_\gT$?
	\end{minipage}}
	\vspace*{-1.0mm}
\end{center}

DP for transfer learning has two key distinctions (\ding{182}-\ding{183}) from the vanilla DP setup. First (\ding{182}), DP must be performed in the source domain (${\gD}_\gS$), while its effectiveness is evaluated based on the target domain (${\gD}_\gT$). This `cross-domain' challenge makes the design of efficient and effective DP highly non-trivial. For example, prior work \cite{jain2022data} utilizes a computationally-intensive leave-one-out analysis. Classical influence function-based methods \cite{koh2017understanding, schioppa2022scaling, guo2020fastif, yang2022dataset, borsos2020coresets, kong2022resolving, zhang2023introduction}, which trace the eventual model's prediction through the learning algorithm and back to its training data, are also computationally infeasible due to the complex bi-level optimizer and the calculation of high-order derivatives.
Second (\ding{183}), the pre-trained source model ($f_\gS$) in today's transfer learning regime is typically of a large scale. This motivates us to develop a DP method that can be independent of the source model, while the pruned source dataset $\hat{\gD}_\gS$ remains effective in the original transfer learning setup. Given this challenge, we will design DP methods for transfer learning using a \textit{simple surrogate source model} to avoid the computation on the large source model $f_\gS$ (see \textbf{Fig. \ref{fig: overview}} for an illustration).

\begin{wrapfigure}{r}{85mm}
\vspace*{-5mm}
\centerline{
\begin{tabular}{cc}
\hspace*{0mm}\includegraphics[width=.3\textwidth,height=!]{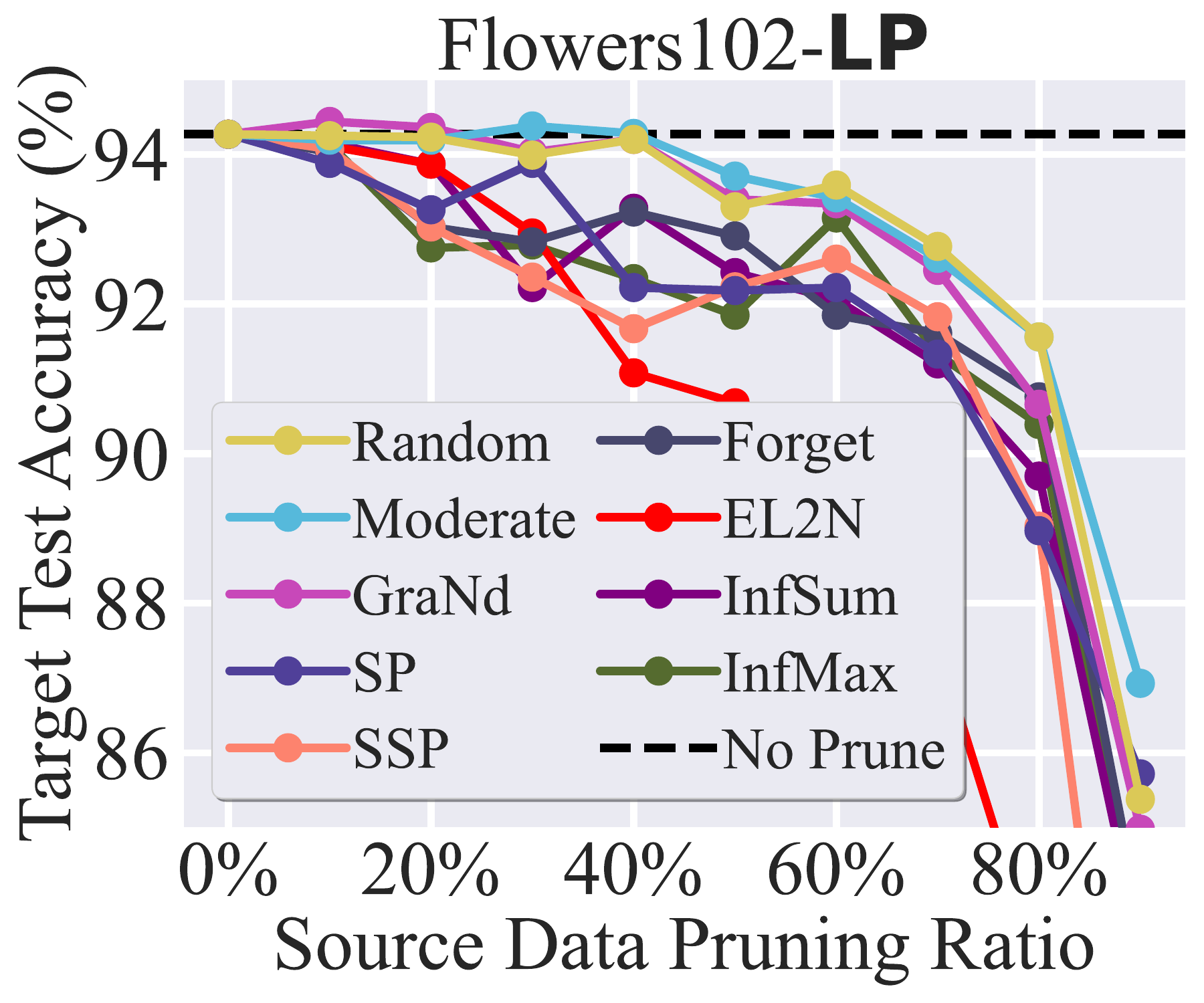}  
&\hspace*{-4mm}\includegraphics[width=.3\textwidth,height=!]{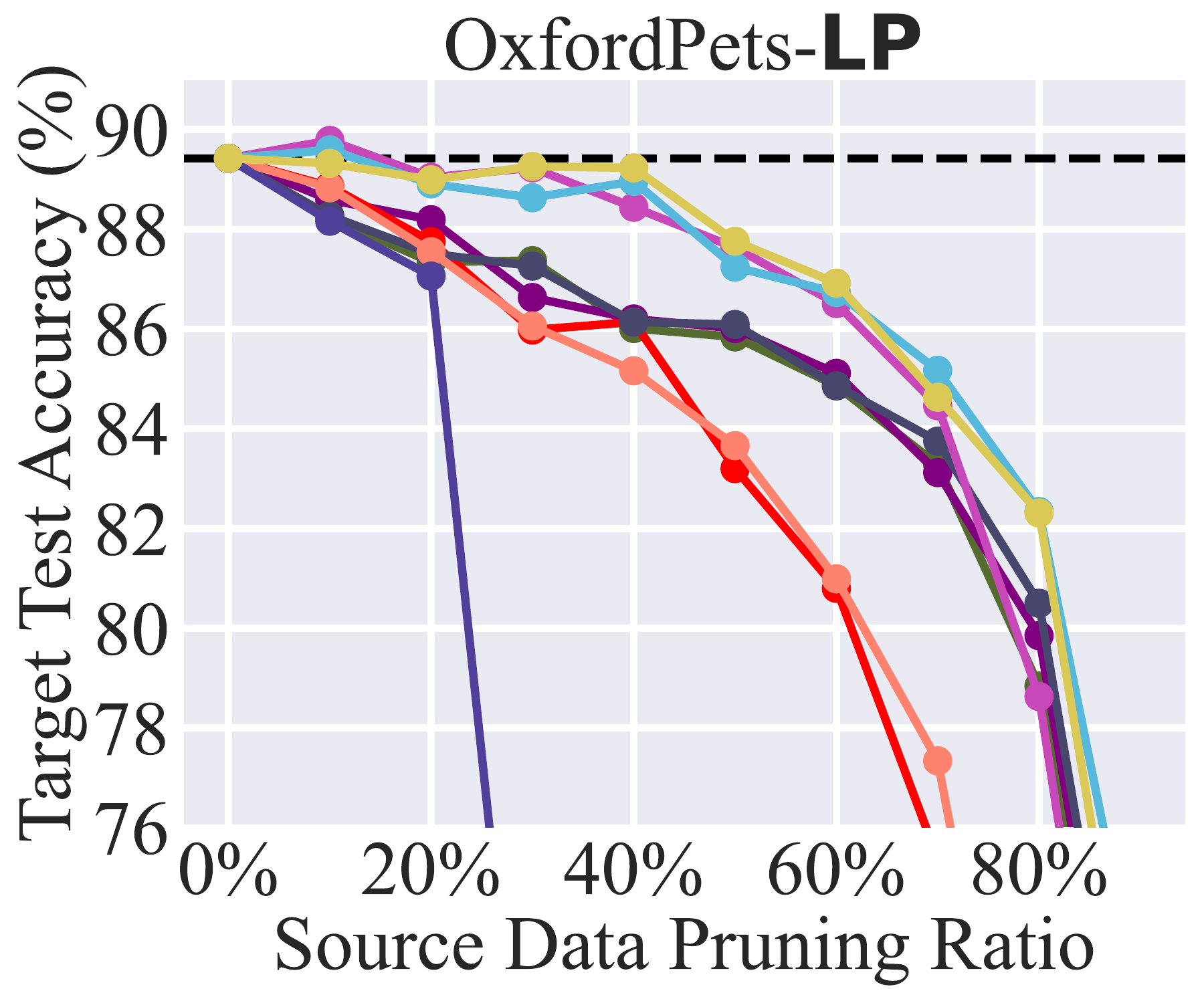}
\end{tabular}}
\vspace*{-3.5mm}
\caption{\footnotesize{
{Transfer learning accuracy of existing DP methods on ImageNet at different pruning ratios, where ResNet-101 is the source model, and linear probing (LP) is used for downstream finetuning on  the target datasets Flowers102 (\textbf{Left}) and OxfordPets (\textbf{Right}).}
}}
\label{fig: in_domain_motivation}
\vspace*{-4mm}
\end{wrapfigure}
\textbf{Conventional DP methods lack effectiveness on transfer learning.}
Given the challenges (\ding{182}-\ding{183}) posed by DP for transfer learning, we further conduct a preliminary study to investigate the effectiveness of existing 8 DP methods, including {\spr} \cite{schioppa2022scaling}, {\ssp} \cite{schioppa2022scaling}, {\gradnorm} \cite{paul2021deep}, {\eln} \cite{paul2021deep}, {\moderate} \cite{moderate2023xia}, {\forget} \cite{feldman2020neural}, {\infmax} \cite{feldman2020neural}, and {\infsum} \cite{feldman2020neural}. Our results show that these methods are \textit{unable} to yield significant improvements over \textit{random} source data pruning. \textbf{Fig.\,\ref{fig: in_domain_motivation}} shows the transfer learning performance of the ResNet-101 \cite{he2016deep} model trained on different pruned versions of the ImageNet dataset when LP-based finetuning is conducted on the downstream Flowers102 \cite{nilsback2008automated} and OxfordPets \cite{parkhi2012cats} datasets. As we can see, in transfer learning, random pruning is a solid baseline for various state-of-the-art DP methods, which have demonstrated superior performance to the former in the non-transfer learning regime. 
Therefore, it is crucial to develop an efficient and effective DP method specifically tailored for transfer learning.

\section{Label/Feature Mapping-based DP for Transfer Learning}
\label{sec: method}

In this section, we first introduce a simple yet powerful DP method called label mapping (\textbf{LM}) by leveraging the class-discriminative capability of a supervised source model. 
We then extend the concept of LM to feature mapping (\textbf{FM}), suitable for   {self-supervised} pretraining.

\textbf{LM-based DP for supervised pretraining.} 
Following the notations used in Sec.\,\ref{sec: prob}, we represent the model obtained through supervised learning on  the source dataset ($\gD_\gS$) as $g_{\mathcal{S}}\circ f_{\mathcal S}$.  This model predicts a source label   given an input example ($\mathbf x$).
In DP for transfer learning, the source data classes serve as the variables to be pruned. Thus, we express $\gD_\gS$ as $\gD_\gS = \{ \gC_{\mathcal{S}}^{(1)}, \ldots, \gC_{\mathcal{S}}^{(N)}\} $ for $N$ source classes,  $\gC_{\mathcal{S}}^{(i)}$ denotes the set of data points belonging to the source class $i$. Additionally, the pruner has access to the target dataset $\gD_\gT = \{ \mathbf t_1, \ldots, \mathbf t_n \}$, which consists of $n$ target data points.
Our objective is to utilize the information provided by $g_{\mathcal{S}}\circ f_{\mathcal S}$, $\gD_\gS$, and $\gD_\gT$ to devise a computationally efficient DP criterion. Importantly, our criterion should not involve model training, distinguishing it from the non-scalable approach in \cite{jain2022data}.
An important observation  by \cite{liu2019towards} in transfer learning is that transferred knowledge improves transferability. This insight is supported by the loss landscape analysis in transfer learning: The finetuned weights may remain within the flat basin of the pretrained weights for enhanced transfer learning performance. 

We next propose to extract the `transferred knowledge' by leveraging the class-discriminative capability of the source model $g_{\mathcal{S}}\circ f_{\mathcal S}$ on the target data samples in $\gD_\gT$. Specifically, we focus on monitoring the responsiveness of the source label predictions made by $g_{\mathcal{S}}\circ f_{\mathcal S}$ when using target samples $\{ \mathbf t_i \in \gD_\gT \}_{i=1}^n$ as input data. Here we resize these target samples to ensure their resolution alignment with source data.
For the $i$th source data class $\mathcal{C}_{\mathcal{S}}^{(i)}$, we then define its pruning score below:

\vspace*{-1em}
{\small{
\begin{align}
    \mathtt{s}_{\mathrm{LM}}(\gC_\gS^{(i)}) = \sum_{j=1}^{n}
    \mathbbm{1} ( g_{\mathcal{S}}\circ f_{\mathcal S} (\mathbf t_j) = i ), ~~ \text{for }  i =1,2,\ldots, N,
    \tag{LM}
    \label{eq: LM}
    \vspace*{-2em}
\end{align}
}}%
where 
$\mathbbm{1}(\cdot)$ represents an indicator function that evaluates to $1$ when {the condition $\cdot$} is satisfied, and $0$ otherwise.  We refer to the aforementioned formula as \ref{eq: LM} (Label Mapping) since the condition $g_{\mathcal{S}}\circ f_{\mathcal S} (\mathbf t_j) = i$ establishes a mapping between the predicted labels of the target samples $\{\mathbf t_j\}_{j=1}^n$ by the source model and the corresponding source labels ${i}$. The larger $\mathtt{s}_{\mathrm{LM}}$ in \eqref{eq: LM} signifies that a specific source class is more frequently utilized to interpret the target data. Consequently, this indicates a  more tightly connected relationship between the source class and the transferred knowledge. Therefore, we 
{prune} (or select) source data classes with {low}  (or high) $\mathtt{s}_{\mathrm{LM}}$ values.

Although \ref{eq: LM} is simple, it offers several advantages. \textit{Firstly}, it is highly efficient in  computation. Given a pre-trained source model $g_{\mathcal{S}}\circ f_{\mathcal S}$, the calculation of \ref{eq: LM} only requires forward passes through the model. In cases where obtaining the pretrained model in the source domain is challenging, our 
\begin{wrapfigure}{r}{85mm}
\vspace*{-4mm}
\centering
\begin{tabular}{cc}
\includegraphics[width=.45\linewidth]{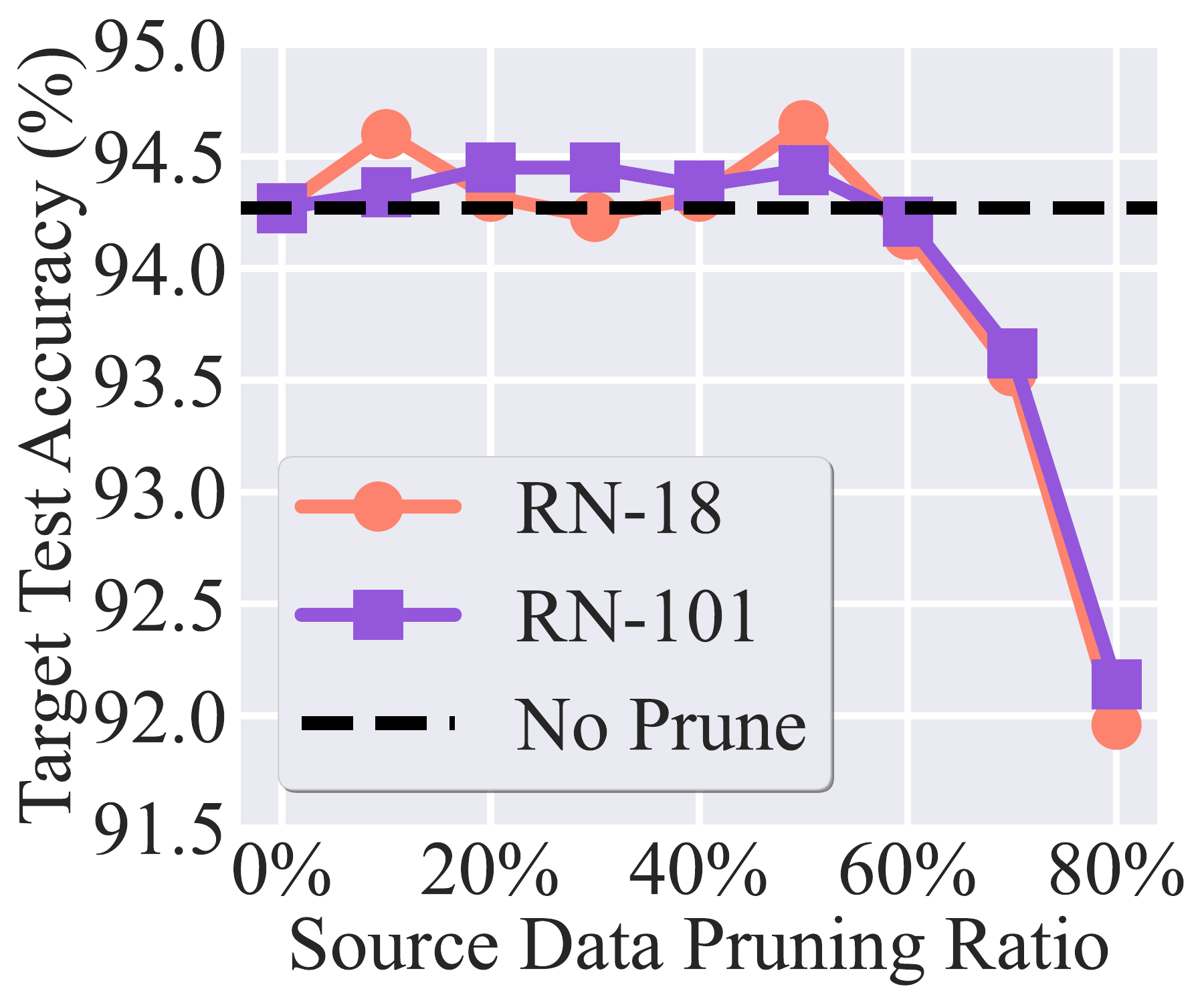}  
&\includegraphics[width=.45\linewidth]{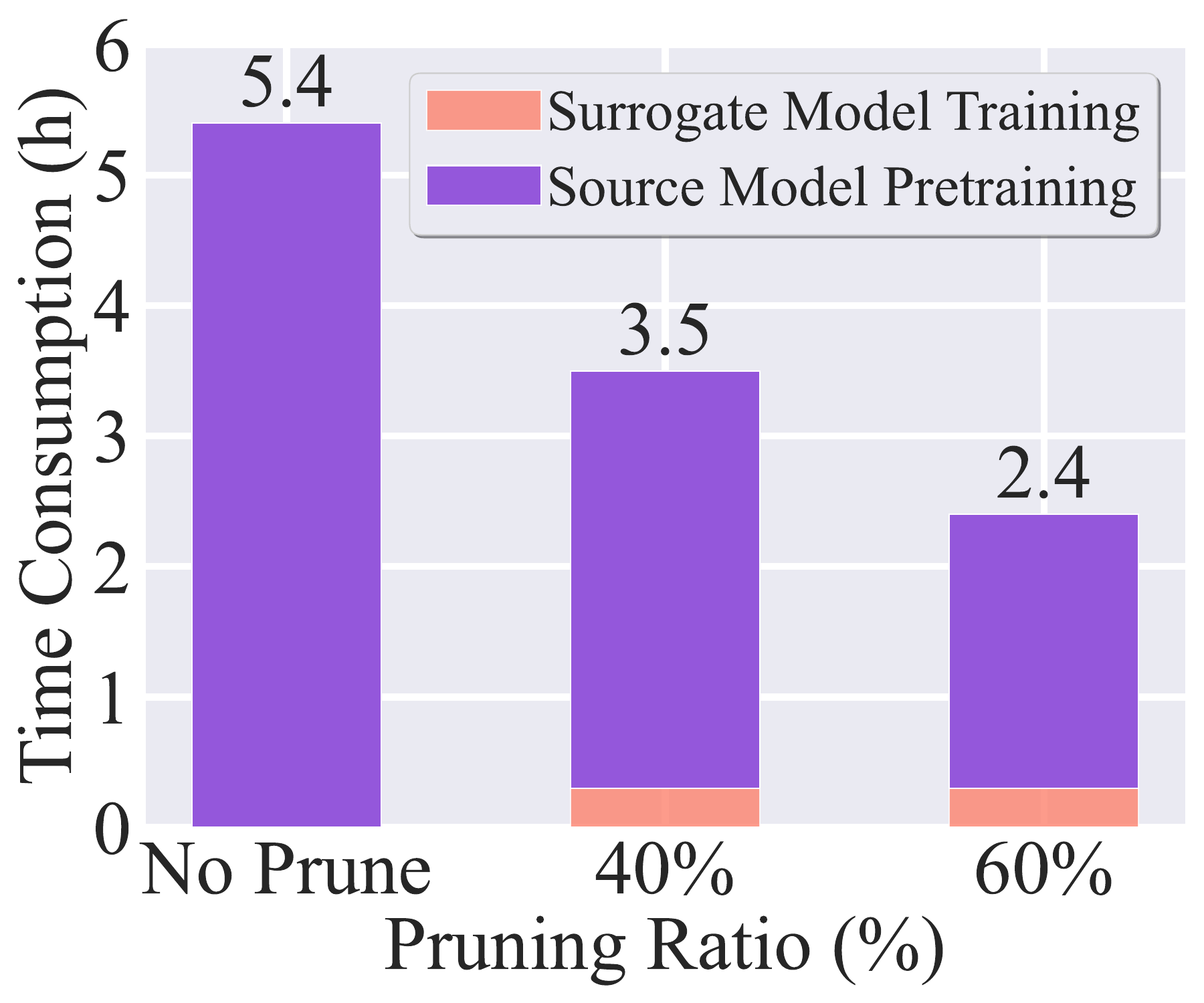} \vspace*{-1mm} \vspace*{-1mm}\\ 
{\footnotesize{(a) Downstream performance.}} 
&{\footnotesize{(b) Time consumption.}}
\end{tabular}
\vspace*{-2.5mm}
\caption{\footnotesize{Preliminary studies on the usage of surrogate models for \ref{eq: LM}. (\textbf{a}) The downstream performance (on Flowers102) of using the source model ResNet-101 trained on the pruned ImageNet delivered by \ref{eq: LM} at different pruning ratios. 
Here \ref{eq: LM} is conducted using either RN-101 or a smaller surrogate model ResNet-18.
(\textbf{b}) Computation time decomposition analysis for obtaining the pretrained model using ResNet-18 as surrogate model with different pruning ratios.
}}
\label{fig: preliminary_LM}
\vspace*{-4mm}
\end{wrapfigure}
proposal supports an alternative approach: training a \textit{smaller and simpler surrogate model} to carry out  \ref{eq: LM}. This surrogate model can effectively replace the complex pretrained model $g_{\mathcal{S}}\circ f_{\mathcal S}$ and facilitate the efficient execution of the pruning process.
As we show in \textbf{Fig.\,\ref{fig: preliminary_LM}a}, employing ResNet-18 \cite{he2016deep} is sufficient to successfully prune the source dataset (ImageNet \cite{russakovsky2015imagenet}). The resulting DP scheme remains effective to improve transfer learning utilizing other larger source models, such as ResNet-101 \cite{he2016deep}.
\textit{In addition}, \textbf{Fig.\,\ref{fig: preliminary_LM}b} shows that the computational overhead incurred by training the small surrogate model (ResNet-18) for DP is insignificant, compared to the time saved during the pretraining phase of a larger source model (ResNet-101) on the pruned dataset for transfer learning. 
\textit{Lastly}, pretraining on the subset found by LM can guide the source model towards a flatter region in the downstream loss landscape, (see results in \textbf{Fig.\,\ref{fig: app_sharpness}}).
The source model trained on the LM-pruned dataset achieves a higher flatness score than baselines, which aligns with the understanding of transfer learning in \cite{liu2019towards}.

\textbf{FM-based DP framework for self-supervised pretraining.} 
The requirement for \textit{labeled} source data in  \ref{eq: LM} may pose limitations on the application of DP methods, particularly in the context of self-supervised learning (\textbf{SSL})-based pertaining. To address this limitation, we introduce a new method called FM (Feature Mapping). Unlike LM, FM determines the DP scheme using only the feature extractor network $f_{\gS}$, the unlabeled source dataset $\gD_\gS$, and the target dataset $\gD_\gT$. This allows us to overcome the dependence on labeled source data, making FM applicable in SSL scenarios.
The inspiration for FM is derived from the deep clustering technique \cite{caron2018deep, yan2020clusterfit, fan2021does} operating in the representation space, which can generate \textit{pseudo source labels} using \textit{cluster indices} provided by \textit{e.g.}, $K$-means. With the assistance of deep clustering, we can represent the unlabeled source dataset $\gD_\gS$ as $\gD_\gS = \{\gC_\gS^{(1)}, \gC_\gS^{(2)}, \ldots, \gC_\gS^{(K)}\}$, where $\gC_\gS^{(k)}$ denotes the set of source data samples within cluster $k$.
Building upon the similar spirit as LM, we propose ranking the importance of pseudo source classes for DP by evaluating the source feature extractor's responsiveness   to target data samples.
To achieve this objective, we quantify the responsiveness of $f_{\gS}$ for a target sample $\mathbf t$ as follows:

\vspace*{-4mm}
{\small{
\begin{align}
    r(\mathbf t) = \argmin_{k\in[K]} ~ \left \| f_\gS(\mathbf t) -  {\mathbb E_{\mathbf x \in \mathcal{C}_{\gS}^{(k)}} [f_{\gS} (\mathbf x)]
    }
    \right \|_2,
    \label{eq: FM_data}
    \vspace*{-4mm}
\end{align}
}}%
where $\mathbb E_{\mathbf x \in \mathcal{C}_{\gS}^{(k)}} [f_{\gS} (\mathbf x)]$ is the  centroid of  source data   within  the cluster $k$, and $r(\mathbf t)$ is  the nearest pseudo label as the responsiveness of  $f_\gS$ against $\mathbf t$. The FM score then integrates \eqref{eq: FM_data} with \eqref{eq: LM}:

\vspace*{-5mm}
{\small{
\begin{align}
    \mathtt{s}_{\mathrm{FM}}(\gC_\gS^{(i)}) = \sum_{j=1}^{n}
    \mathbbm{1} ( r (\mathbf t_j) = i ), ~~ \text{for }  i =1,2,\ldots, K,
    \tag{FM}
    \label{eq: FM}
\end{align}
}}%
where different from \eqref{eq: LM}, $K$ represents the number of pseudo source classes produced by deep clustering, and $r(\mathbf t_j)$ corresponds to the source feature extractor's prediction on $\mathbf t_j$. It is important to note that the value of $K$ is a free parameter of the deep clustering process. Our empirical study in Fig.\,\ref{fig: app_cluster_num_study} shows that \ref{eq: FM}  is quite robust to the choice of $K$ without sacrificing the benefit of DP for transfer learning compared to using the unpruned source dataset. Lastly, it is worth mentioning that   FM   can also be applied in the context of supervised pretraining   by specifying data labels as clusters.

\section{Experiments}
\label{sec: exp}
In this section, we provide a comprehensive set of experiments and analyses to showcase the effectiveness of our proposed methods (LM and FM) in diverse transfer learning scenarios.

\subsection{Experiment setup}
\textbf{Datasets and models.} In line with existing transfer learning benchmarks \cite{jain2022data, evci2022head2toe}, we utilize ImageNet-1K \cite{russakovsky2015imagenet} for pretraining and \textbf{8} datasets as downstream tasks. These datasets include DTD \cite{cimpoi2014describing}, Flowers102 \cite{nilsback2008automated}, UCF101 \cite{soomro2012ucf101}, Food101 \cite{bossard2014food}, SUN397 \cite{xiao2010sun}, OxfordPets \cite{parkhi2012cats}, StanfordCars \cite{krause20133d}, and CIFAR10 \cite{krizhevsky2009learning}. Please refer to Tab.\,\ref{tab: app_downstream_data} for more details about the datasets. As discussed in Sec.\,\ref{sec: method}, we utilize ResNet-18 (RN-18) \cite{he2016deep} as the surrogate source model for  pruning source classes. This method significantly reduces the computational cost associated with DP, making the process more efficient.
Subsequently, a range of larger models, \textit{e.g.,} ResNet-101 (RN-101) and ViT-B/16 \cite{dosovitskiy2020image}, are trained on the (pruned) ImageNet and then finetuned on  downstream tasks.

\textbf{Baselines, training, and evaluation.}
{By examining the performance of the existing \textbf{8} DP baselines as shown in Fig.\,\ref{fig: in_domain_motivation}, our experiments focus on two of the most effective methods:}
\ding{172} {\gradnorm} \cite{paul2021deep} and \ding{173} {\moderate} \cite{moderate2023xia}, together with \ding{174} {\random} (the random pruning strategy).  In Fig.\,\ref{fig: app_complete_main_figure}, we show more results of the rest DP baselines.
Unfortunately, we are unable to include the existing data attribution method \cite{jain2022data} as our baseline, as it does not release its pruning results, and we are unable to repeat its experiments due to the need for intensive computations.
Unless specified otherwise, we focus on the supervised pretraining setting in the experiments.
For   self-supervised pretraining, we follow the implementation of \textsc{MoCov2} \cite{chen2020improved}. The finetuning strategies employed include LP (linear probing), which finetunes the classification head with fixed feature extractor, and FF (full finetuning), which finetunes the entire source model. 
For FM-based DP method, we utilize K-means clustering to group the ImageNet training data points into $K=2000$ clusters for the computation of \eqref{eq: FM_data}. 

In accordance with the terminology convention in model pruning \cite{frankle2018lottery, chen2020lottery, chen2021lottery, zhang2022advancing}, we refer the term `\textbf{winning subsets}' to the obtained source subsets that do \textit{not} compromise downstream performance. Among these winning subsets, we identify the one with the highest pruning ratio as the `\textbf{best winning subset}'. We then evaluate the performance of DP methods from the following aspects: the downstream performance of the model pretrained on the pruned source dataset obtained by various DP methods, and  the pruning ratio of the best winning subset achieved by DP methods, accompanied by the corresponding time saved in the pretraining phase.

\begin{figure}[hbt]
\vspace*{-2mm}
\centerline{
\begin{tabular}{cccc}
    \hspace*{-2mm} \includegraphics[width=.23\textwidth,height=!]{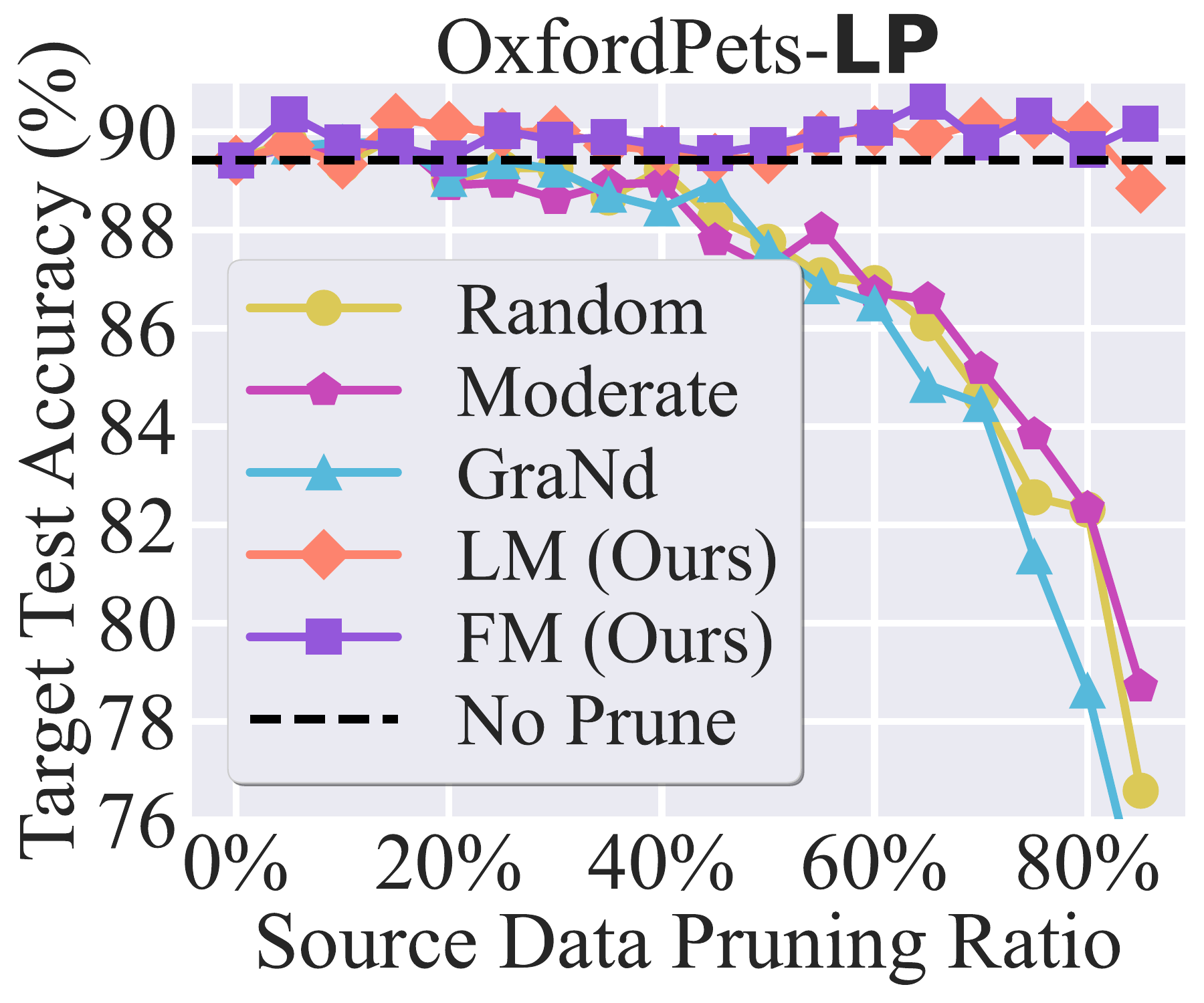} &
    \hspace*{-4mm}  \includegraphics[width=.23\textwidth,height=!]{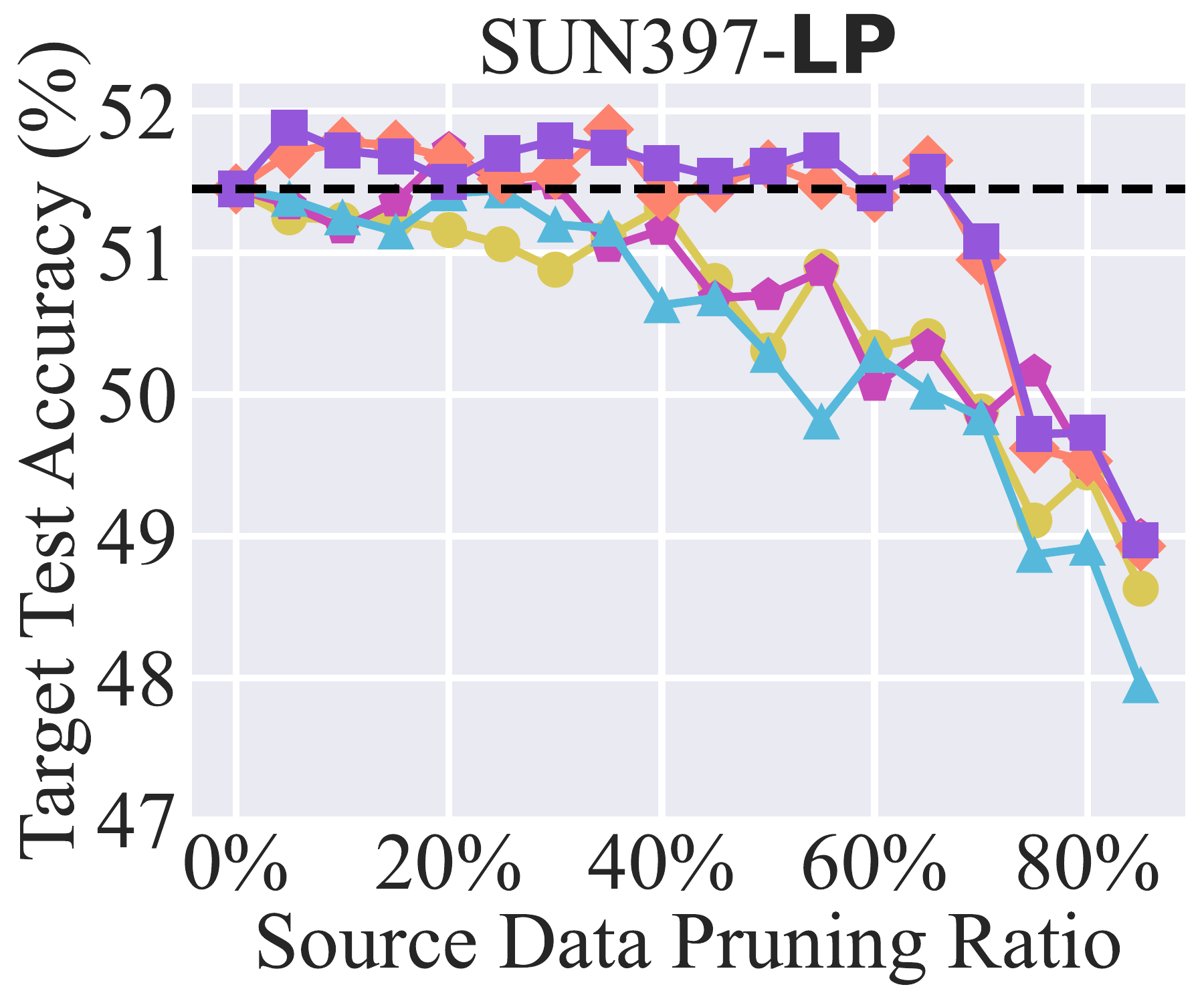} &
    \hspace*{-4mm}  \includegraphics[width=.23\textwidth,height=!]{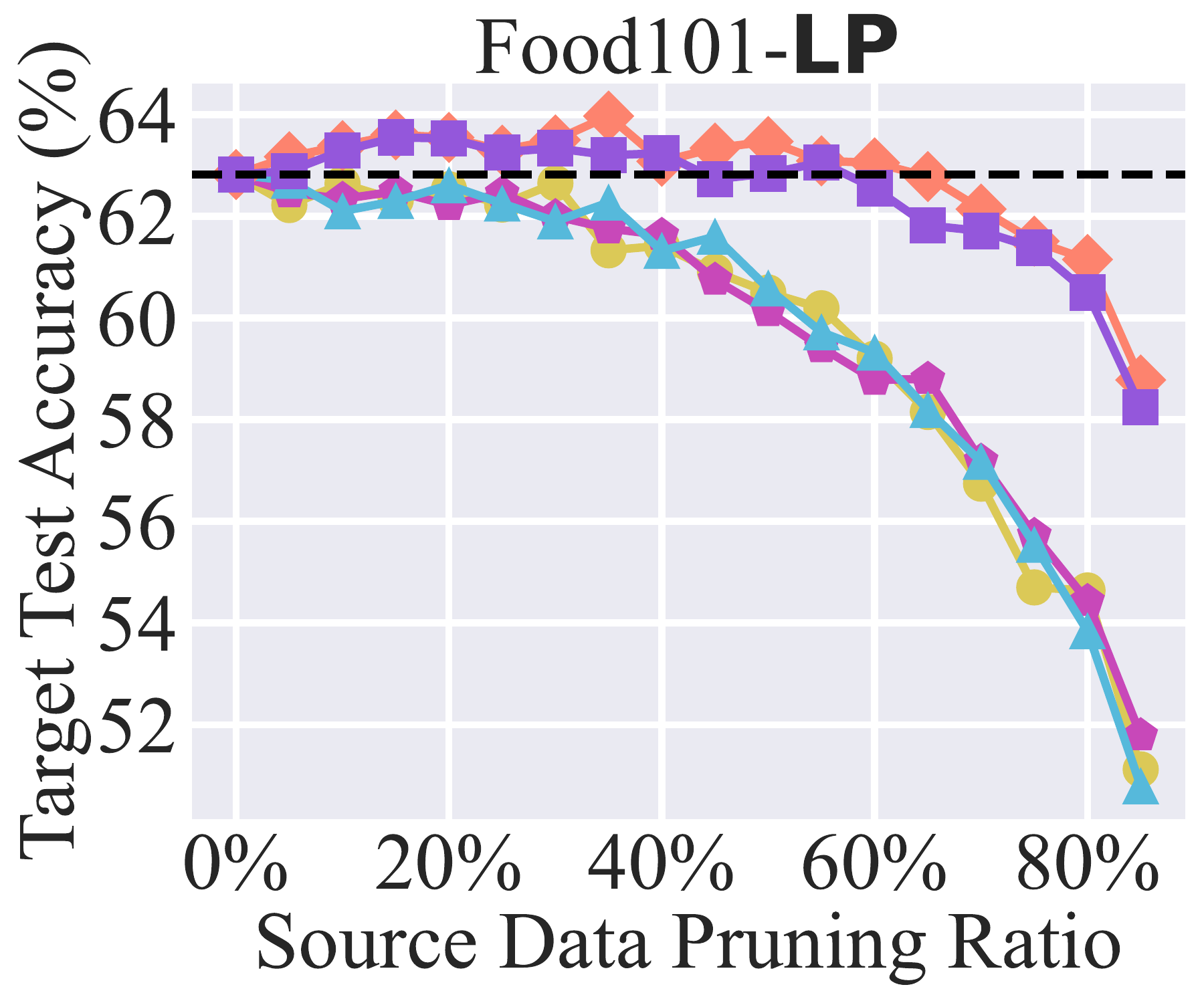} &
    \hspace*{-4mm} \includegraphics[width=.23\textwidth,height=!]{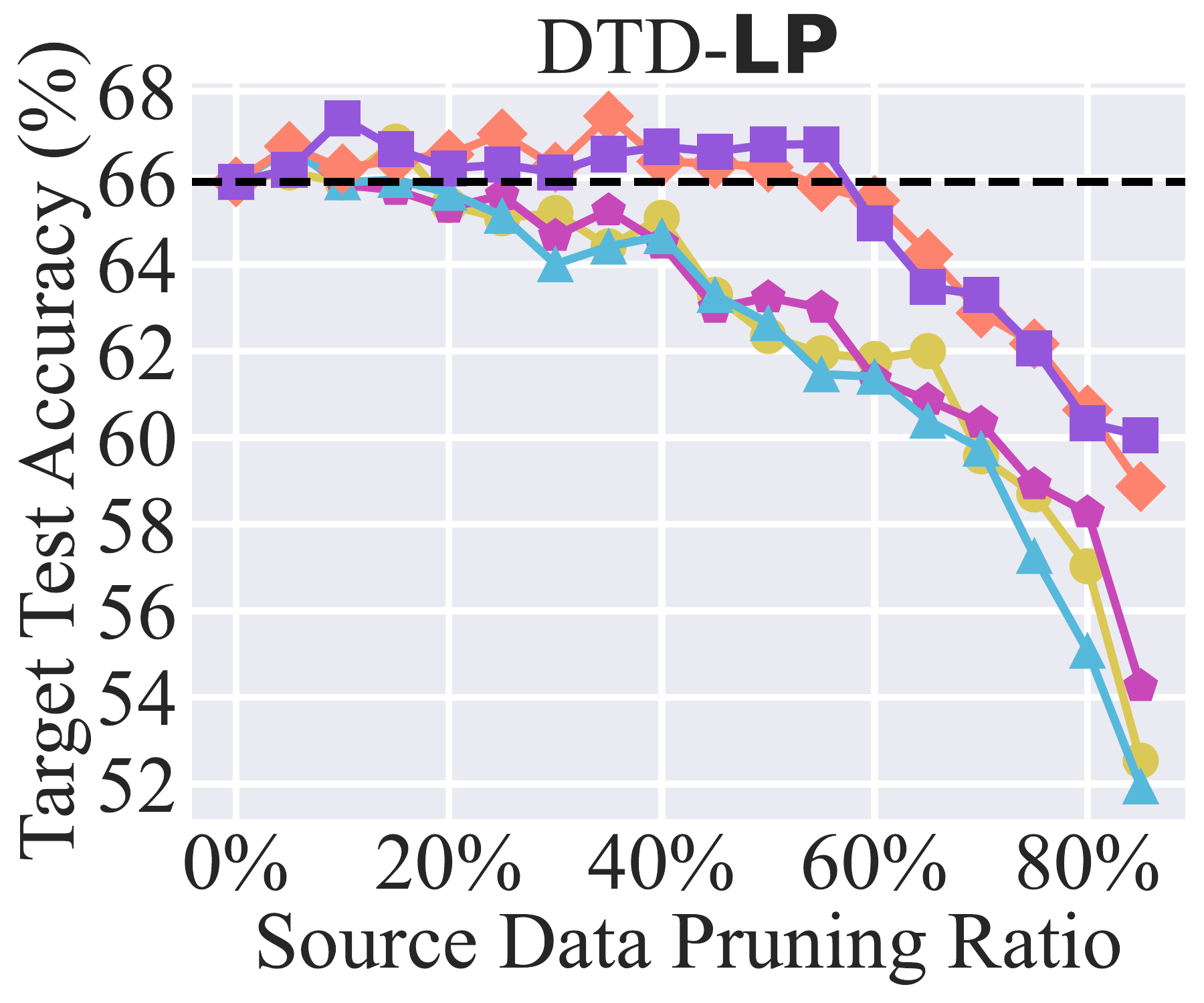} \\

    \includegraphics[width=.23\textwidth,height=!]{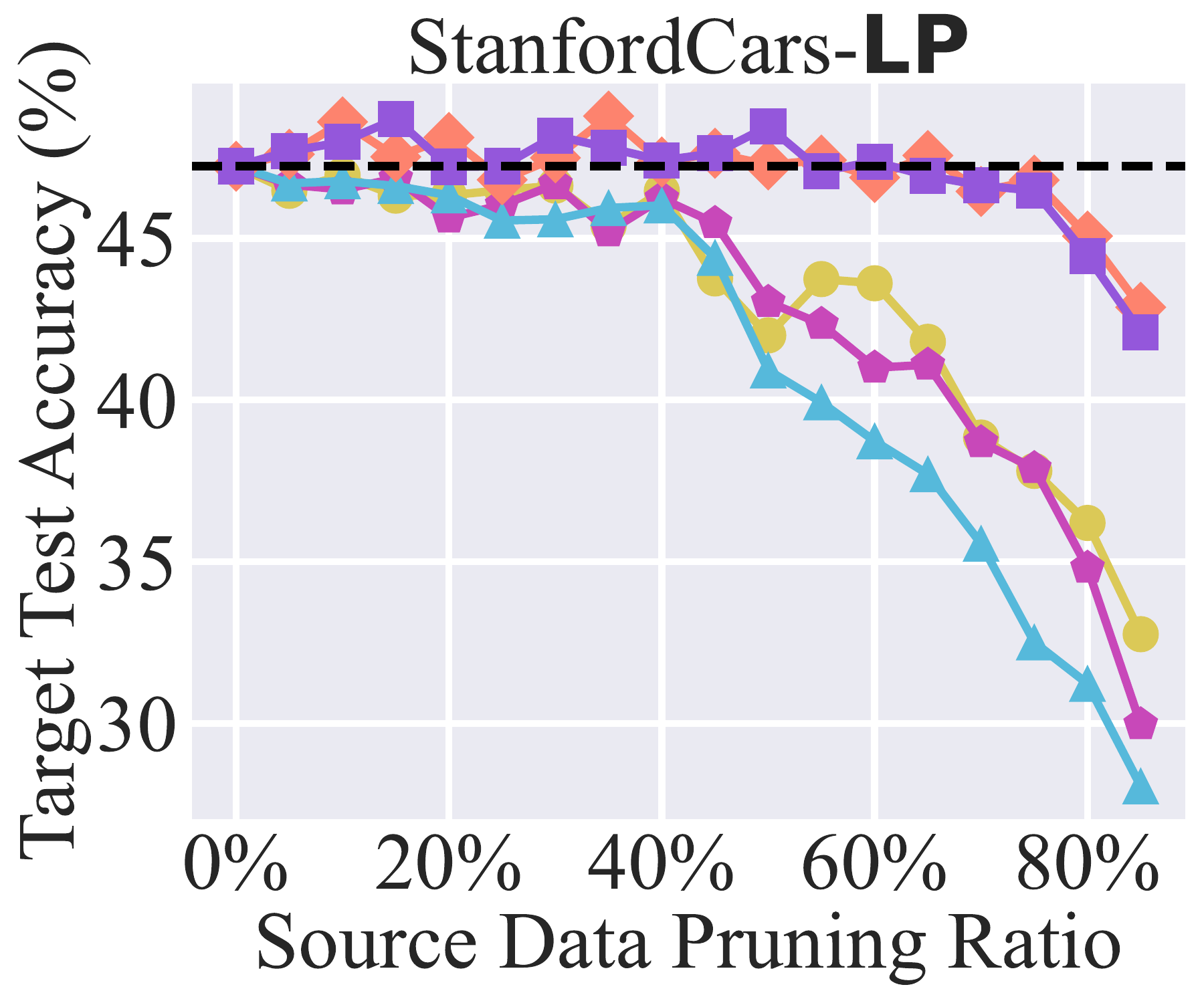} &
    \hspace*{-4mm}  \includegraphics[width=.23\textwidth,height=!]{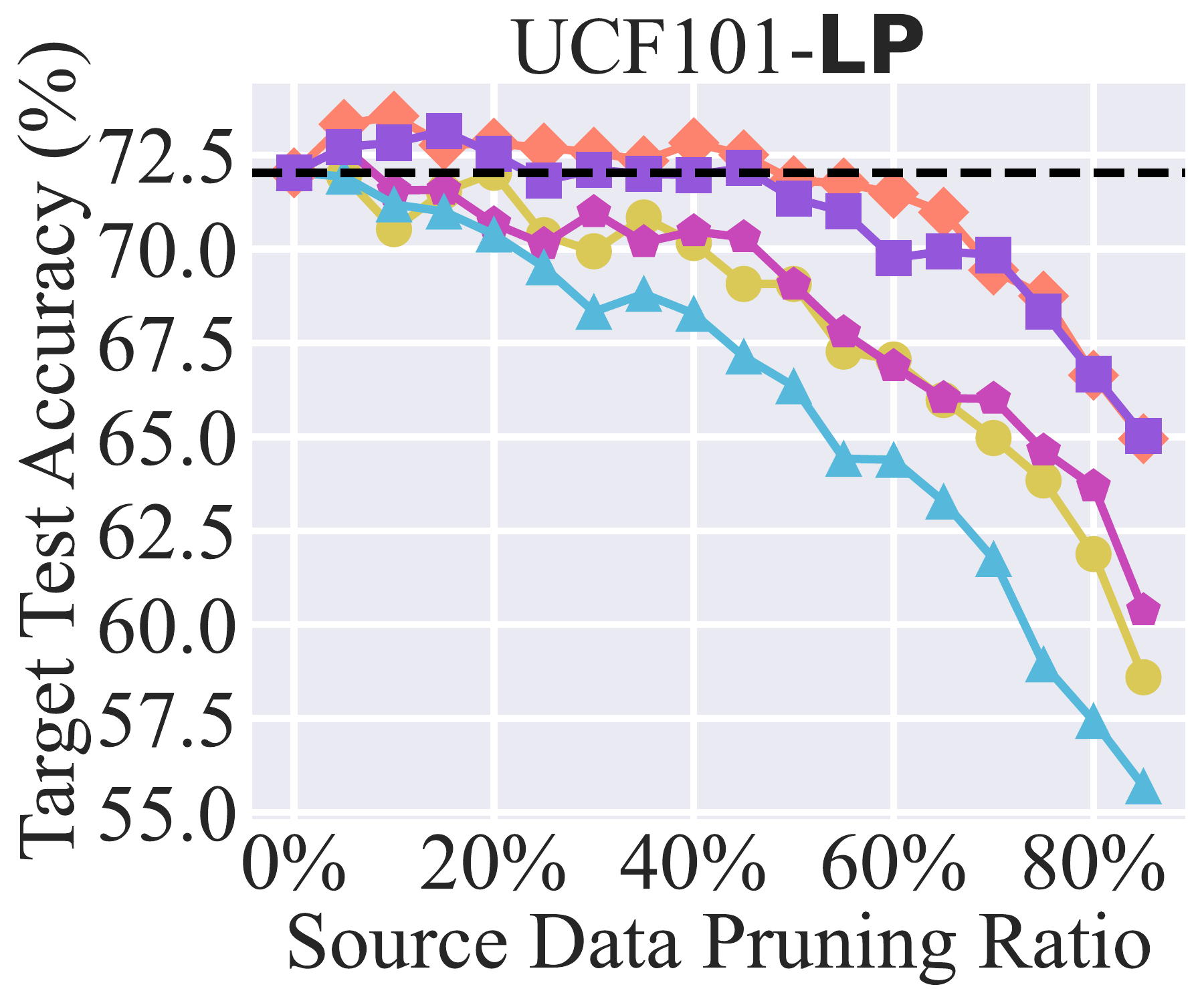} &
    \hspace*{-4mm}  \includegraphics[width=.23\textwidth,height=!]{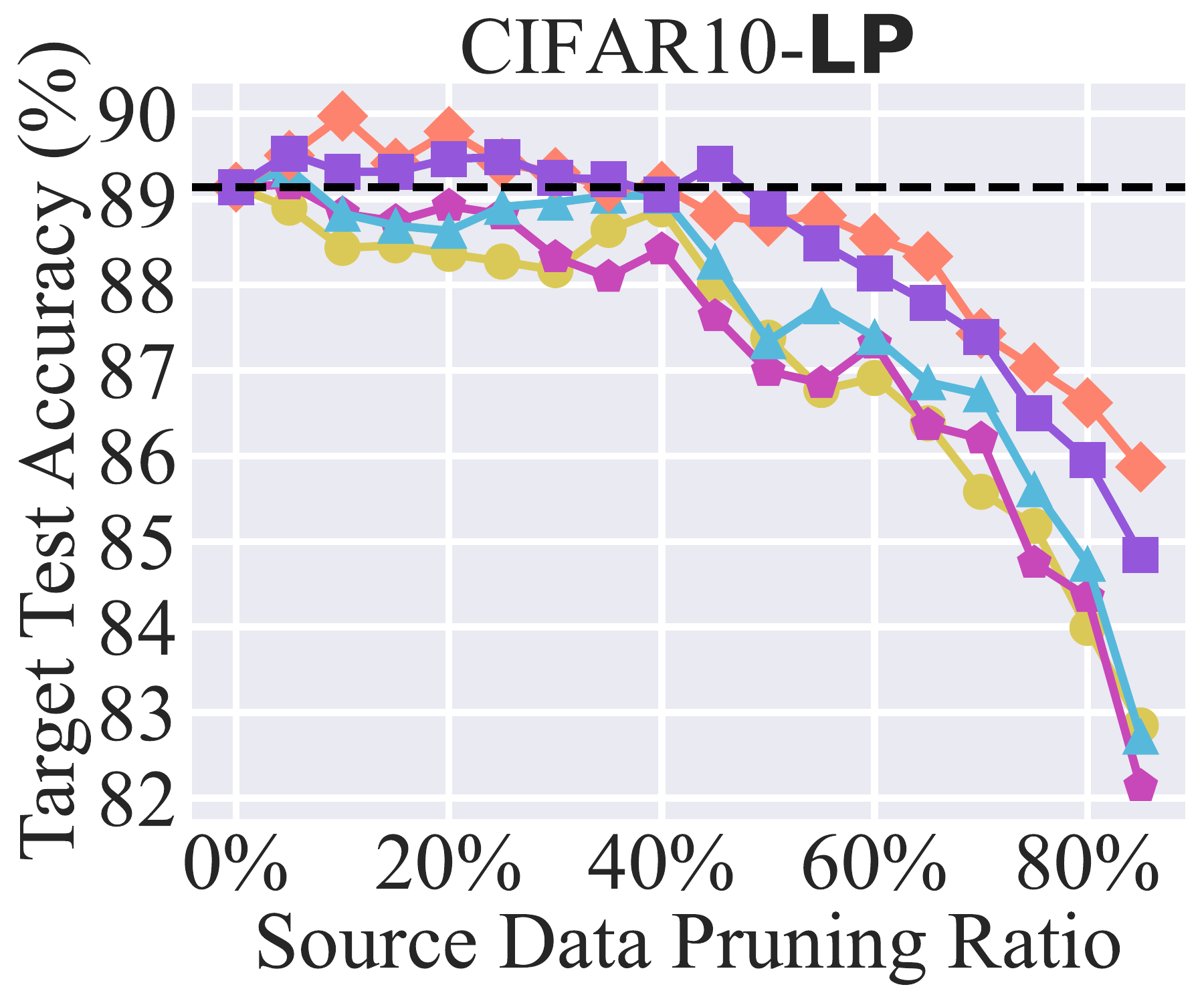} &
    \hspace*{-4mm} \includegraphics[width=.23\textwidth,height=!]{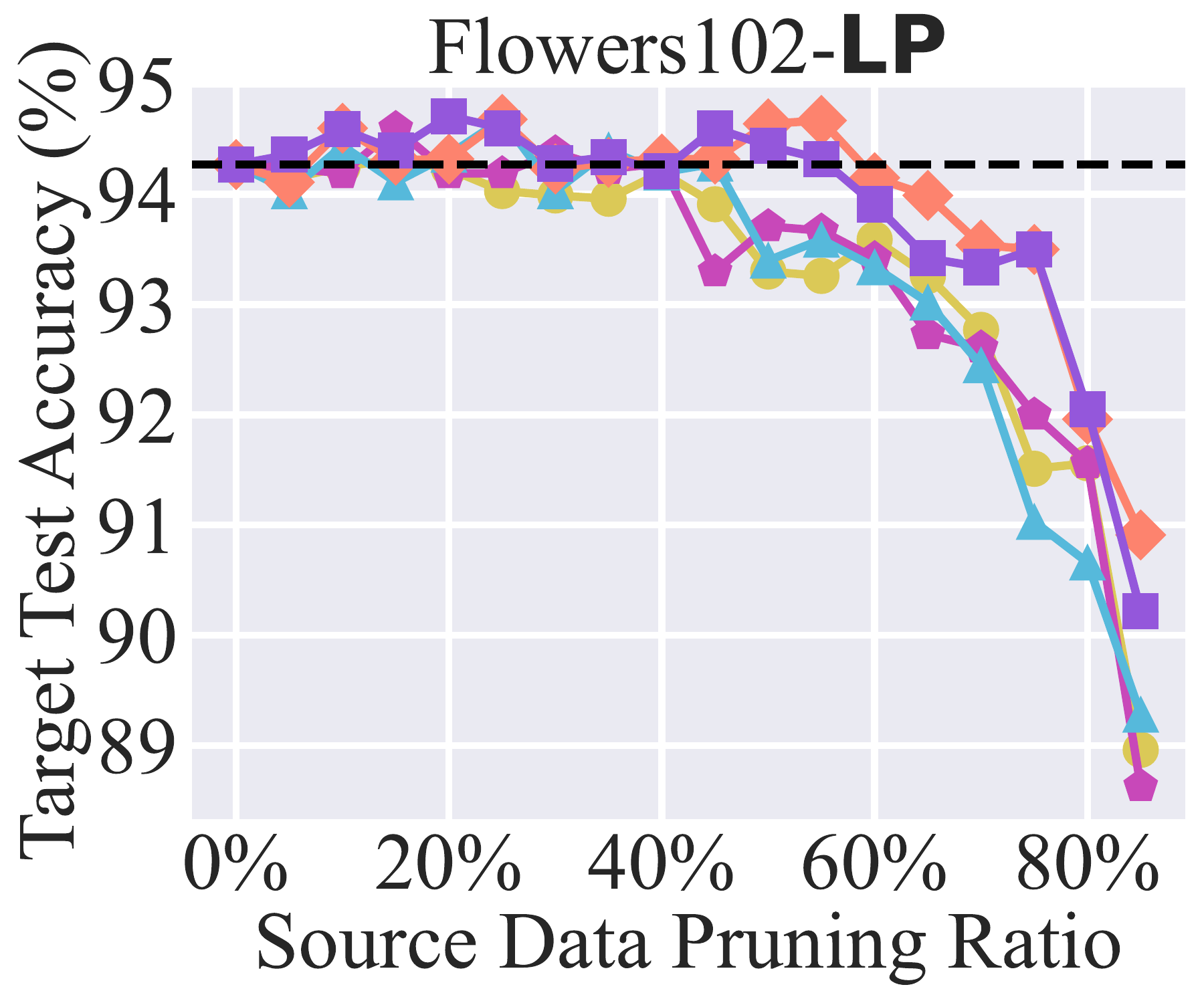} \\
    
    \hspace*{-2mm} \includegraphics[width=.23\textwidth,height=!]{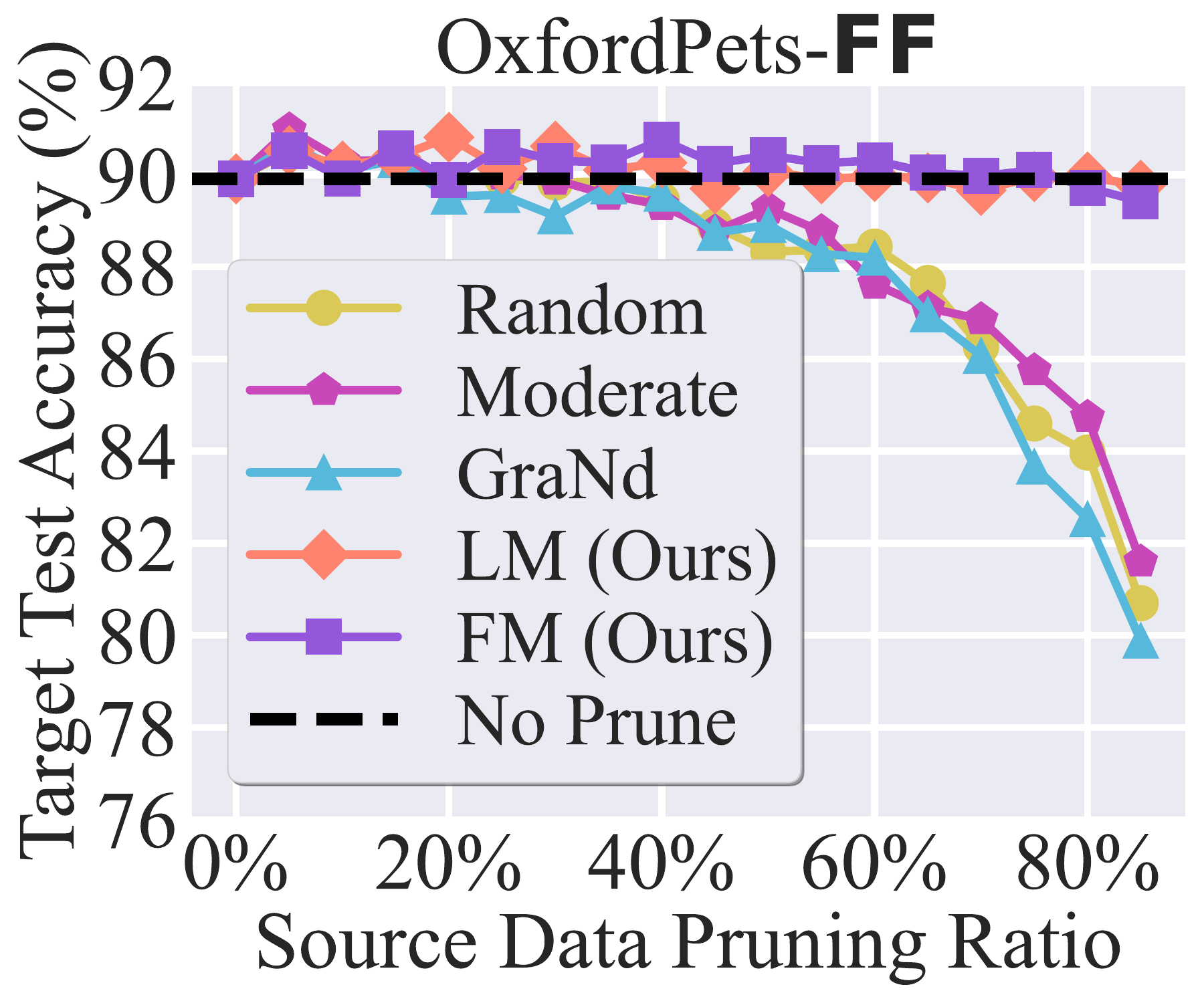} &
    \hspace*{-4mm}  \includegraphics[width=.23\textwidth,height=!]{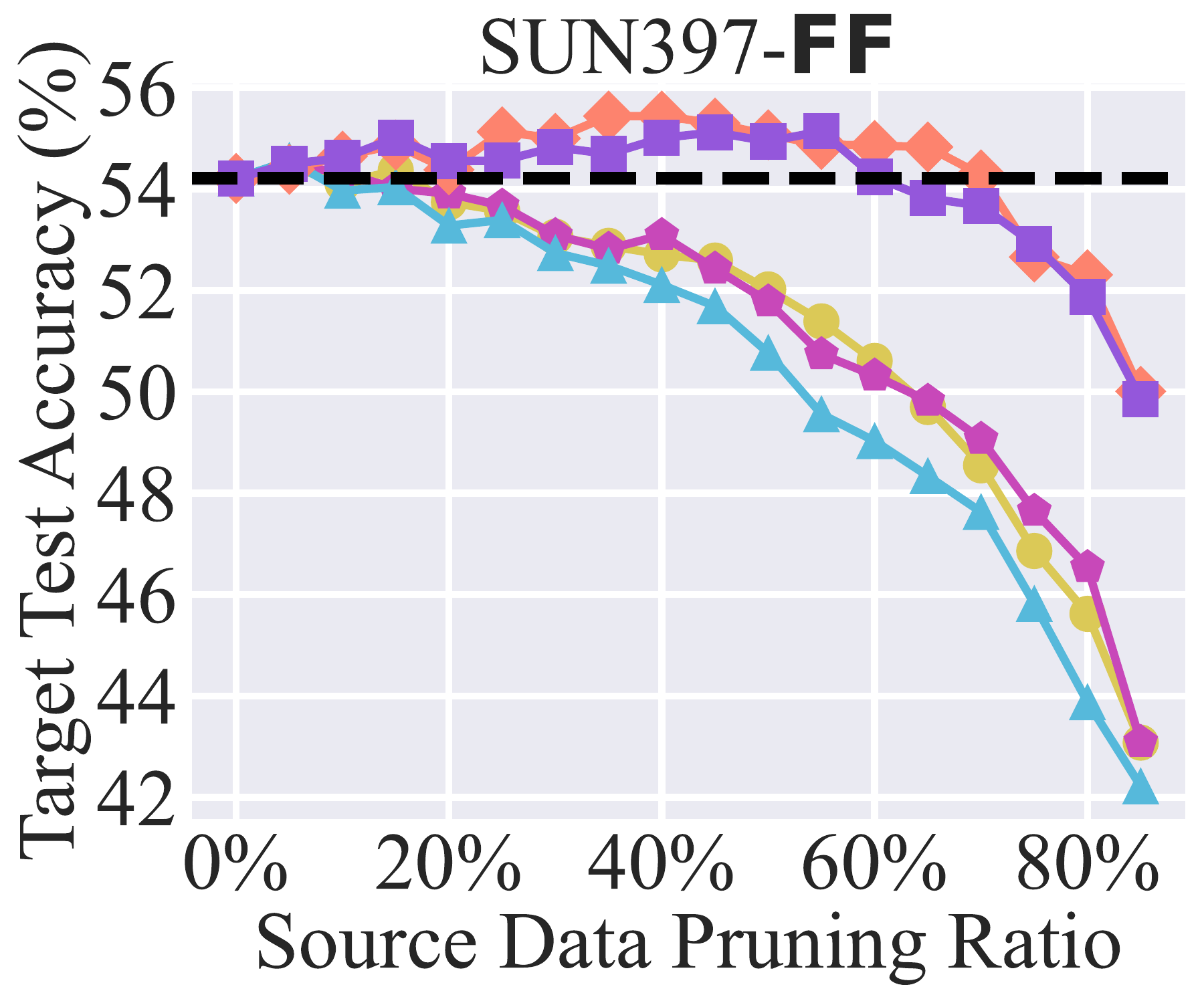} &
    \hspace*{-4mm}  \includegraphics[width=.23\textwidth,height=!]{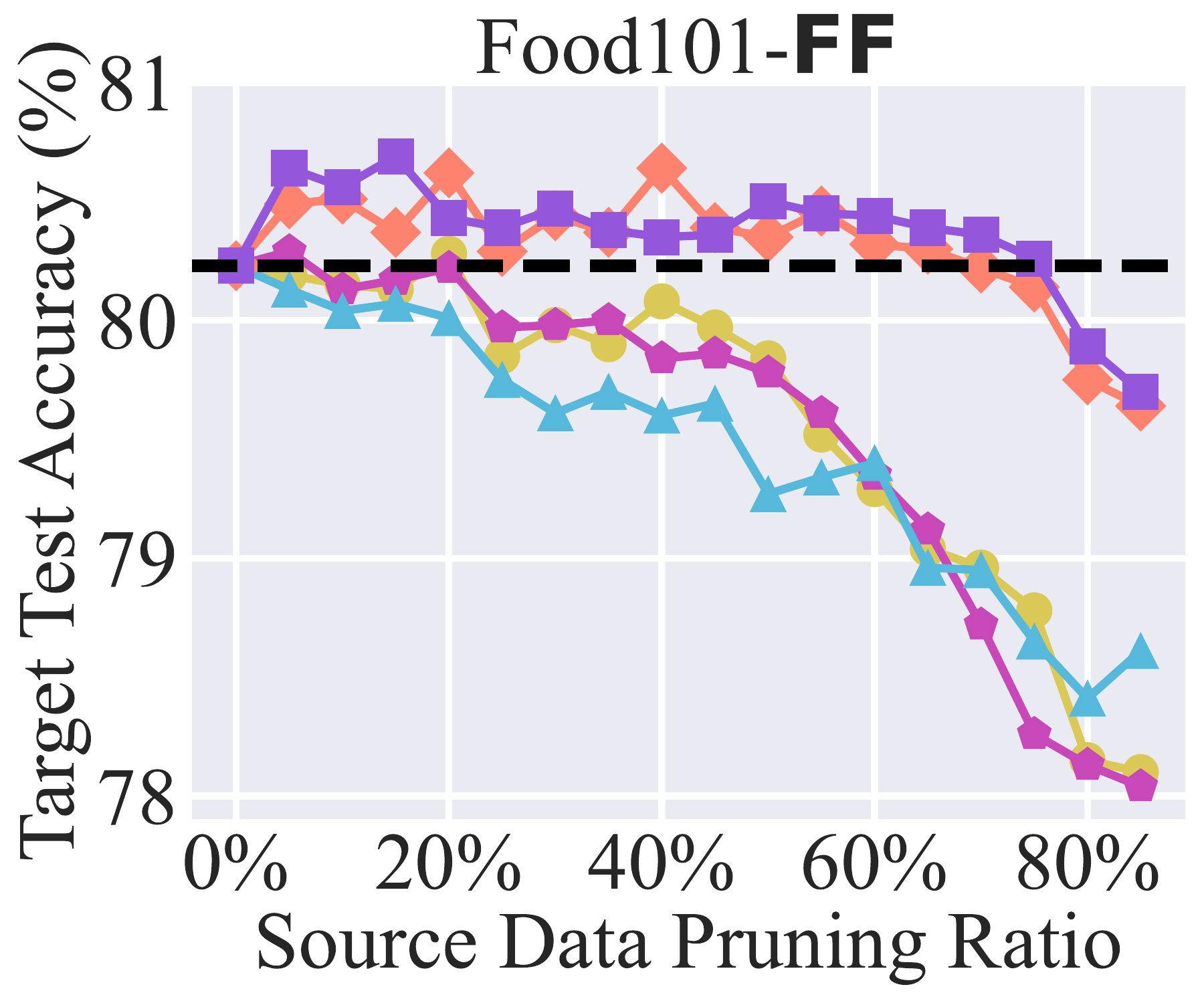} &
    \hspace*{-4mm} \includegraphics[width=.23\textwidth,height=!]{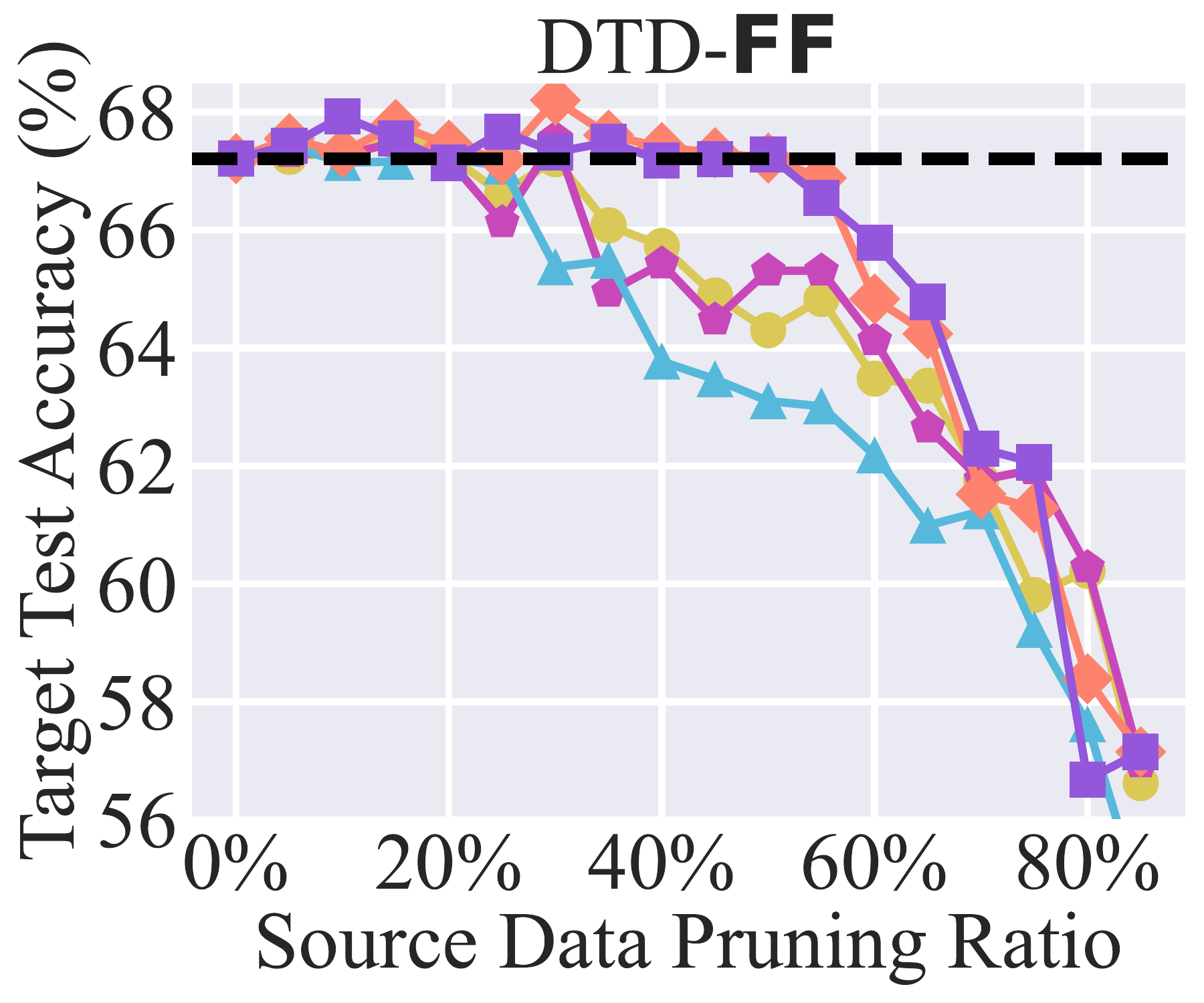} \\

    \hspace*{-2mm} \includegraphics[width=.23\textwidth,height=!]{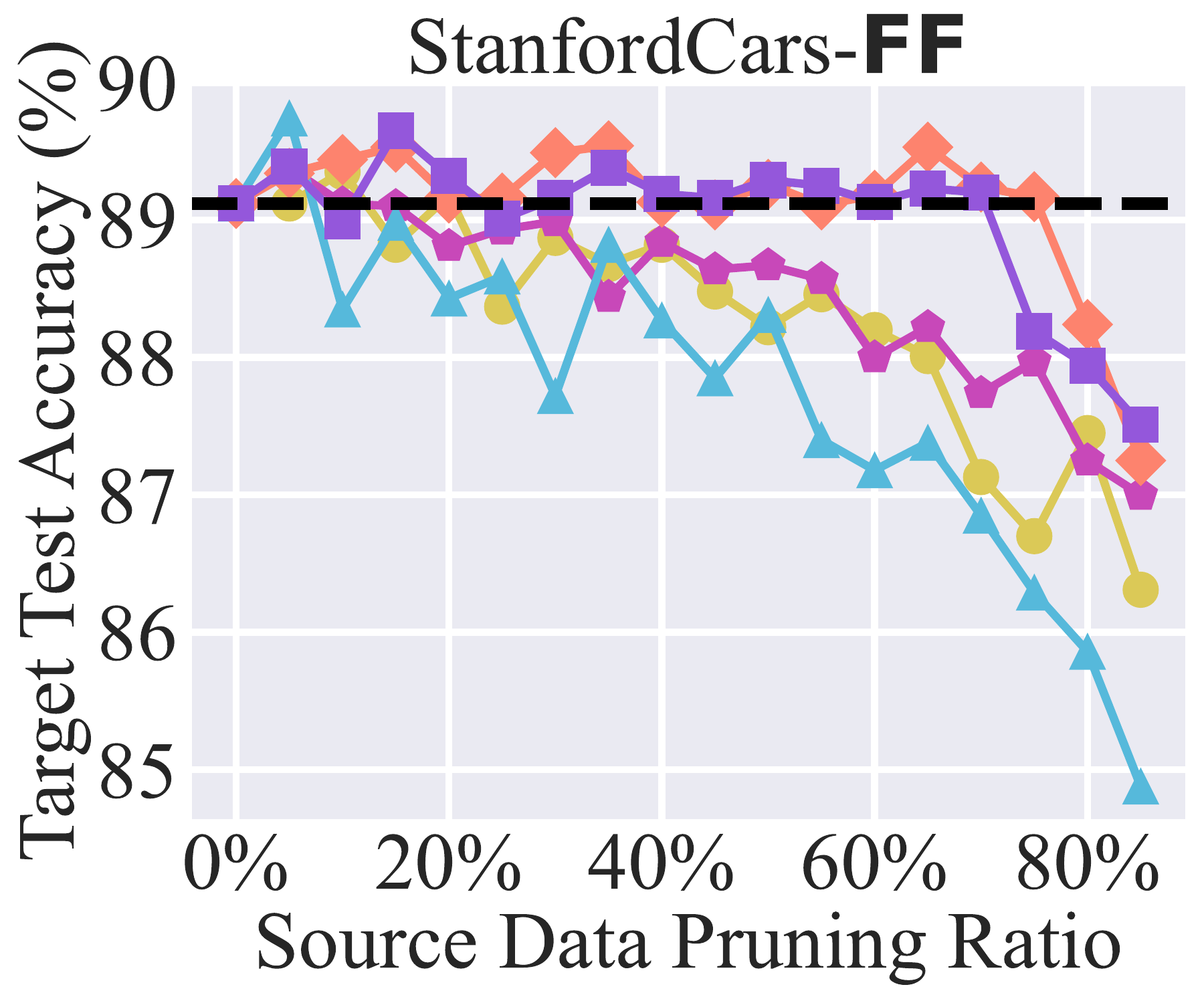} &
    \hspace*{-4mm}  \includegraphics[width=.23\textwidth,height=!]{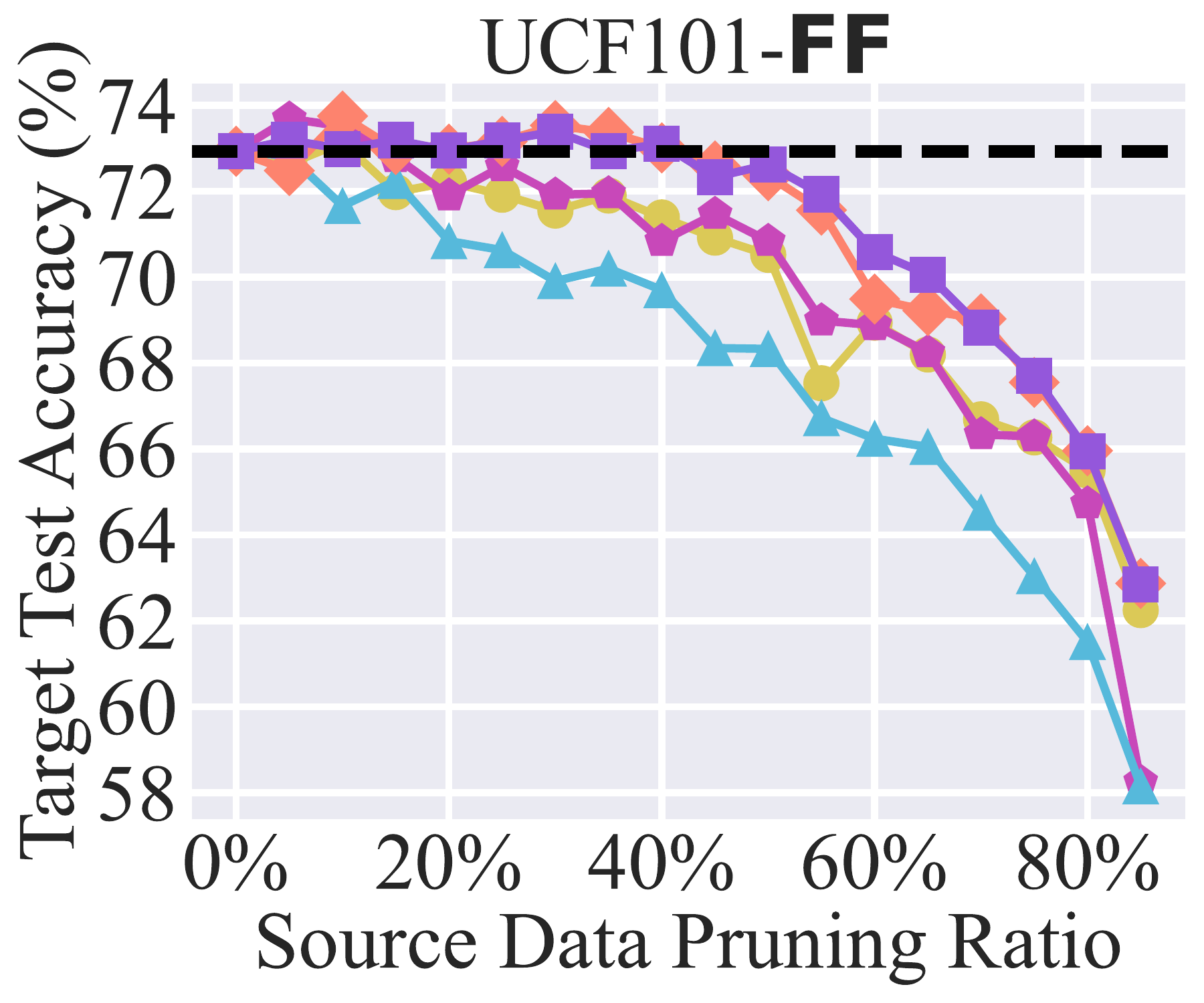} &
    \hspace*{-4mm}  \includegraphics[width=.23\textwidth,height=!]{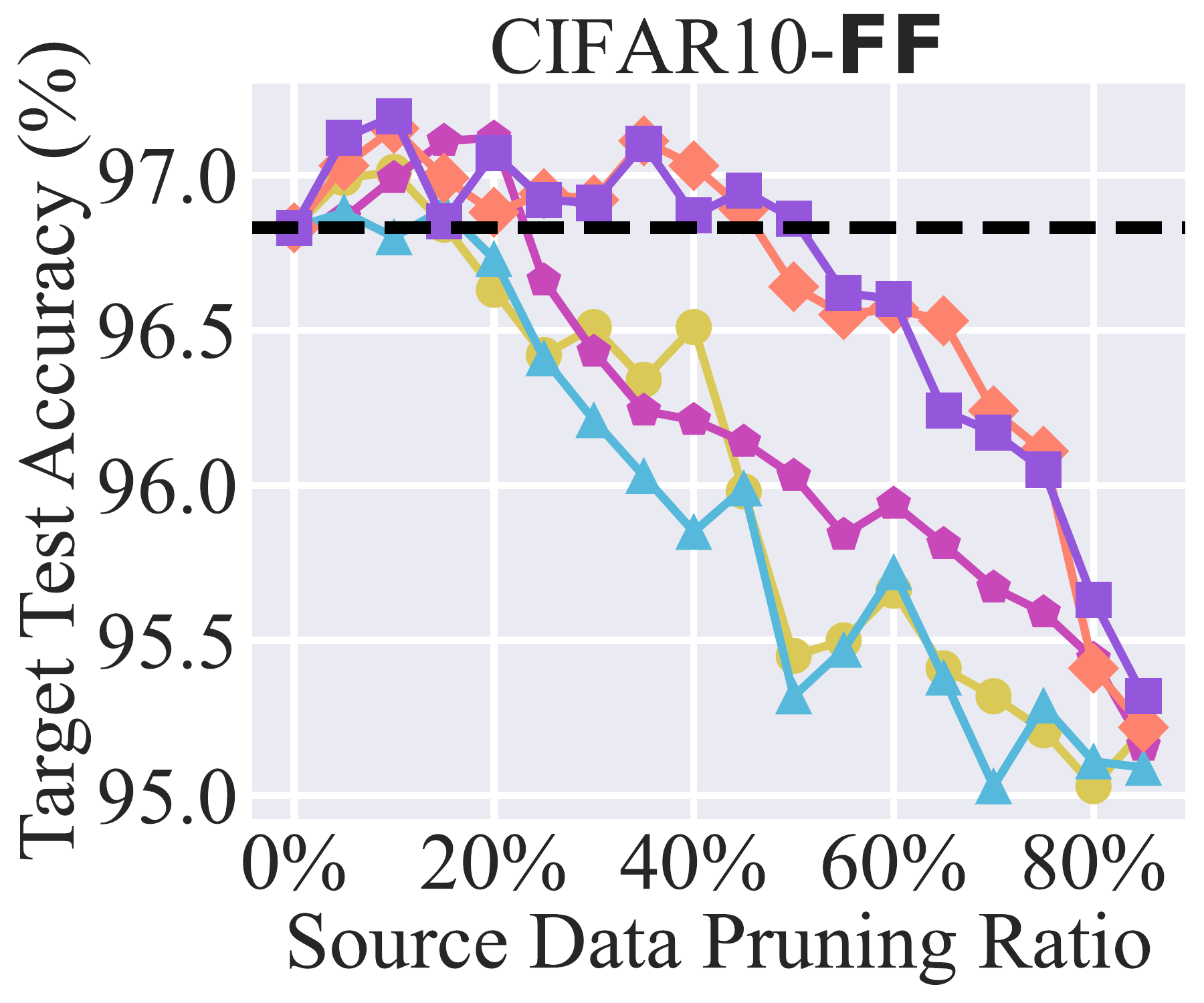} &
    \hspace*{-4mm} \includegraphics[width=.23\textwidth,height=!]{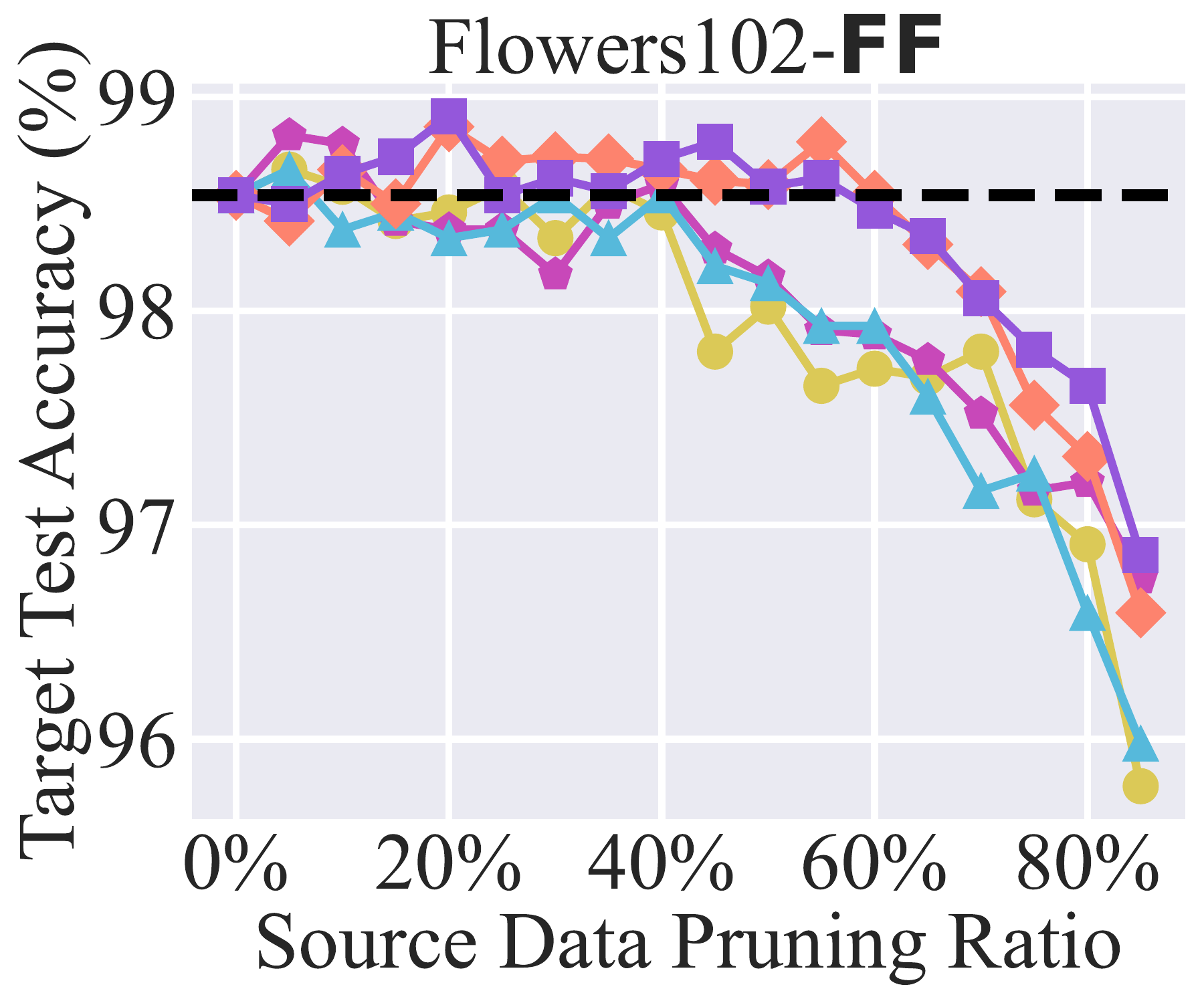} \\
\end{tabular}}
\vspace*{-2mm}
\caption{Source dataset pruning trajectory given by downstream testing accuracy (\%) vs. source dataset pruning ratio (\%) in the supervised pretraining setting.  Here the source model {RN-101} is trained on each full/pruned source dataset (ImageNet) and finetuned on different downstream tasks through LP and FF. The downstream performance without pruning (No Prune) is marked with the black dashed line. Results are averaged over three independent trials (see exact numbers and variances in Tab.\,\ref{tab: app_number_main_lp}). 
}
\vspace*{-5mm}
\label{fig: main_lp}
\end{figure}
\subsection{Experiment results}

\textbf{LM/FM improves transfer learning accuracy by identifying `winning subsets'.} 
We first  showcase the significant improvements achieved by our proposed DP methods (LM and FM) compared to baselines. Our methods successfully identify winning subsets of the ImageNet, yielding transfer accuracy on par or even better than scenarios without pruning.

\textbf{Fig.\,\ref{fig: main_lp}} 
presents the downstream accuracy of transfer learning vs. different pruning ratios. Here DP is performed using the surrogate model (RN-18) on ImageNet for $8$ downstream tasks. The source model (RN-101) is then trained on the pruned ImageNet and the transfer learning accuracy is assessed through LP (linear probing) and FF (full finetuning). We also present the downstream performance without pruning the source dataset (No Prune) as a reference for winning subsets.
As we can see, both LM and FM significantly outperform the baselines by a substantial margin. Notably, LM and FM consistently identify winning subsets with significantly larger pruning ratios in all settings. This highlights the effectiveness of our proposed methods in achieving substantial dataset pruning without hurting downstream performance.
Furthermore, we observe that the downstream performance of LM and FM initially improves and then declines as the pruning ratio increases. 
This is not surprising, since the initial increase in performance corresponds to the scenario where harmful source classes are removed, consistent with \cite{jain2022data}. 
When the source dataset continues to shrink, the performance inevitably decreases, as a natural consequence of reducing the size of the source dataset. 
Moreover, in some datasets of small sizes (\textit{e.g.}, OxfordPets), LP achieves performance comparable to FF. This is attributed to the fact that the performance gap between LP and FF using the large-scale RN-101 model on small-scale datasets tends to diminish.

\begin{table}[htb]
\centering
\caption{The pruning ratio of the \textbf{best winning subsets} obtained by each DP method for different downstream tasks. The other experiment setup is consistent with \textbf{Fig.\,\ref{fig: main_lp}}. The achieved largest pruning ratio   is highlighted in \textbf{bold} in each column. And N/A indicates the scenario where no winning subsets are found.}
\resizebox{0.8\textwidth}{!}{%
\begin{tabular}{c|cc|cc|cc|cc|cc|cc|cc|cc}
\toprule[1pt]

Dataset & \multicolumn{2}{c|}{OxfordPets} & \multicolumn{2}{c|}{SUN397} & \multicolumn{2}{c|}{Food101} & \multicolumn{2}{c|}{DTD} & \multicolumn{2}{c|}{StanfordCars} & \multicolumn{2}{c|}{UCF101} & \multicolumn{2}{c|}{CIFAR10} & \multicolumn{2}{c}{Flowers102} \\ 
Finetune method & LP & FF & LP & FF & LP & FF & LP & FF & LP & FF & LP & FF & LP & FF & LP & FF 
\\ \midrule
{\random}   & $20\%$ & $20\%$ & $15\%$ & N/A & N/A & $20\%$ & $10\%$ & $20\%$ & N/A & $10\%$ & $20\%$ & $10\%$ & N/A & $15\%$ & $40\%$ & $35\%$ \\
{\moderate} & $15\%$ & $20\%$ & $15\%$ & $20\%$ & N/A & $20\%$ & $15\%$ & $30\%$ & N/A & $5\%$ & $5\%$ & $10\%$ & N/A & $20\%$ & $40\%$ & $35\%$ \\
{\gradnorm} & $25\%$ & $20\%$ & $10\%$ & $20\%$ & $5\%$ & N/A & $15\%$ & $20\%$ & N/A & $5\%$ & $5\%$ & N/A & $40\%$ & $15\%$ & $45\%$ & $40\%$ \\
LM (ours)         & $80\%$ & $\textbf{80\%}$ & $\mathbf{65\%}$ & $\mathbf{70\%}$ & $\mathbf{65\%}$ & $70\%$ & $\mathbf{55\%}$ & $\mathbf{50\%}$ & $\mathbf{65\%}$ & $\mathbf{75\%}$ & $\mathbf{45\%}$ & $\mathbf{40\%}$ & $40\%$ & $40\%$ & $\mathbf{55\%}$ & $\mathbf{60\%}$ \\
FM  (ours)         & $\mathbf{85\%}$ & $75\%$ & $\mathbf{65\%}$ & $60\%$ & $55\%$ & $\mathbf{75\%}$ & $\mathbf{55\%}$ & $\mathbf{50\%}$ & $60\%$ & $70\%$ & $\mathbf{45\%}$ & $\mathbf{40\%}$ & $\mathbf{45\%}$ & $\mathbf{45\%}$ & $\mathbf{55\%}$ & $55\%$ \\
\bottomrule[1pt]
\end{tabular}
}
\label{tab: main_winning_ratio}
\vspace*{-4mm}
\end{table}
\textbf{Tab.\,\ref{tab: main_winning_ratio}} provides a summary of the pruning ratios achieved by the best winning subsets identified using different DP methods for all $8$ downstream datasets. Both LM and FM methods successfully remove more than $45\%$ of the source classes without downstream performance drop. 
In contrast, all the baselines experience significant performance degradation when the pruning ratio exceeds $40\%$.

\begin{table}[htb]
\vspace*{-5mm}
\centering
\caption{The downstream performance with different source data pruning ratios in the SSL pretraining setting. A randomly initialized RN-101 is self-supervised pretrained using \textsc{MoCo v2} on each full/pruned source dataset and finetuned on the downstream task through LP. The best result in each pruning ratio is marked in \textbf{bold} and the performance surpassing the unpruned setting (pruning ratio $0\%$) is highlighted in \colorbox{LightCyan}{cyan}. }
\label{tab: main_ssl_lp}
\resizebox{.9\textwidth}{!}{%
\begin{tabular}{c|ccccc|ccccc|ccccc}
\toprule[1pt]
\midrule
\multirow{2}{*}{\begin{tabular}[c]{@{}c@{}}Dataset\\ Pruning Ratio\end{tabular}} & \multicolumn{5}{c|}{OxfordPets}                         & \multicolumn{5}{c|}{SUN397}                                & \multicolumn{5}{c}{Flowers102}                        \\ 
                         & 0\%                    & 50\%  & 60\%  & 70\%  & 80\%  & 0\%                    & 50\%  & 60\%  & 70\%  & 80\%  & 0\%                    & 50\%  & 60\%  & 70\%  & 80\%  \\
\midrule
\random & \multirow{4}{*}{69.26} & 62.32 & 61.27 & 59.09 & 53.75 & \multirow{4}{*}{47.36} & 45.63 & 45.08 & 43.54 & 39.81    & \multirow{4}{*}{85.17} & 82.23 & 82.60 & 81.03 & 80.02 \\
\moderate &                        & 63.37 & 62.45 & 63.31 & 57.42 &                        & 45.73 & 45.14 & 44.23 & 40.82  &                        & 82.45 & 81.45 & 81.69 & 81.32 \\
\gradnorm &                        & 64.42 & 63.34 & 61.14 & 56.42 &                        & 45.72 & 45.58 & 45.24 & 41.72  &                        & 82.85 & 82.44 & 82.14 & 81.73 \\
FM (ours) &                        & \cellcolor{LightCyan}\textbf{69.92} & \cellcolor{LightCyan}\textbf{69.99} & \cellcolor{LightCyan}\textbf{70.29} & \cellcolor{LightCyan}\textbf{70.21} &                        & \cellcolor{LightCyan}\textbf{48.46} & \cellcolor{LightCyan}\textbf{48.58} & \cellcolor{LightCyan}\textbf{47.90} & \textbf{46.00}                        &                        & \cellcolor{LightCyan}\textbf{85.22} & \cellcolor{LightCyan}\textbf{85.42} & \textbf{84.37} & \textbf{84.61}  \\
\midrule
\bottomrule[1pt]
\end{tabular}%
}
\vspace*{-1mm}
\end{table}
{\textbf{Tab.\,\ref{tab: main_ssl_lp}}} highlights the effectiveness of FM-based DP in the self-supervised pretraining setup for three representative downstream tasks.
As we can see, the transfer learning accuracy achieved by using FM consistently outperforms baselines in the self-supervised pretraining paradigm. FM can identify winning subsets for transfer learning even in the challenging regime of large pruning ratios, ranging from $50\%$ to $80\%$. For instance, in the case of SUN397, FM-based winning subsets achieves a pruning ratio up to $70\%$, and for Flowers102 the maximum pruning ratio is $60\%$. These pruning merits align with the findings for FM in the supervised pretraining setup, as illustrated in Tab. \ref{tab: main_winning_ratio}.

\begin{wraptable}{r}{70mm}
\centering
\vspace*{-4mm}
\caption{\footnotesize{Time consumption of LM/FM in Fig.\ref{fig: main_lp} to obtain the pretrained model. 
The reported time consumption covers surrogate model (RN-18) training, LM/FM dataset pruning, and source model pretraining (RN-101).
}}
\label{tab: time_efficiency}
\vspace*{-3mm}
\resizebox{.99\linewidth}{!}{%
\begin{tabular}{c|ccccc}
\toprule[1pt]
Pruning Ratio & 0\% & 20\% & 40\% & 60\% & 80\% \\ \midrule
\begin{tabular}[c]{@{}c@{}}Time \\ Consumption 
(h)\end{tabular} & 5.4 & \begin{tabular}[c]{@{}c@{}}4.6\\ (15\%$\downarrow$)\end{tabular} & \begin{tabular}[c]{@{}c@{}}3.5\\ (35\%$\downarrow$)\end{tabular} & \begin{tabular}[c]{@{}c@{}}2.4\\ (56\%$\downarrow$)\end{tabular} & \begin{tabular}[c]{@{}c@{}}1.3\\ (76\%$\downarrow$)\end{tabular} \\
\toprule[1pt]
\end{tabular}%
}
\vspace*{-1.5em}
\end{wraptable}
\textbf{DP enhances the efficiency of source pretraining.} 
\textbf{Tab.\,\ref{tab: time_efficiency}} displays the computation time required to obtain the pretrained source model using LM at different pruning ratios. 
The reported time consumption includes the entire pipeline, encompassing surrogate model training (RN-18), DP process, and source model training (RN-101) on the pruned ImageNet dataset. 
\revision{The runtime cost of the conventional transfer learning on the full ImageNet dataset for RN-101 is also listed as a reference.}
\revision{As we can see, DP enjoys high efficiency merit of source training.}
Taking the $5.4$ hours required for source training on the full ImageNet dataset as a reference, LM-enabled $20\%$ pruned ImageNet achieves a $15\%$ reduction in training time. Moreover, the efficiency advantage increases to $76\%$ when the pruning ratio reaches $80\%$ and these computational benefits do not sacrifice transfer learning accuracy at all. 

\begin{wraptable}{r}{.5\linewidth}
\centering
\vspace*{-3mm}
\caption{\footnotesize{Time consumption comparison with a pruning ratio of $60\%$ of different dataset pruning methods. Other settings follow Fig.\,\ref{fig: main_lp}.
}}
\label{tab: time_baselines}
\vspace*{-3mm}
\resizebox{\linewidth}{!}{%
\begin{tabular}{c|cccc}
\toprule[1pt]
Method & \moderate & \gradnorm & \textsc{Brute-Force} \cite{jain2022data} & ours \\ \midrule
\begin{tabular}[c]{@{}c@{}}Time \\ Consumption 
(h)\end{tabular} & $7.6$ & $18.4$ & $>500$ & 2.4 \\
\toprule[1pt]
\end{tabular}%
}
\vspace*{-1.5em}
\end{wraptable}
Next, we compare the efficiency of our methods with the state-of-the-art dataset pruning baselines in Tab.\,\ref{tab: time_baselines}, including {\gradnorm}, {\moderate}, and the brute-force method proposed in \cite{jain2022data}.
\revision{We showcase the efficiency comparison with a pruning ratio of $60\%$. As we can see, our method (even using a surrogate model) achieves a substantial computation efficiency improvement over other methods. In particular, \cite{jain2022data} involves training thousands of source models. Thus, its computational cost is typically unaffortable in experiments.}

\begin{wraptable}{r}{.45\linewidth}
\vspace*{-4mm}
\centering
\caption{Downstream performance of models pretrained on full/pruned source dataset (\textsc{Dense}/\textsc{LM}) w/wo adversarial pretraining (Adv). For \textsc{Dense-Adv} and \textsc{LM-Adv}, $3$-step adversarial training \cite{madry2017towards} is used for pretraining. Pruning ratio, downstream test accuracy (Acc.), and time consumptions for obtaining pretrained models are reported.}
\vspace*{-2mm}
\label{tab: exp_adv}
\resizebox{\linewidth}{!}{%
\begin{tabular}{c|cccc|cccc}
\toprule
\multirow{3}{*}{Method} & \multicolumn{4}{c}{SUN397} & \multicolumn{4}{c}{DTD} \\
 & Pruning & \multicolumn{2}{c}{Acc.(\%)} & Time & Pruning &  \multicolumn{2}{c}{Acc.(\%)}  & Time \\
 & Ratio & LP & FF & (h) & Ratio & LP & FF & (h) \\ \midrule
\textsc{Dense} & N/A & 51.45 & 54.21 & 5.4 & N/A & 65.91 & 67.21 & 5.4 \\
\textsc{Dense-Adv} & N/A & 52.97 & 55.67 & 13.7 & N/A & 67.23 & 68.92 & 13.7 \\
\textsc{LM} & 70 & 50.95 & 54.28 & 1.9 & 50 & 66.25 & 67.22 & 2.9 \\
\textsc{LM-Adv} & 70 & 52.07 & 55.49 & 4.2 & 50 & 67.02 & 68.54 & 6.7 \\
\bottomrule
\end{tabular}%
}
\vspace*{-5mm}
\end{wraptable}
\textbf{DP enables efficient adversarial pretraining.} 
{\textbf{Tab.\,\ref{tab: exp_adv}}  showcases the improvement in transfer learning accuracy and efficiency achieved by our proposed LP-based DP method when incorporating \textit{adversarial training} (\textbf{AT}) \cite{madry2017towards} on either the full or pruned ImageNet dataset. 
This transfer learning setup is motivated by the findings in \cite{salman2020adversarially}, showing   that enhancing the robustness of the source model against adversarial attacks through AT can improve transfer learning accuracy in both LP and FF-based finetuning scenarios. We then report downstream accuracies using LP and FF on two specific downstream datasets: SUN397 and DTD, which are intentionally chosen due to the large room for improvement in transfer learning accuracy, as shown in  Fig.\,\ref{fig: main_lp}.  We also determine the pruning ratios for LM by selecting those that led to the best winning subsets. Our experimental results demonstrate that employing AT on both the unpruned and pruned source datasets can improve transfer learning accuracy. Specifically, we refer to AT on the original unpruned ImageNet dataset as \textsc{Dense-AT}, and AT on the LM-pruned ImageNet dataset as \textsc{LM-AT}. One notable advantage of integrating LM into AT is the significant improvement in computation efficiency. A key highlight of this approach is that \textsc{LM-AT} achieves a similar computation time as the standard source training on ImageNet (\textsc{Dense}), while exhibiting almost no accuracy drop compared to \textsc{Dense-AT}. This observation demonstrates the potential to accelerate   AT   through  DP. 
}  

\begin{wrapfigure}{r}{.5\linewidth}
\vspace*{0mm}
\centerline{
\begin{tabular}{cc}
    \hspace*{-2mm} \includegraphics[width=.24\textwidth,height=!]{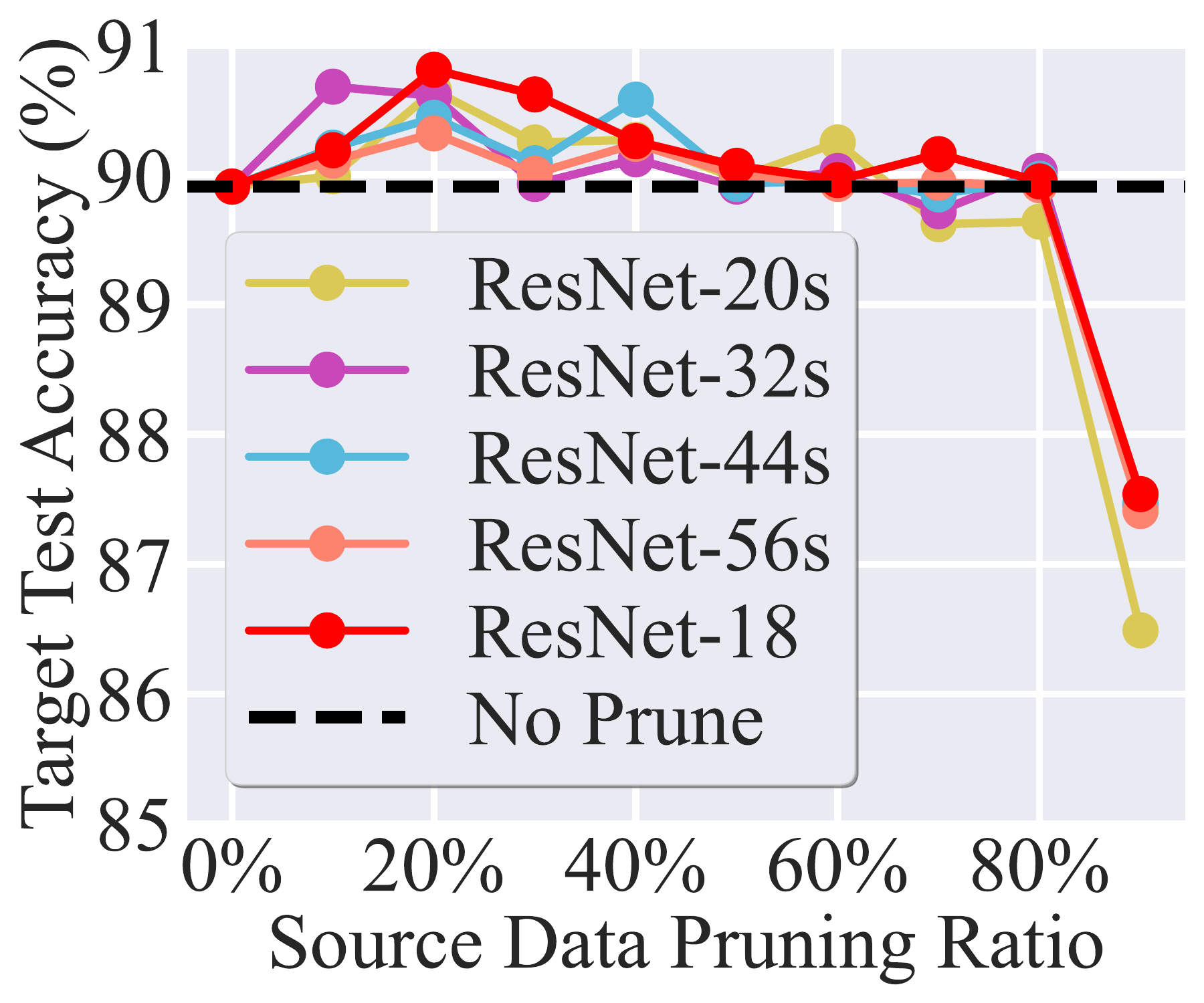} &
    \hspace*{-4mm}  \includegraphics[width=.24\textwidth,height=!]{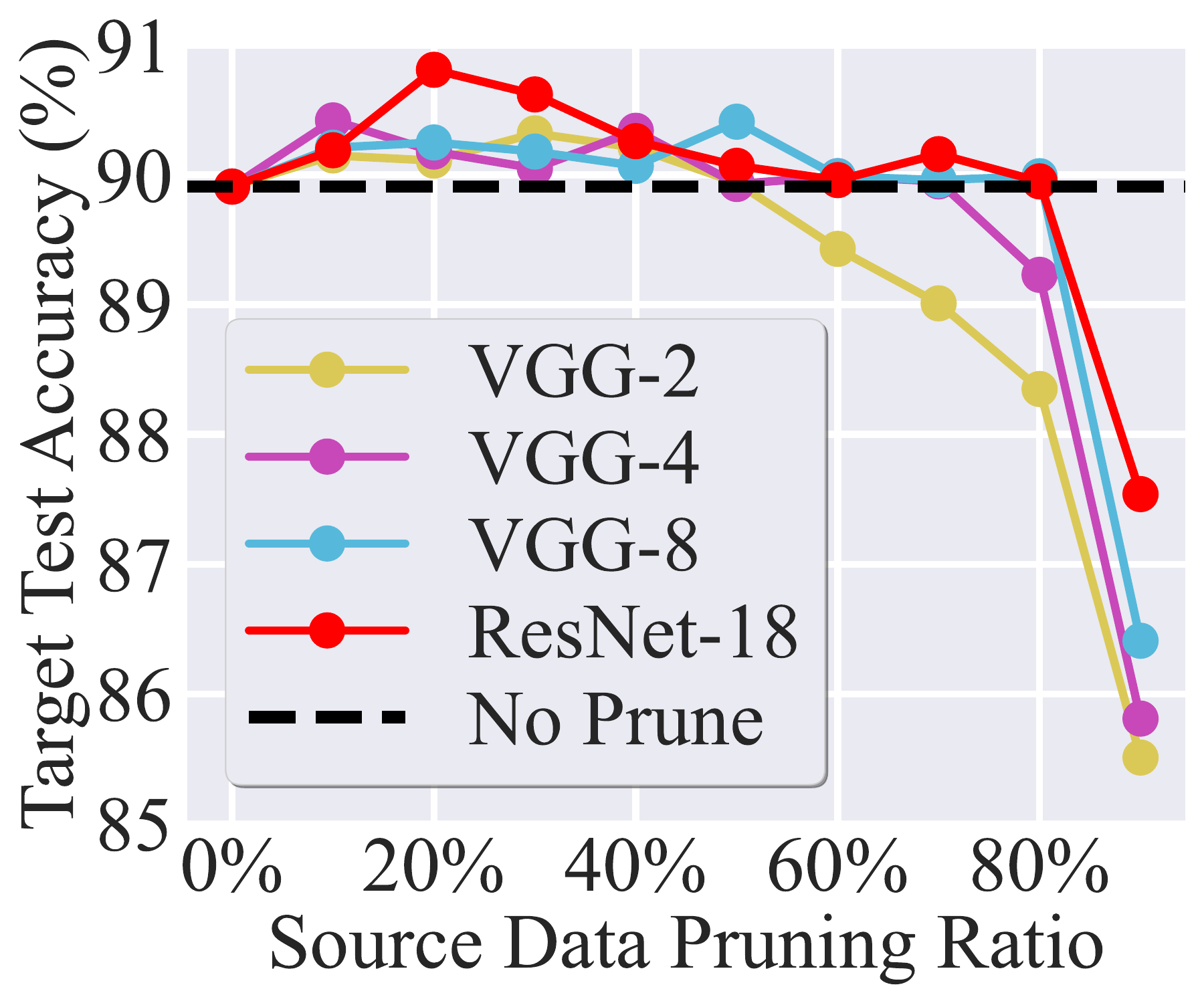} \\
    \hspace*{-2mm} \footnotesize{(a) ResNets family}
    & \hspace*{-4mm} \footnotesize{(b) VGG family}\\
\end{tabular}
}
\vspace*{-2mm}
\caption{\footnotesize 
Source dataset pruning trajectory given the downstream task {OxfordPets} using different surrogate models. Other settings follow Fig.\,\ref{fig: main_lp}.
}
\vspace*{-5mm}
\label{fig: surrogate_model_study}
\end{wrapfigure}

\textbf{\revision{Robustness against the surrogate model size.}} To explore the sensitivity of our proposed method to the surrogate model’s size and accuracy, we  
show the transfer learning performance on the task {OxfordPets} against different surrogate model sizes in \textbf{Fig.\,\ref{fig: surrogate_model_study}}. It is evident that even though the performance of the surrogate model on the source dataset ({ImageNet}) decreases, the downstream performance of RN-101 pretrained on the LM-based pruned source subsets remains relatively stable. 
Further, \textbf{Tab.\,\ref{tab: surrogate_model}} compares the class indices selected by the corresponding surrogate models. Interestingly, the most relevant source class selections exhibit a high level of agreement across surrogate models of differing sizes. 
LM shows robust behavior with different surrogate models, even if they yield different performance on the source dataset due to their different model capacities. 
For example, RN-32s only achieves a test accuracy of $40.77\%$ on the source dataset, but it manages to achieve the same maximum pruning ratio of the winning subsets as RN-18. This underscores that LM can accommodate a small surrogate model of $1\%$ the size of the pretrained model (Rn-32s vs. RN-101). Moreover, it reveals that the surrogate model does not need to attain exceptional performance in the source domain for dataset pruning in transfer learning.

\begin{table}[htb]
\centering
\vspace*{-4mm}
\caption{Downstream performance of different DP methods using ViT-B/16 as the source model and FF as the finetuning method. Other settings are consistent with Fig.\,\ref{fig: main_lp}. The best performance at each pruning ratio is marked in \textbf{bold}, and performance exceeding the unpruned setting ($0\%$ pruning ratio) is highlighted in \colorbox{LightCyan}{cyan}.
}
\label{tab: main_vit}
\resizebox{.9\textwidth}{!}{%
\begin{tabular}{c|ccccc|ccccc|ccccc}
\toprule[1pt]
\midrule
\multirow{2}{*}{\begin{tabular}[c]{@{}c@{}}Dataset\\ Pruning Ratio\end{tabular}} & \multicolumn{5}{c|}{OxfordPets}                         & \multicolumn{5}{c|}{SUN397} & \multicolumn{5}{c}{Flowers102} \\ 
& 0\%  & 20\%  & 40\%  & 60\%  & 80\%  
& 0\%  & 20\%  & 40\%  & 60\%  & 80\%   
& 0\%  & 20\%  & 40\%  & 60\%  & 80\%   \\
\midrule
\random 
& \multirow{4}{*}{85.75} & 85.63 & 83.72 & 75.94 & 69.83
& \multirow{4}{*}{43.97} & 43.74 & 41.18 & 36.72 & 29.41
& \multirow{4}{*}{90.95} & 90.25 & 89.38 & 87.47 & 84.44 \\
\moderate &  
& 85.39 & 83.97 & 82.39 & 78.32 
&                        
& 43.97 & 41.82 & 38.89 & 31.15  
&                        
& 90.86 & 89.77 & 88.34 & 86.69 \\
\gradnorm &  
& \cellcolor{LightCyan}85.77 & 83.88 & 82.35 & 79.12 
&                        
& 43.32 & 40.93 & 38.52 & 31.00  
&                        
& 90.75 & 89.89 & 88.54 & 85.39 \\
LM &                        & \cellcolor{LightCyan}\textbf{86.15} & \cellcolor{LightCyan}\textbf{85.72} & \textbf{84.77} & \textbf{80.64} &                        & \cellcolor{LightCyan}\textbf{44.24} & \textbf{42.79} & \textbf{39.59} & \textbf{32.35}                        &                        & \cellcolor{LightCyan}\textbf{91.19} & \textbf{89.97} & \textbf{89.07} & \textbf{87.23}  \\
\midrule
\bottomrule[1pt]
\end{tabular}%
}
\vspace*{-2mm}
\end{table}

\textbf{DP for ViT.} In \textbf{Tab.\,\ref{tab: main_vit}}, we present the consistent transfer learning improvement achieved by LM for Vision Transformer (ViT) pretraining. Specifically, we utilize ViT-B/16 as the source model, RN-18 as the surrogate model for DP, and use FF on the source model for the downstream tasks studied in Tab.\,\ref{tab: main_ssl_lp}.  As we can see, LM yields a better transfer learning accuracy than baselines at different pruning ratios, consistent with the previous results on RN-101. This observation highlights the model-agnostic advantage of our proposed method in transfer learning. We also observe the pruning ratio of the best winning subset achieved for ViT is smaller compared to ResNet. This discrepancy is possibly due to the higher data requirements of ViT during pretraining \cite{dosovitskiy2020image}.

\textbf{Additional results.} 
We conduct ablation studies on the cluster number for FM in Fig.\,\ref{fig: app_cluster_num_study} and the results of selecting data proposed by LM in reverse order in Fig.\,\ref{fig: app_reverse_order}. 
\revision{We shows FM can be applied to more advanced SSL framework in Tab.\,\ref{tab: mocov3}. We also investigate the performance of our proposed method in the multi-task setting in Fig.\,\ref{fig: multi_task_study}, and the scenarios with data biases in Tab.\,\ref{tab: cifar10c}. We further examine our method on the few-shot benchmark in Tab.\,\ref{tab: vtab}.}
We also present feature distribution analysis in Fig.\,\ref{fig: app_feature_dist} and image examples for the top classes selected by LM/FM in Fig.\,\ref{fig: app_ebe_all}.

\textbf{\revision{Limitations.}} Despite the remarkable empirical results above, we admit LM/FM mainly caters to exploring the source data influence in a specific downstream task, and the largest pruning ratio is reduced as more downstream tasks are considered ({see Fig.\,\ref{fig: multi_task_study}}), which we consider as a limitation. Meanwhile, we acknowledge that the promising performance achieved by our method still lacks rigorous theoretical analysis, despite the feature distribution and model flatness analysis, which we believe will be a challenging but exciting future work on DP for transfer learning.

\section{Conclusion}
\label{sec: conclusion}
In this paper, we first formally define and investigate the problem of DP (dataset pruning) for transfer learning. Recognizing the ineffectiveness of conventional DP methods in the transfer learning setting, we proposed two novel and efficient DP techniques, label mapping and feature mapping. Extensive experiments demonstrate that both methods can effectively prune a large ratio of source dataset and substantially reduce the pretraining computational costs without sacrificing the downstream performance in various transfer learning settings.

\newpage
\section*{Acknowledgement}
The work of Y. Zhang, Y. Zhang, A. Chen, J. Jia, J. Liu, and S. Liu was supported in part by the Cisco Research Award, the NSF Grant IIS-2207052, and the ARO Award W911NF2310343. The work of S. Chang was supported by the Cisco Research Award and the NSF Grant IIS-2207052. The work of M. Hong was supported by NSF grants CIF-1910385 and EPCN-2311007.

{{
\bibliographystyle{IEEEtranN}
\bibliography{refs}

\begin{thebibliography}{124}
\providecommand{\natexlab}[1]{#1}
\providecommand{\url}[1]{#1}
\csname url@samestyle\endcsname
\providecommand{\newblock}{\relax}
\providecommand{\bibinfo}[2]{#2}
\providecommand{\BIBentrySTDinterwordspacing}{\spaceskip=0pt\relax}
\providecommand{\BIBentryALTinterwordstretchfactor}{4}
\providecommand{\BIBentryALTinterwordspacing}{\spaceskip=\fontdimen2\font plus
\BIBentryALTinterwordstretchfactor\fontdimen3\font minus
  \fontdimen4\font\relax}
\providecommand{\BIBforeignlanguage}[2]{{%
\expandafter\ifx\csname l@#1\endcsname\relax
\typeout{** WARNING: IEEEtranN.bst: No hyphenation pattern has been}%
\typeout{** loaded for the language `#1'. Using the pattern for}%
\typeout{** the default language instead.}%
\else
\language=\csname l@#1\endcsname
\fi
#2}}
\providecommand{\BIBdecl}{\relax}
\BIBdecl

\bibitem[Zhao et~al.(2020)Zhao, Mopuri, and Bilen]{zhao2020dataset}
B.~Zhao, K.~R. Mopuri, and H.~Bilen, ``Dataset condensation with gradient
  matching,'' \emph{arXiv preprint arXiv:2006.05929}, 2020.

\bibitem[Zhao and Bilen(2021)]{zhao2021dataset}
B.~Zhao and H.~Bilen, ``Dataset condensation with differentiable siamese
  augmentation,'' in \emph{International Conference on Machine Learning}.\hskip
  1em plus 0.5em minus 0.4em\relax PMLR, 2021, pp. 12\,674--12\,685.

\bibitem[Xia et~al.(2023)Xia, Liu, Yu, Shen, Han, and Liu]{moderate2023xia}
X.~Xia, J.~Liu, J.~Yu, X.~Shen, B.~Han, and T.~Liu, ``Moderate coreset: A
  universal method of data selection for real-world data-efficient deep
  learning,'' in \emph{The Eleventh International Conference on Learning
  Representations}, 2023.

\bibitem[Geirhos et~al.(2018)Geirhos, Rubisch, Michaelis, Bethge, Wichmann, and
  Brendel]{geirhos2018imagenet}
R.~Geirhos, P.~Rubisch, C.~Michaelis, M.~Bethge, F.~A. Wichmann, and
  W.~Brendel, ``Imagenet-trained cnns are biased towards texture; increasing
  shape bias improves accuracy and robustness,'' \emph{arXiv preprint
  arXiv:1811.12231}, 2018.

\bibitem[Ilyas et~al.(2019)Ilyas, Santurkar, Tsipras, Engstrom, Tran, and
  Madry]{ilyas2019adversarial}
A.~Ilyas, S.~Santurkar, D.~Tsipras, L.~Engstrom, B.~Tran, and A.~Madry,
  ``Adversarial examples are not bugs, they are features,'' \emph{Advances in
  neural information processing systems}, vol.~32, 2019.

\bibitem[Gu et~al.(2017)Gu, Dolan-Gavitt, and Garg]{gu2017badnets}
T.~Gu, B.~Dolan-Gavitt, and S.~Garg, ``Badnets: Identifying vulnerabilities in
  the machine learning model supply chain,'' \emph{arXiv preprint
  arXiv:1708.06733}, 2017.

\bibitem[Wei et~al.(2020)Wei, Feng, Chen, and An]{wei2020combating}
H.~Wei, L.~Feng, X.~Chen, and B.~An, ``Combating noisy labels by agreement: A
  joint training method with co-regularization,'' in \emph{Proceedings of the
  IEEE/CVF conference on computer vision and pattern recognition}, 2020, pp.
  13\,726--13\,735.

\bibitem[Jiang et~al.(2018)Jiang, Zhou, Leung, Li, and
  Fei-Fei]{jiang2018mentornet}
L.~Jiang, Z.~Zhou, T.~Leung, L.-J. Li, and L.~Fei-Fei, ``Mentornet: Learning
  data-driven curriculum for very deep neural networks on corrupted labels,''
  in \emph{International conference on machine learning}.\hskip 1em plus 0.5em
  minus 0.4em\relax PMLR, 2018, pp. 2304--2313.

\bibitem[Guo et~al.(2022)Guo, Zhao, and Bai]{guo2022deepcore}
C.~Guo, B.~Zhao, and Y.~Bai, ``Deepcore: A comprehensive library for coreset
  selection in deep learning,'' in \emph{Database and Expert Systems
  Applications: 33rd International Conference, DEXA 2022, Vienna, Austria,
  August 22--24, 2022, Proceedings, Part I}.\hskip 1em plus 0.5em minus
  0.4em\relax Springer, 2022, pp. 181--195.

\bibitem[Borsos et~al.(2020)Borsos, Mutny, and Krause]{borsos2020coresets}
Z.~Borsos, M.~Mutny, and A.~Krause, ``Coresets via bilevel optimization for
  continual learning and streaming,'' \emph{Advances in Neural Information
  Processing Systems}, vol.~33, pp. 14\,879--14\,890, 2020.

\bibitem[Toneva et~al.(2018)Toneva, Sordoni, Combes, Trischler, Bengio, and
  Gordon]{toneva2018empirical}
M.~Toneva, A.~Sordoni, R.~T.~d. Combes, A.~Trischler, Y.~Bengio, and G.~J.
  Gordon, ``An empirical study of example forgetting during deep neural network
  learning,'' \emph{arXiv preprint arXiv:1812.05159}, 2018.

\bibitem[Castro et~al.(2018)Castro, Mar{\'\i}n-Jim{\'e}nez, Guil, Schmid, and
  Alahari]{castro2018end}
F.~M. Castro, M.~J. Mar{\'\i}n-Jim{\'e}nez, N.~Guil, C.~Schmid, and K.~Alahari,
  ``End-to-end incremental learning,'' in \emph{Proceedings of the European
  conference on computer vision (ECCV)}, 2018, pp. 233--248.

\bibitem[Aljundi et~al.(2019)Aljundi, Lin, Goujaud, and
  Bengio]{aljundi2019gradient}
R.~Aljundi, M.~Lin, B.~Goujaud, and Y.~Bengio, ``Gradient based sample
  selection for online continual learning,'' \emph{Advances in neural
  information processing systems}, vol.~32, 2019.

\bibitem[Sener and Savarese(2017)]{sener2017active}
O.~Sener and S.~Savarese, ``Active learning for convolutional neural networks:
  A core-set approach,'' \emph{arXiv preprint arXiv:1708.00489}, 2017.

\bibitem[Chatterjee and Hadi(1986)]{chatterjee1986influential}
S.~Chatterjee and A.~S. Hadi, ``Influential observations, high leverage points,
  and outliers in linear regression,'' \emph{Statistical science}, pp.
  379--393, 1986.

\bibitem[Cook(1986)]{cook1986assessment}
R.~D. Cook, ``Assessment of local influence,'' \emph{Journal of the Royal
  Statistical Society: Series B (Methodological)}, vol.~48, no.~2, pp.
  133--155, 1986.

\bibitem[Thomas and Cook(1990)]{thomas1990assessing}
W.~Thomas and R.~D. Cook, ``Assessing influence on predictions from generalized
  linear models,'' \emph{Technometrics}, vol.~32, no.~1, pp. 59--65, 1990.

\bibitem[Wei et~al.(1998)Wei, Hu, and Fung]{wei1998generalized}
B.-C. Wei, Y.-Q. Hu, and W.-K. Fung, ``Generalized leverage and its
  applications,'' \emph{Scandinavian Journal of statistics}, vol.~25, no.~1,
  pp. 25--37, 1998.

\bibitem[Koh and Liang(2017)]{koh2017understanding}
P.~W. Koh and P.~Liang, ``Understanding black-box predictions via influence
  functions,'' in \emph{International conference on machine learning}.\hskip
  1em plus 0.5em minus 0.4em\relax PMLR, 2017, pp. 1885--1894.

\bibitem[Schioppa et~al.(2022)Schioppa, Zablotskaia, Vilar, and
  Sokolov]{schioppa2022scaling}
A.~Schioppa, P.~Zablotskaia, D.~Vilar, and A.~Sokolov, ``Scaling up influence
  functions,'' in \emph{Proceedings of the AAAI Conference on Artificial
  Intelligence}, vol.~36, 2022, pp. 8179--8186.

\bibitem[Guo et~al.(2020{\natexlab{a}})Guo, Rajani, Hase, Bansal, and
  Xiong]{guo2020fastif}
H.~Guo, N.~F. Rajani, P.~Hase, M.~Bansal, and C.~Xiong, ``Fastif: Scalable
  influence functions for efficient model interpretation and debugging,''
  \emph{arXiv preprint arXiv:2012.15781}, 2020.

\bibitem[Yang et~al.(2022)Yang, Xie, Peng, Xu, Sun, and Li]{yang2022dataset}
S.~Yang, Z.~Xie, H.~Peng, M.~Xu, M.~Sun, and P.~Li, ``Dataset pruning: Reducing
  training data by examining generalization influence,'' \emph{arXiv preprint
  arXiv:2205.09329}, 2022.

\bibitem[Kong et~al.(2022)Kong, Shen, and Huang]{kong2022resolving}
\BIBentryALTinterwordspacing
S.~Kong, Y.~Shen, and L.~Huang, ``Resolving training biases via influence-based
  data relabeling,'' in \emph{International Conference on Learning
  Representations}, 2022. [Online]. Available:
  \url{https://openreview.net/forum?id=EskfH0bwNVn}
\BIBentrySTDinterwordspacing

\bibitem[Pleiss et~al.(2020)Pleiss, Zhang, Elenberg, and
  Weinberger]{pleiss2020identifying}
G.~Pleiss, T.~Zhang, E.~Elenberg, and K.~Q. Weinberger, ``Identifying
  mislabeled data using the area under the margin ranking,'' \emph{Advances in
  Neural Information Processing Systems}, vol.~33, pp. 17\,044--17\,056, 2020.

\bibitem[Welling(2009)]{welling2009herding}
M.~Welling, ``Herding dynamical weights to learn,'' in \emph{Proceedings of the
  26th Annual International Conference on Machine Learning}, 2009, pp.
  1121--1128.

\bibitem[Paul et~al.(2021)Paul, Ganguli, and Dziugaite]{paul2021deep}
M.~Paul, S.~Ganguli, and G.~K. Dziugaite, ``Deep learning on a data diet:
  Finding important examples early in training,'' \emph{Advances in Neural
  Information Processing Systems}, vol.~34, pp. 20\,596--20\,607, 2021.

\bibitem[Pruthi et~al.(2020)Pruthi, Liu, Kale, and
  Sundararajan]{pruthi2020estimating}
G.~Pruthi, F.~Liu, S.~Kale, and M.~Sundararajan, ``Estimating training data
  influence by tracing gradient descent,'' \emph{Advances in Neural Information
  Processing Systems}, vol.~33, pp. 19\,920--19\,930, 2020.

\bibitem[Rebuffi et~al.(2017)Rebuffi, Kolesnikov, Sperl, and
  Lampert]{rebuffi2017icarl}
S.-A. Rebuffi, A.~Kolesnikov, G.~Sperl, and C.~H. Lampert, ``icarl: Incremental
  classifier and representation learning,'' in \emph{Proceedings of the IEEE
  conference on Computer Vision and Pattern Recognition}, 2017, pp. 2001--2010.

\bibitem[Feldman and Langberg(2011)]{feldman2011unified}
D.~Feldman and M.~Langberg, ``A unified framework for approximating and
  clustering data,'' in \emph{Proceedings of the forty-third annual ACM
  symposium on Theory of computing}, 2011, pp. 569--578.

\bibitem[Ju et~al.(2022)Ju, Jung, Oh, and Kim]{ju2022extending}
J.~Ju, H.~Jung, Y.~Oh, and J.~Kim, ``Extending contrastive learning to
  unsupervised coreset selection,'' \emph{IEEE Access}, vol.~10, pp.
  7704--7715, 2022.

\bibitem[Huggins et~al.(2016)Huggins, Campbell, and
  Broderick]{huggins2016coresets}
J.~Huggins, T.~Campbell, and T.~Broderick, ``Coresets for scalable bayesian
  logistic regression,'' \emph{Advances in Neural Information Processing
  Systems}, vol.~29, 2016.

\bibitem[Campbell and Broderick(2019)]{campbell2019automated}
T.~Campbell and T.~Broderick, ``Automated scalable bayesian inference via
  hilbert coresets,'' \emph{The Journal of Machine Learning Research}, vol.~20,
  no.~1, pp. 551--588, 2019.

\bibitem[Nguyen et~al.(2017)Nguyen, Li, Bui, and Turner]{nguyen2017variational}
C.~V. Nguyen, Y.~Li, T.~D. Bui, and R.~E. Turner, ``Variational continual
  learning,'' \emph{arXiv preprint arXiv:1710.10628}, 2017.

\bibitem[Farquhar and Gal(2018)]{farquhar2018towards}
S.~Farquhar and Y.~Gal, ``Towards robust evaluations of continual learning,''
  \emph{arXiv preprint arXiv:1805.09733}, 2018.

\bibitem[Kim et~al.(2023)Kim, Bae, and Yun]{kim2023coreset}
S.~Kim, S.~Bae, and S.-Y. Yun, ``Coreset sampling from open-set for
  fine-grained self-supervised learning,'' in \emph{Proceedings of the IEEE/CVF
  Conference on Computer Vision and Pattern Recognition}, 2023, pp. 7537--7547.

\bibitem[Yang et~al.(2023)Yang, Kang, and Mirzasoleiman]{yang2023towards}
Y.~Yang, H.~Kang, and B.~Mirzasoleiman, ``Towards sustainable learning:
  Coresets for data-efficient deep learning,'' \emph{arXiv preprint
  arXiv:2306.01244}, 2023.

\bibitem[Pan and Yang(2010)]{pan2010survey}
S.~J. Pan and Q.~Yang, ``A survey on transfer learning,'' \emph{IEEE
  Transactions on knowledge and data engineering}, vol.~22, no.~10, pp.
  1345--1359, 2010.

\bibitem[Torrey and Shavlik(2010)]{torrey2010transfer}
L.~Torrey and J.~Shavlik, ``Transfer learning,'' in \emph{Handbook of research
  on machine learning applications and trends: algorithms, methods, and
  techniques}.\hskip 1em plus 0.5em minus 0.4em\relax IGI global, 2010, pp.
  242--264.

\bibitem[Yang et~al.(2013)Yang, Hanneke, and Carbonell]{yang2013theory}
L.~Yang, S.~Hanneke, and J.~Carbonell, ``A theory of transfer learning with
  applications to active learning,'' \emph{Machine learning}, vol.~90, pp.
  161--189, 2013.

\bibitem[Sorscher et~al.(2022)Sorscher, Geirhos, Shekhar, Ganguli, and
  Morcos]{sorscher2022beyond}
\BIBentryALTinterwordspacing
B.~Sorscher, R.~Geirhos, S.~Shekhar, S.~Ganguli, and A.~S. Morcos, ``Beyond
  neural scaling laws: beating power law scaling via data pruning,'' in
  \emph{Advances in Neural Information Processing Systems}, A.~H. Oh,
  A.~Agarwal, D.~Belgrave, and K.~Cho, Eds., 2022. [Online]. Available:
  \url{https://openreview.net/forum?id=UmvSlP-PyV}
\BIBentrySTDinterwordspacing

\bibitem[Jain et~al.(2022)Jain, Salman, Khaddaj, Wong, Park, and
  Madry]{jain2022data}
S.~Jain, H.~Salman, A.~Khaddaj, E.~Wong, S.~M. Park, and A.~Madry, ``A
  data-based perspective on transfer learning,'' \emph{arXiv preprint
  arXiv:2207.05739}, 2022.

\bibitem[Sattigeri et~al.(2022)Sattigeri, Ghosh, Padhi, Dognin, and
  Varshney]{sattigeri2022fair}
P.~Sattigeri, S.~Ghosh, I.~Padhi, P.~Dognin, and K.~R. Varshney, ``Fair
  infinitesimal jackknife: Mitigating the influence of biased training data
  points without refitting,'' \emph{arXiv preprint arXiv:2212.06803}, 2022.

\bibitem[Chen et~al.(2020{\natexlab{a}})Chen, Kornblith, Norouzi, and
  Hinton]{chen2020simple}
T.~Chen, S.~Kornblith, M.~Norouzi, and G.~Hinton, ``A simple framework for
  contrastive learning of visual representations,'' in \emph{International
  conference on machine learning}.\hskip 1em plus 0.5em minus 0.4em\relax PMLR,
  2020, pp. 1597--1607.

\bibitem[He et~al.(2020)He, Fan, Wu, Xie, and Girshick]{he2020momentum}
K.~He, H.~Fan, Y.~Wu, S.~Xie, and R.~Girshick, ``Momentum contrast for
  unsupervised visual representation learning,'' in \emph{Proceedings of the
  IEEE/CVF conference on computer vision and pattern recognition}, 2020, pp.
  9729--9738.

\bibitem[Chen et~al.(2020{\natexlab{b}})Chen, Fan, Girshick, and
  He]{chen2020improved}
X.~Chen, H.~Fan, R.~Girshick, and K.~He, ``Improved baselines with momentum
  contrastive learning,'' \emph{arXiv preprint arXiv:2003.04297}, 2020.

\bibitem[Chen et~al.(2020{\natexlab{c}})Chen, Kornblith, Swersky, Norouzi, and
  Hinton]{chen2020big}
T.~Chen, S.~Kornblith, K.~Swersky, M.~Norouzi, and G.~E. Hinton, ``Big
  self-supervised models are strong semi-supervised learners,'' \emph{Advances
  in neural information processing systems}, vol.~33, pp. 22\,243--22\,255,
  2020.

\bibitem[Wu et~al.(2018)Wu, Xiong, Yu, and Lin]{wu2018unsupervised}
Z.~Wu, Y.~Xiong, S.~X. Yu, and D.~Lin, ``Unsupervised feature learning via
  non-parametric instance discrimination,'' in \emph{Proceedings of the IEEE
  conference on computer vision and pattern recognition}, 2018, pp. 3733--3742.

\bibitem[Tian et~al.(2020)Tian, Sun, Poole, Krishnan, Schmid, and
  Isola]{tian2020makes}
Y.~Tian, C.~Sun, B.~Poole, D.~Krishnan, C.~Schmid, and P.~Isola, ``What makes
  for good views for contrastive learning?'' \emph{Advances in neural
  information processing systems}, vol.~33, pp. 6827--6839, 2020.

\bibitem[Grill et~al.(2020)Grill, Strub, Altch{\'e}, Tallec, Richemond,
  Buchatskaya, Doersch, Avila~Pires, Guo, Gheshlaghi~Azar,
  et~al.]{grill2020bootstrap}
J.-B. Grill, F.~Strub, F.~Altch{\'e}, C.~Tallec, P.~Richemond, E.~Buchatskaya,
  C.~Doersch, B.~Avila~Pires, Z.~Guo, M.~Gheshlaghi~Azar \emph{et~al.},
  ``Bootstrap your own latent-a new approach to self-supervised learning,''
  \emph{Advances in neural information processing systems}, vol.~33, pp.
  21\,271--21\,284, 2020.

\bibitem[Asano et~al.(2019)Asano, Rupprecht, and Vedaldi]{asano2019self}
Y.~M. Asano, C.~Rupprecht, and A.~Vedaldi, ``Self-labelling via simultaneous
  clustering and representation learning,'' \emph{arXiv preprint
  arXiv:1911.05371}, 2019.

\bibitem[Caron et~al.(2020)Caron, Misra, Mairal, Goyal, Bojanowski, and
  Joulin]{caron2020unsupervised}
M.~Caron, I.~Misra, J.~Mairal, P.~Goyal, P.~Bojanowski, and A.~Joulin,
  ``Unsupervised learning of visual features by contrasting cluster
  assignments,'' \emph{Advances in neural information processing systems},
  vol.~33, pp. 9912--9924, 2020.

\bibitem[Li et~al.(2020)Li, Zhou, Xiong, and Hoi]{li2020prototypical}
J.~Li, P.~Zhou, C.~Xiong, and S.~C. Hoi, ``Prototypical contrastive learning of
  unsupervised representations,'' \emph{arXiv preprint arXiv:2005.04966}, 2020.

\bibitem[Feldman and Zhang(2020)]{feldman2020neural}
V.~Feldman and C.~Zhang, ``What neural networks memorize and why: Discovering
  the long tail via influence estimation,'' \emph{Advances in Neural
  Information Processing Systems}, vol.~33, pp. 2881--2891, 2020.

\bibitem[He et~al.(2016)He, Zhang, Ren, and Sun]{he2016deep}
K.~He, X.~Zhang, S.~Ren, and J.~Sun, ``Deep residual learning for image
  recognition,'' in \emph{Proceedings of the IEEE conference on computer vision
  and pattern recognition}, 2016, pp. 770--778.

\bibitem[Russakovsky et~al.(2015)Russakovsky, Deng, Su, Krause, Satheesh, Ma,
  Huang, Karpathy, Khosla, Bernstein, et~al.]{russakovsky2015imagenet}
O.~Russakovsky, J.~Deng, H.~Su, J.~Krause, S.~Satheesh, S.~Ma, Z.~Huang,
  A.~Karpathy, A.~Khosla, M.~Bernstein \emph{et~al.}, ``Imagenet large scale
  visual recognition challenge,'' \emph{International journal of computer
  vision}, vol. 115, pp. 211--252, 2015.

\bibitem[Parkhi et~al.(2012)Parkhi, Vedaldi, Zisserman, and
  Jawahar]{parkhi2012cats}
O.~M. Parkhi, A.~Vedaldi, A.~Zisserman, and C.~Jawahar, ``Cats and dogs,'' in
  \emph{2012 IEEE conference on computer vision and pattern recognition}.\hskip
  1em plus 0.5em minus 0.4em\relax IEEE, 2012, pp. 3498--3505.

\bibitem[Krause et~al.(2013)Krause, Stark, Deng, and Fei-Fei]{krause20133d}
J.~Krause, M.~Stark, J.~Deng, and L.~Fei-Fei, ``3d object representations for
  fine-grained categorization,'' in \emph{Proceedings of the IEEE international
  conference on computer vision workshops}, 2013, pp. 554--561.

\bibitem[Salman et~al.(2020)Salman, Ilyas, Engstrom, Kapoor, and
  Madry]{salman2020adversarially}
H.~Salman, A.~Ilyas, L.~Engstrom, A.~Kapoor, and A.~Madry, ``Do adversarially
  robust imagenet models transfer better?'' \emph{Advances in Neural
  Information Processing Systems}, vol.~33, pp. 3533--3545, 2020.

\bibitem[Agarwal et~al.(2004)Agarwal, Har-Peled, and
  Varadarajan]{agarwal2004approximating}
P.~K. Agarwal, S.~Har-Peled, and K.~R. Varadarajan, ``Approximating extent
  measures of points,'' \emph{Journal of the ACM (JACM)}, vol.~51, no.~4, pp.
  606--635, 2004.

\bibitem[Har-Peled and Mazumdar(2004)]{har2004coresets}
S.~Har-Peled and S.~Mazumdar, ``On coresets for k-means and k-median
  clustering,'' in \emph{Proceedings of the thirty-sixth annual ACM symposium
  on Theory of computing}, 2004, pp. 291--300.

\bibitem[Feldman et~al.(2020)Feldman, Schmidt, and Sohler]{feldman2020turning}
D.~Feldman, M.~Schmidt, and C.~Sohler, ``Turning big data into tiny data:
  Constant-size coresets for k-means, pca, and projective clustering,''
  \emph{SIAM Journal on Computing}, vol.~49, no.~3, pp. 601--657, 2020.

\bibitem[Basu et~al.(2020)Basu, Pope, and Feizi]{basu2020influence}
S.~Basu, P.~Pope, and S.~Feizi, ``Influence functions in deep learning are
  fragile,'' \emph{arXiv preprint arXiv:2006.14651}, 2020.

\bibitem[Jia et~al.(2023)Jia, Zhang, Song, Liu, and Hero]{jia2023robustness}
J.~Jia, Y.~Zhang, D.~Song, S.~Liu, and A.~Hero, ``Robustness-preserving
  lifelong learning via dataset condensation,'' in \emph{ICASSP 2023-2023 IEEE
  International Conference on Acoustics, Speech and Signal Processing
  (ICASSP)}.\hskip 1em plus 0.5em minus 0.4em\relax IEEE, 2023, pp. 1--5.

\bibitem[Xie et~al.(2023)Xie, Lu, Yan, Yang, Tomizuka, and Zhan]{xie2023active}
Y.~Xie, H.~Lu, J.~Yan, X.~Yang, M.~Tomizuka, and W.~Zhan, ``Active finetuning:
  Exploiting annotation budget in the pretraining-finetuning paradigm,''
  \emph{arXiv preprint arXiv:2303.14382}, 2023.

\bibitem[Ghorbani and Zou(2019)]{ghorbani2019data}
A.~Ghorbani and J.~Zou, ``Data shapley: Equitable valuation of data for machine
  learning,'' in \emph{International Conference on Machine Learning}.\hskip 1em
  plus 0.5em minus 0.4em\relax PMLR, 2019, pp. 2242--2251.

\bibitem[Jia et~al.(2019{\natexlab{a}})Jia, Dao, Wang, Hubis, Hynes, G{\"u}rel,
  Li, Zhang, Song, and Spanos]{jia2019towards}
R.~Jia, D.~Dao, B.~Wang, F.~A. Hubis, N.~Hynes, N.~M. G{\"u}rel, B.~Li,
  C.~Zhang, D.~Song, and C.~J. Spanos, ``Towards efficient data valuation based
  on the shapley value,'' in \emph{The 22nd International Conference on
  Artificial Intelligence and Statistics}.\hskip 1em plus 0.5em minus
  0.4em\relax PMLR, 2019, pp. 1167--1176.

\bibitem[Jia et~al.(2019{\natexlab{b}})Jia, Dao, Wang, Hubis, Gurel, Li, Zhang,
  Spanos, and Song]{jia2019efficient}
R.~Jia, D.~Dao, B.~Wang, F.~A. Hubis, N.~M. Gurel, B.~Li, C.~Zhang, C.~J.
  Spanos, and D.~Song, ``Efficient task-specific data valuation for nearest
  neighbor algorithms,'' \emph{arXiv preprint arXiv:1908.08619}, 2019.

\bibitem[Lin et~al.(2022)Lin, Zhang, L{\'e}cuyer, Li, Panda, and
  Sen]{lin2022measuring}
J.~Lin, A.~Zhang, M.~L{\'e}cuyer, J.~Li, A.~Panda, and S.~Sen, ``Measuring the
  effect of training data on deep learning predictions via randomized
  experiments,'' in \emph{International Conference on Machine Learning}.\hskip
  1em plus 0.5em minus 0.4em\relax PMLR, 2022, pp. 13\,468--13\,504.

\bibitem[Guu et~al.(2023)Guu, Webson, Pavlick, Dixon, Tenney, and
  Bolukbasi]{guu2023simfluence}
K.~Guu, A.~Webson, E.~Pavlick, L.~Dixon, I.~Tenney, and T.~Bolukbasi,
  ``Simfluence: Modeling the influence of individual training examples by
  simulating training runs,'' \emph{arXiv preprint arXiv:2303.08114}, 2023.

\bibitem[Hammoudeh and Lowd(2022)]{hammoudeh2022identifying}
Z.~Hammoudeh and D.~Lowd, ``Identifying a training-set attack's target using
  renormalized influence estimation,'' in \emph{Proceedings of the 2022 ACM
  SIGSAC Conference on Computer and Communications Security}, 2022, pp.
  1367--1381.

\bibitem[Park et~al.(2023)Park, Georgiev, Ilyas, Leclerc, and
  Madry]{park2023trak}
S.~M. Park, K.~Georgiev, A.~Ilyas, G.~Leclerc, and A.~Madry, ``Trak:
  Attributing model behavior at scale,'' \emph{arXiv preprint
  arXiv:2303.14186}, 2023.

\bibitem[Just et~al.(2023)Just, Kang, Wang, Zeng, Ko, Jin, and
  Jia]{just2023lava}
H.~A. Just, F.~Kang, J.~T. Wang, Y.~Zeng, M.~Ko, M.~Jin, and R.~Jia, ``Lava:
  Data valuation without pre-specified learning algorithms,'' \emph{arXiv
  preprint arXiv:2305.00054}, 2023.

\bibitem[Ilyas et~al.(2022)Ilyas, Park, Engstrom, Leclerc, and
  Madry]{ilyas2022datamodels}
A.~Ilyas, S.~M. Park, L.~Engstrom, G.~Leclerc, and A.~Madry, ``Datamodels:
  Predicting predictions from training data,'' \emph{arXiv preprint
  arXiv:2202.00622}, 2022.

\bibitem[Yeh et~al.(2018)Yeh, Kim, Yen, and Ravikumar]{yeh2018representer}
C.-K. Yeh, J.~Kim, I.~E.-H. Yen, and P.~K. Ravikumar, ``Representer point
  selection for explaining deep neural networks,'' \emph{Advances in neural
  information processing systems}, vol.~31, 2018.

\bibitem[Song and Ermon(2019)]{song2019generative}
Y.~Song and S.~Ermon, ``Generative modeling by estimating gradients of the data
  distribution,'' \emph{Advances in neural information processing systems},
  vol.~32, 2019.

\bibitem[Deng et~al.(2021)Deng, Zhang, Vodrahalli, Kawaguchi, and
  Zou]{deng2021adversarial}
Z.~Deng, L.~Zhang, K.~Vodrahalli, K.~Kawaguchi, and J.~Y. Zou, ``Adversarial
  training helps transfer learning via better representations,'' \emph{Advances
  in Neural Information Processing Systems}, vol.~34, pp. 25\,179--25\,191,
  2021.

\bibitem[Liu et~al.(2022)Liu, Xie, Li, and Ma]{liu2022same}
H.~Liu, S.~M. Xie, Z.~Li, and T.~Ma, ``Same pre-training loss, better
  downstream: Implicit bias matters for language models,'' \emph{arXiv preprint
  arXiv:2210.14199}, 2022.

\bibitem[Liu et~al.(2019)Liu, Long, Wang, and Jordan]{liu2019towards}
H.~Liu, M.~Long, J.~Wang, and M.~I. Jordan, ``Towards understanding the
  transferability of deep representations,'' \emph{arXiv preprint
  arXiv:1909.12031}, 2019.

\bibitem[Rosenstein et~al.(2005)Rosenstein, Marx, Kaelbling, and
  Dietterich]{rosenstein2005transfer}
M.~T. Rosenstein, Z.~Marx, L.~P. Kaelbling, and T.~G. Dietterich, ``To transfer
  or not to transfer,'' in \emph{NIPS 2005 workshop on transfer learning}, vol.
  898, no.~3, 2005.

\bibitem[Duan et~al.(2012)Duan, Xu, and Chang]{duan2012exploiting}
L.~Duan, D.~Xu, and S.-F. Chang, ``Exploiting web images for event recognition
  in consumer videos: A multiple source domain adaptation approach,'' in
  \emph{2012 IEEE Conference on computer vision and pattern recognition}.\hskip
  1em plus 0.5em minus 0.4em\relax IEEE, 2012, pp. 1338--1345.

\bibitem[Ge et~al.(2014)Ge, Gao, Ngo, Li, and Zhang]{ge2014handling}
L.~Ge, J.~Gao, H.~Ngo, K.~Li, and A.~Zhang, ``On handling negative transfer and
  imbalanced distributions in multiple source transfer learning,''
  \emph{Statistical Analysis and Data Mining: The ASA Data Science Journal},
  vol.~7, no.~4, pp. 254--271, 2014.

\bibitem[Yosinski et~al.(2014)Yosinski, Clune, Bengio, and
  Lipson]{yosinski2014transferable}
J.~Yosinski, J.~Clune, Y.~Bengio, and H.~Lipson, ``How transferable are
  features in deep neural networks?'' \emph{Advances in neural information
  processing systems}, vol.~27, 2014.

\bibitem[Cao et~al.(2018)Cao, Long, Wang, and Jordan]{cao2018partial}
Z.~Cao, M.~Long, J.~Wang, and M.~I. Jordan, ``Partial transfer learning with
  selective adversarial networks,'' in \emph{Proceedings of the IEEE conference
  on computer vision and pattern recognition}, 2018, pp. 2724--2732.

\bibitem[Wang et~al.(2019)Wang, Dai, P{\'o}czos, and
  Carbonell]{wang2019characterizing}
Z.~Wang, Z.~Dai, B.~P{\'o}czos, and J.~Carbonell, ``Characterizing and avoiding
  negative transfer,'' in \emph{Proceedings of the IEEE/CVF conference on
  computer vision and pattern recognition}, 2019, pp. 11\,293--11\,302.

\bibitem[Guo et~al.(2020{\natexlab{b}})Guo, Rush, and Kim]{guo2020parameter}
D.~Guo, A.~M. Rush, and Y.~Kim, ``Parameter-efficient transfer learning with
  diff pruning,'' \emph{arXiv preprint arXiv:2012.07463}, 2020.

\bibitem[Liu et~al.(2021)Liu, Cai, Guo, and Chen]{liu2021transtailor}
B.~Liu, Y.~Cai, Y.~Guo, and X.~Chen, ``Transtailor: Pruning the pre-trained
  model for improved transfer learning,'' in \emph{Proceedings of the AAAI
  Conference on Artificial Intelligence}, vol.~35, no.~10, 2021, pp.
  8627--8634.

\bibitem[Chen et~al.(2021{\natexlab{a}})Chen, Frankle, Chang, Liu, Zhang,
  Carbin, and Wang]{chen2021lottery}
T.~Chen, J.~Frankle, S.~Chang, S.~Liu, Y.~Zhang, M.~Carbin, and Z.~Wang, ``The
  lottery tickets hypothesis for supervised and self-supervised pre-training in
  computer vision models,'' in \emph{Proceedings of the IEEE/CVF Conference on
  Computer Vision and Pattern Recognition}, 2021, pp. 16\,306--16\,316.

\bibitem[Chen et~al.(2021{\natexlab{b}})Chen, Cheng, Gan, Yuan, Zhang, and
  Wang]{chen2021chasing}
T.~Chen, Y.~Cheng, Z.~Gan, L.~Yuan, L.~Zhang, and Z.~Wang, ``Chasing sparsity
  in vision transformers: An end-to-end exploration,'' \emph{Advances in Neural
  Information Processing Systems}, vol.~34, pp. 19\,974--19\,988, 2021.

\bibitem[Myung et~al.(2022)Myung, Huh, Jang, Choe, Ryu, Kim, Kim, and
  Jeong]{myung2022pac}
S.~Myung, I.~Huh, W.~Jang, J.~M. Choe, J.~Ryu, D.~Kim, K.-E. Kim, and C.~Jeong,
  ``Pac-net: A model pruning approach to inductive transfer learning,'' in
  \emph{International Conference on Machine Learning}.\hskip 1em plus 0.5em
  minus 0.4em\relax PMLR, 2022, pp. 16\,240--16\,252.

\bibitem[Bahng et~al.(2022)Bahng, Jahanian, Sankaranarayanan, and
  Isola]{bahng2022visual}
H.~Bahng, A.~Jahanian, S.~Sankaranarayanan, and P.~Isola, ``Exploring visual
  prompts for adapting large-scale models,'' \emph{arXiv preprint
  arXiv:2203.17274}, vol.~1, no.~3, p.~4, 2022.

\bibitem[Elsayed et~al.(2018)Elsayed, Goodfellow, and
  Sohl-Dickstein]{elsayed2018adversarial}
G.~F. Elsayed, I.~Goodfellow, and J.~Sohl-Dickstein, ``Adversarial
  reprogramming of neural networks,'' \emph{arXiv preprint arXiv:1806.11146},
  2018.

\bibitem[Chen(2022)]{chen2022model}
P.-Y. Chen, ``Model reprogramming: Resource-efficient cross-domain machine
  learning,'' \emph{arXiv preprint arXiv:2202.10629}, 2022.

\bibitem[Zheng et~al.(2021)Zheng, Feng, Xia, Jiang, Demontis, Pintor, Biggio,
  and Roli]{zheng2021adversarial}
Y.~Zheng, X.~Feng, Z.~Xia, X.~Jiang, A.~Demontis, M.~Pintor, B.~Biggio, and
  F.~Roli, ``Why adversarial reprogramming works, when it fails, and how to
  tell the difference,'' \emph{arXiv preprint arXiv:2108.11673}, 2021.

\bibitem[Neekhara et~al.(2018)Neekhara, Hussain, Dubnov, and
  Koushanfar]{neekhara2018adversarial}
P.~Neekhara, S.~Hussain, S.~Dubnov, and F.~Koushanfar, ``Adversarial
  reprogramming of text classification neural networks,'' \emph{arXiv preprint
  arXiv:1809.01829}, 2018.

\bibitem[Neekhara et~al.(2022)Neekhara, Hussain, Du, Dubnov, Koushanfar, and
  McAuley]{neekhara2022cross}
P.~Neekhara, S.~Hussain, J.~Du, S.~Dubnov, F.~Koushanfar, and J.~McAuley,
  ``Cross-modal adversarial reprogramming,'' in \emph{Proceedings of the
  IEEE/CVF Winter Conference on Applications of Computer Vision}, 2022, pp.
  2427--2435.

\bibitem[Chen et~al.(2021{\natexlab{c}})Chen, Fan, and Ye]{chen2021adversarial}
L.~Chen, Y.~Fan, and Y.~Ye, ``Adversarial reprogramming of pretrained neural
  networks for fraud detection,'' in \emph{Proceedings of the 30th ACM
  International Conference on Information \& Knowledge Management}, 2021, pp.
  2935--2939.

\bibitem[Chen et~al.(2022)Chen, Yao, Chen, Zhang, and
  Liu]{chen2022understanding}
A.~Chen, Y.~Yao, P.-Y. Chen, Y.~Zhang, and S.~Liu, ``Understanding and
  improving visual prompting: A label-mapping perspective,'' \emph{arXiv
  preprint arXiv:2211.11635}, 2022.

\bibitem[Chen et~al.(2023)Chen, Lorenz, Yao, Chen, and Liu]{chen2023visual}
A.~Chen, P.~Lorenz, Y.~Yao, P.-Y. Chen, and S.~Liu, ``Visual prompting for
  adversarial robustness,'' in \emph{ICASSP 2023-2023 IEEE International
  Conference on Acoustics, Speech and Signal Processing (ICASSP)}.\hskip 1em
  plus 0.5em minus 0.4em\relax IEEE, 2023, pp. 1--5.

\bibitem[Zhang et~al.(2022{\natexlab{a}})Zhang, Zhang, Zhang, Fan, Li, Liu, and
  Chang]{zhang2022fairness}
G.~Zhang, Y.~Zhang, Y.~Zhang, W.~Fan, Q.~Li, S.~Liu, and S.~Chang, ``Fairness
  reprogramming,'' \emph{arXiv preprint arXiv:2209.10222}, 2022.

\bibitem[Zhang et~al.(2021)Zhang, Zhao, Yu, and Poupart]{zhang2021quantifying}
G.~Zhang, H.~Zhao, Y.~Yu, and P.~Poupart, ``Quantifying and improving
  transferability in domain generalization,'' \emph{Advances in Neural
  Information Processing Systems}, vol.~34, pp. 10\,957--10\,970, 2021.

\bibitem[Zhang et~al.(2023)Zhang, Khanduri, Tsaknakis, Yao, Hong, and
  Liu]{zhang2023introduction}
Y.~Zhang, P.~Khanduri, I.~Tsaknakis, Y.~Yao, M.~Hong, and S.~Liu, ``An
  introduction to bi-level optimization: Foundations and applications in signal
  processing and machine learning,'' \emph{arXiv preprint arXiv:2308.00788},
  2023.

\bibitem[Nilsback and Zisserman(2008)]{nilsback2008automated}
M.-E. Nilsback and A.~Zisserman, ``Automated flower classification over a large
  number of classes,'' in \emph{2008 Sixth Indian Conference on Computer
  Vision, Graphics \& Image Processing}.\hskip 1em plus 0.5em minus 0.4em\relax
  IEEE, 2008, pp. 722--729.

\bibitem[Caron et~al.(2018)Caron, Bojanowski, Joulin, and Douze]{caron2018deep}
M.~Caron, P.~Bojanowski, A.~Joulin, and M.~Douze, ``Deep clustering for
  unsupervised learning of visual features,'' in \emph{Proceedings of the
  European conference on computer vision (ECCV)}, 2018, pp. 132--149.

\bibitem[Yan et~al.(2020)Yan, Misra, Gupta, Ghadiyaram, and
  Mahajan]{yan2020clusterfit}
X.~Yan, I.~Misra, A.~Gupta, D.~Ghadiyaram, and D.~Mahajan, ``Clusterfit:
  Improving generalization of visual representations,'' in \emph{Proceedings of
  the IEEE/CVF Conference on Computer Vision and Pattern Recognition}, 2020,
  pp. 6509--6518.

\bibitem[Fan et~al.(2021{\natexlab{a}})Fan, Liu, Chen, Zhang, and
  Gan]{fan2021does}
L.~Fan, S.~Liu, P.-Y. Chen, G.~Zhang, and C.~Gan, ``When does contrastive
  learning preserve adversarial robustness from pretraining to finetuning?''
  \emph{Advances in Neural Information Processing Systems}, vol.~34, pp.
  21\,480--21\,492, 2021.

\bibitem[Evci et~al.(2022)Evci, Dumoulin, Larochelle, and
  Mozer]{evci2022head2toe}
U.~Evci, V.~Dumoulin, H.~Larochelle, and M.~C. Mozer, ``Head2toe: Utilizing
  intermediate representations for better transfer learning,'' in
  \emph{International Conference on Machine Learning}.\hskip 1em plus 0.5em
  minus 0.4em\relax PMLR, 2022, pp. 6009--6033.

\bibitem[Cimpoi et~al.(2014)Cimpoi, Maji, Kokkinos, Mohamed, and
  Vedaldi]{cimpoi2014describing}
M.~Cimpoi, S.~Maji, I.~Kokkinos, S.~Mohamed, and A.~Vedaldi, ``Describing
  textures in the wild,'' in \emph{Proceedings of the IEEE conference on
  computer vision and pattern recognition}, 2014, pp. 3606--3613.

\bibitem[Soomro et~al.(2012)Soomro, Zamir, and Shah]{soomro2012ucf101}
K.~Soomro, A.~R. Zamir, and M.~Shah, ``Ucf101: A dataset of 101 human actions
  classes from videos in the wild,'' \emph{arXiv preprint arXiv:1212.0402},
  2012.

\bibitem[Bossard et~al.(2014)Bossard, Guillaumin, and Gool]{bossard2014food}
L.~Bossard, M.~Guillaumin, and L.~V. Gool, ``Food-101--mining discriminative
  components with random forests,'' in \emph{European conference on computer
  vision}.\hskip 1em plus 0.5em minus 0.4em\relax Springer, 2014, pp. 446--461.

\bibitem[Xiao et~al.(2010)Xiao, Hays, Ehinger, Oliva, and
  Torralba]{xiao2010sun}
J.~Xiao, J.~Hays, K.~A. Ehinger, A.~Oliva, and A.~Torralba, ``Sun database:
  Large-scale scene recognition from abbey to zoo,'' in \emph{2010 IEEE
  computer society conference on computer vision and pattern
  recognition}.\hskip 1em plus 0.5em minus 0.4em\relax IEEE, 2010, pp.
  3485--3492.

\bibitem[Krizhevsky et~al.(2009)Krizhevsky, Hinton,
  et~al.]{krizhevsky2009learning}
A.~Krizhevsky, G.~Hinton \emph{et~al.}, ``Learning multiple layers of features
  from tiny images,'' \emph{cs.utoronto.ca}, 2009.

\bibitem[Dosovitskiy et~al.(2020)Dosovitskiy, Beyer, Kolesnikov, Weissenborn,
  Zhai, Unterthiner, Dehghani, Minderer, Heigold, Gelly,
  et~al.]{dosovitskiy2020image}
A.~Dosovitskiy, L.~Beyer, A.~Kolesnikov, D.~Weissenborn, X.~Zhai,
  T.~Unterthiner, M.~Dehghani, M.~Minderer, G.~Heigold, S.~Gelly \emph{et~al.},
  ``An image is worth 16x16 words: Transformers for image recognition at
  scale,'' \emph{arXiv preprint arXiv:2010.11929}, 2020.

\bibitem[Frankle and Carbin(2018)]{frankle2018lottery}
J.~Frankle and M.~Carbin, ``The lottery ticket hypothesis: Finding sparse,
  trainable neural networks,'' \emph{arXiv preprint arXiv:1803.03635}, 2018.

\bibitem[Chen et~al.(2020{\natexlab{d}})Chen, Frankle, Chang, Liu, Zhang, Wang,
  and Carbin]{chen2020lottery}
T.~Chen, J.~Frankle, S.~Chang, S.~Liu, Y.~Zhang, Z.~Wang, and M.~Carbin, ``The
  lottery ticket hypothesis for pre-trained bert networks,'' \emph{Advances in
  neural information processing systems}, vol.~33, pp. 15\,834--15\,846, 2020.

\bibitem[Zhang et~al.(2022{\natexlab{b}})Zhang, Yao, Ram, Zhao, Chen, Hong,
  Wang, and Liu]{zhang2022advancing}
Y.~Zhang, Y.~Yao, P.~Ram, P.~Zhao, T.~Chen, M.~Hong, Y.~Wang, and S.~Liu,
  ``Advancing model pruning via bi-level optimization,'' \emph{arXiv preprint
  arXiv:2210.04092}, 2022.

\bibitem[Madry et~al.(2017)Madry, Makelov, Schmidt, Tsipras, and
  Vladu]{madry2017towards}
A.~Madry, A.~Makelov, L.~Schmidt, D.~Tsipras, and A.~Vladu, ``Towards deep
  learning models resistant to adversarial attacks,'' \emph{arXiv preprint
  arXiv:1706.06083}, 2017.

\bibitem[Fan et~al.(2021{\natexlab{b}})Fan, Xiong, Mangalam, Li, Yan, Malik,
  and Feichtenhofer]{fan2021multiscale}
H.~Fan, B.~Xiong, K.~Mangalam, Y.~Li, Z.~Yan, J.~Malik, and C.~Feichtenhofer,
  ``Multiscale vision transformers,'' in \emph{Proceedings of the IEEE/CVF
  international conference on computer vision}, 2021, pp. 6824--6835.

\bibitem[Hendrycks and Dietterich(2019)]{hendrycks2019benchmarking}
D.~Hendrycks and T.~Dietterich, ``Benchmarking neural network robustness to
  common corruptions and perturbations,'' \emph{arXiv preprint
  arXiv:1903.12261}, 2019.

\bibitem[Zhai et~al.(2019)Zhai, Puigcerver, Kolesnikov, Ruyssen, Riquelme,
  Lucic, Djolonga, Pinto, Neumann, Dosovitskiy, et~al.]{zhai2019large}
X.~Zhai, J.~Puigcerver, A.~Kolesnikov, P.~Ruyssen, C.~Riquelme, M.~Lucic,
  J.~Djolonga, A.~S. Pinto, M.~Neumann, A.~Dosovitskiy \emph{et~al.}, ``A
  large-scale study of representation learning with the visual task adaptation
  benchmark,'' \emph{arXiv preprint arXiv:1910.04867}, 2019.

\bibitem[Keskar et~al.(2016)Keskar, Mudigere, Nocedal, Smelyanskiy, and
  Tang]{keskar2016large}
N.~S. Keskar, D.~Mudigere, J.~Nocedal, M.~Smelyanskiy, and P.~T.~P. Tang, ``On
  large-batch training for deep learning: Generalization gap and sharp
  minima,'' \emph{arXiv preprint arXiv:1609.04836}, 2016.

\bibitem[Bisla et~al.(2022)Bisla, Wang, and Choromanska]{bisla2022low}
D.~Bisla, J.~Wang, and A.~Choromanska, ``Low-pass filtering sgd for recovering
  flat optima in the deep learning optimization landscape,'' in
  \emph{International Conference on Artificial Intelligence and
  Statistics}.\hskip 1em plus 0.5em minus 0.4em\relax PMLR, 2022, pp.
  8299--8339.

\bibitem[Maddox et~al.(2020)Maddox, Benton, and Wilson]{maddox2020rethinking}
W.~J. Maddox, G.~Benton, and A.~G. Wilson, ``Rethinking parameter counting in
  deep models: Effective dimensionality revisited,'' \emph{arXiv preprint
  arXiv:2003.02139}, 2020.

\bibitem[Liang et~al.(2019)Liang, Poggio, Rakhlin, and Stokes]{liang2019fisher}
T.~Liang, T.~Poggio, A.~Rakhlin, and J.~Stokes, ``Fisher-rao metric, geometry,
  and complexity of neural networks,'' in \emph{The 22nd international
  conference on artificial intelligence and statistics}.\hskip 1em plus 0.5em
  minus 0.4em\relax PMLR, 2019, pp. 888--896.

\bibitem[Van~der Maaten and Hinton(2008)]{van2008visualizing}
L.~Van~der Maaten and G.~Hinton, ``Visualizing data using t-sne.''
  \emph{Journal of machine learning research}, vol.~9, no.~11, 2008.

\end{thebibliography}
}}

\clearpage\newpage
\section*{\Large{Appendix}}
\setcounter{section}{0}
\setcounter{figure}{0}
\setcounter{table}{0}
\makeatletter 
\renewcommand{\thesection}{\Alph{section}}
\renewcommand{\theHsection}{\Alph{section}}
\renewcommand{\thefigure}{A\arabic{figure}} 
\renewcommand{\theHfigure}{A\arabic{figure}} 
\renewcommand{\thetable}{A\arabic{table}}
\renewcommand{\theHtable}{A\arabic{table}}
\makeatother

\renewcommand{\thetable}{A\arabic{table}}
\setcounter{mylemma}{0}
\renewcommand{\themylemma}{A\arabic{mylemma}}
\setcounter{algorithm}{0}
\renewcommand{\thealgorithm}{A\arabic{algorithm}}
\setcounter{equation}{0}
\renewcommand{\theequation}{A\arabic{equation}}

\section{Experiment Settings}
\label{app: settings}

\subsection{Training Settings}
\label{app: train_settings}

\paragraph{General setups for DP (dataset pruning).} Regardless of the choice of the downstream tasks and DP methods, each DP process will go through a three-stage paradigm, namely \ding{182} source dataset pruning, \ding{183} pretraining based on the pruned source dataset, and \ding{184} finetuning on the target dataset using either LP (linear probing) or FF (full finetuning). Regarding baseline methods for source dataset pruning, we strictly follow the baselines' settings provided in the literature. As stated before, we utilize a small surrogate model (ResNet-18) pretrained on the full ImageNet to conduct  DP. The proportion of pruned class numbers to the total source class numbers determines pruning ratios for our methods: LM (label mapping) and FM (feature mapping). To ensure a fair comparison with other non-class-wise baselines ({\random}, {\gradnorm}, {\moderate}), the same pruning ratio is applied to the total number of training data points.

\paragraph{Pretraining setups.} We keep our pretraining procedure consistent for all models and strictly follow the settings released by the existing work \cite{jain2022data}. All the models are trained from scratch using Stochastic Gradient Descent (SGD) to minimize the standard cross-entropy loss in a multi-class classification setting. We use a batch size of 1024, momentum of 0.9, and a weight decay of $5\times10^{-4}$. The training utilizes a cyclic learning rate schedule, which begins with an initial learning rate of 0.5 and peaks at the second epoch. During training, we incorporate data augmentations such as random resized cropping and random horizontal flipping.

\paragraph{Downstream task finetuning settings.} Our approach to finetuning the pretrained model on downstream tasks involves using LP and FF. Details of the downstream datasets and the training configurations are presented in Tab.\,\ref{tab: app_downstream_data}, following   \cite{chen2023visual}. For LP, we employ the Adam optimizer, a multi-step decaying scheduler, and an initial learning rate of 0.1 across 50 total training epochs. As for FF, we utilize the Adam optimizer over 200 epochs with a cosine-annealing scheduler, an initial learning rate of $0.01$, and a weight decay of $5\times10^{-4}$.
All finetuning experiments employ a batch size of 256 and standard data augmentations, such as random resized cropping and horizontal flipping.

\paragraph{Training settings for SSL (self-supervised learning) and ViT.} Our SSL training settings   follow the configurations provided by \textsc{MoCov2}. Details of the pretraining and finetuning stages can be accessed at \url{https://github.com/facebookresearch/moco}. For the training of ViTs, we rely on the setting released in the original ViT paper \cite{dosovitskiy2020image} (see ViT/B in Table 3).

\begin{table}[h]
\centering
\caption{\footnotesize{Dataset attributes and training configurations of  8 downstream image classification datasets considered in this work.}}
\resizebox{0.7\columnwidth}{!}{%
\begin{tabular}{c|c|c|c|c|c}
\toprule[1pt]
\midrule
Dataset & Train Size & Test Size & Class Number & Batch Size & Rescaled Resolution \\
\midrule
Flowers102
& 4093
& 2463
& 102
& 256
& 224$\times$224
\\
DTD
& 2820
& 1692
& 47
& 256
& 224$\times$224
\\
UCF101
& 7639
& 3783
& 101
& 256
& 224$\times$224
\\
Food101
& 50500
& 30300
& 101
& 256
& 224$\times$224
\\
OxfordPets
& 2944
& 3669
& 37
& 256
& 224$\times$224
\\
StanfordCars
& 6509
& 8041
& 196
& 256
& 224$\times$224
\\
SUN397
& 15888
& 19850
& 397
& 256
& 224$\times$224
\\
CIFAR10
& 50000
& 10000
& 10
& 256
& 160$\times$160
\\
\midrule
\bottomrule[1pt]
\end{tabular}%
}
\label{tab: app_downstream_data}
\end{table}

\section{Additional Results}
\label{app: additional}

\begin{figure}[thb]
    \centerline{
\begin{tabular}{cccc}
    \hspace*{-2mm} \includegraphics[width=.23\textwidth,height=!]{figures/indomain_motivation/ID_rn101_flowers102_LP.pdf} &
    \hspace*{-4mm}  \includegraphics[width=.23\textwidth,height=!]{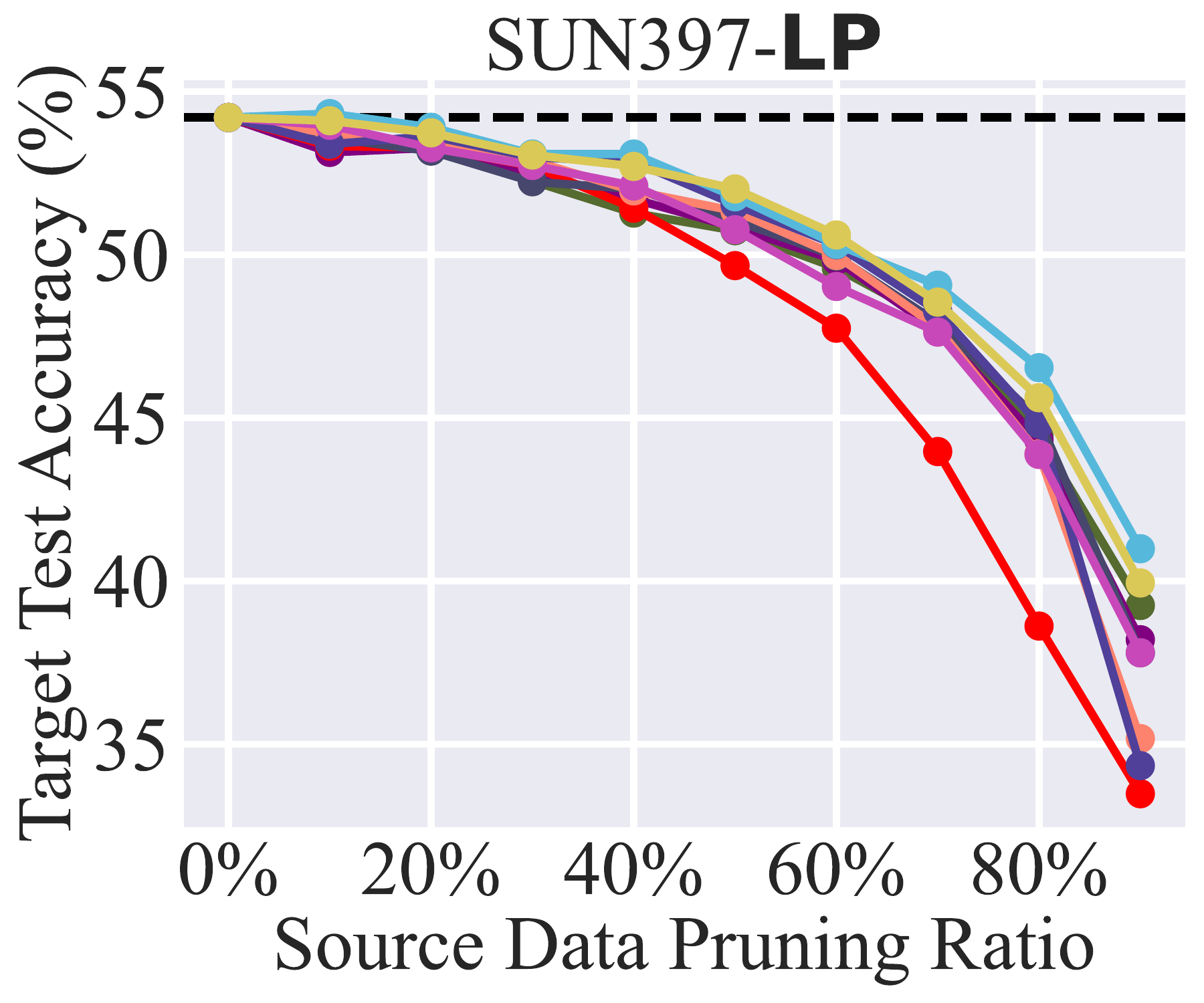} &
    \hspace*{-4mm}  \includegraphics[width=.23\textwidth,height=!]{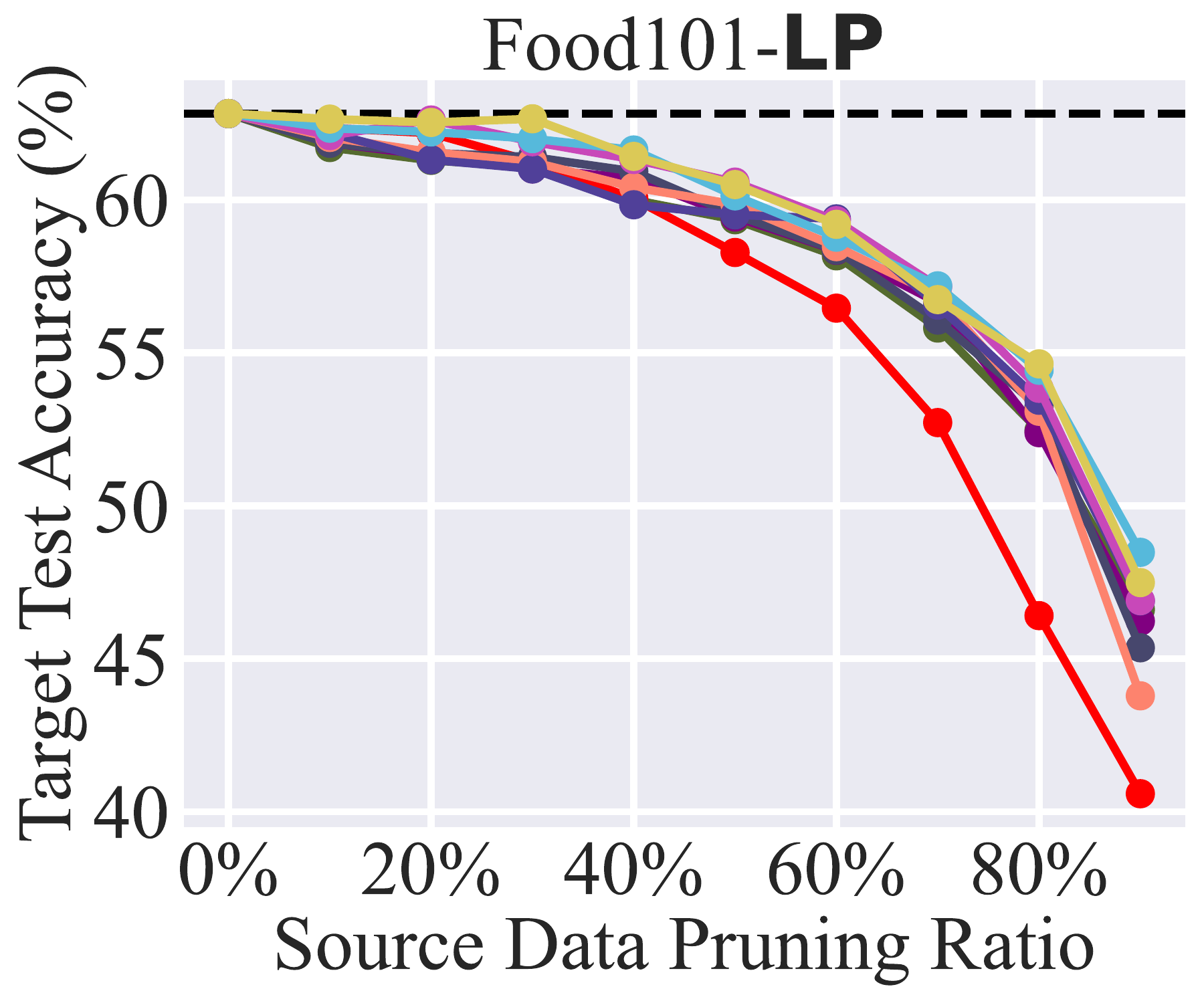} &
    \hspace*{-4mm} \includegraphics[width=.23\textwidth,height=!]{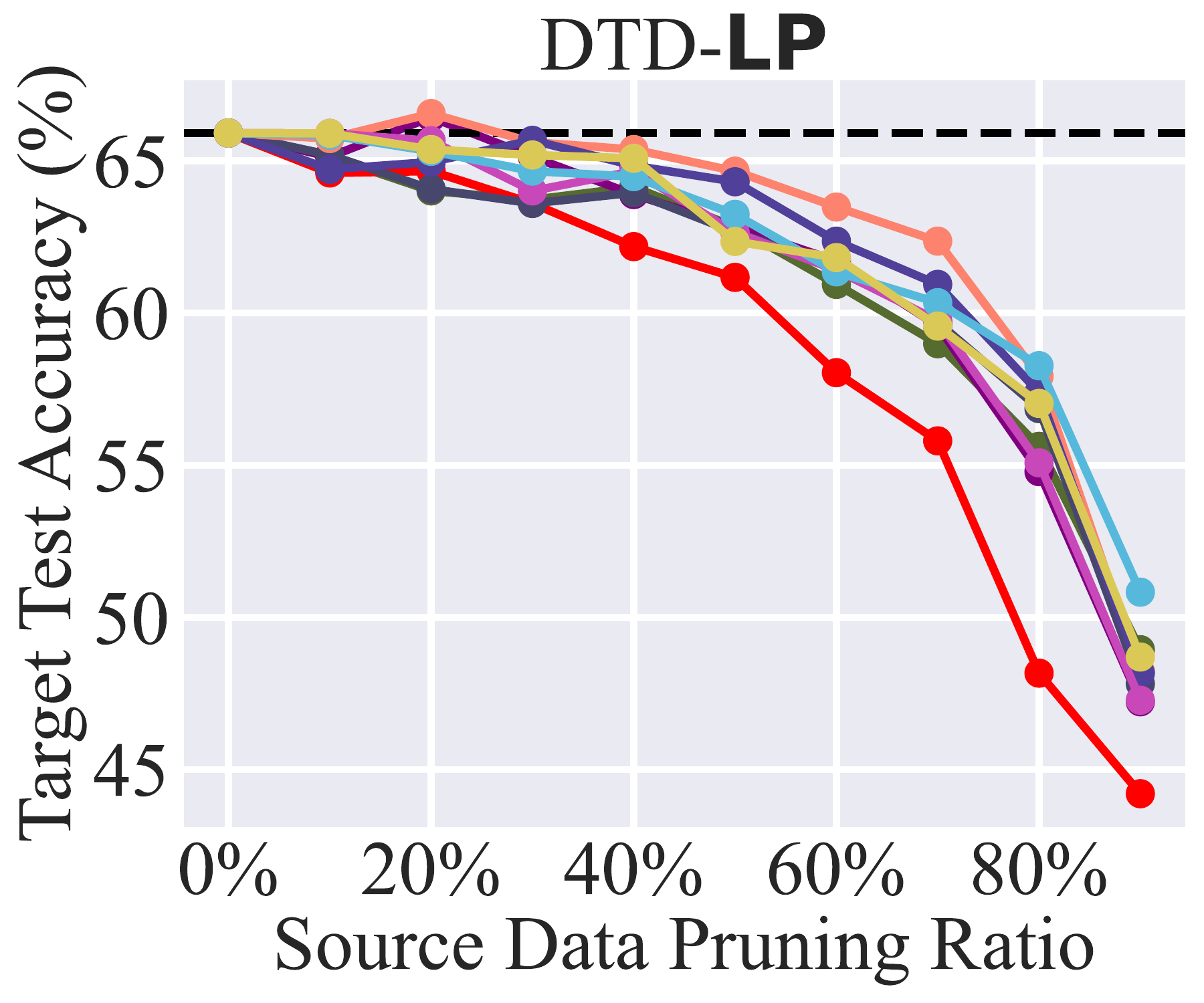} \\

    \includegraphics[width=.23\textwidth,height=!]{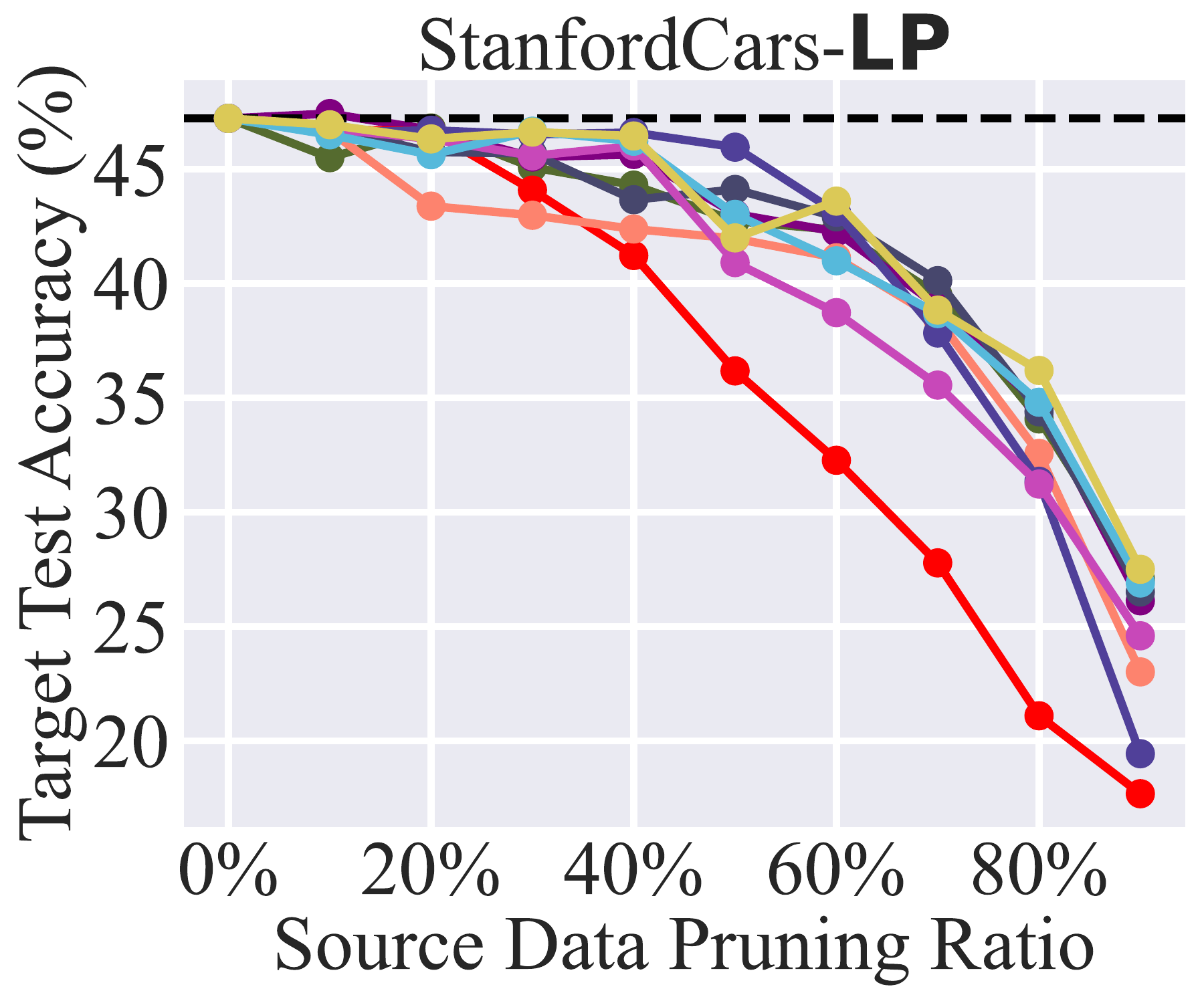} &
    \hspace*{-4mm}  \includegraphics[width=.23\textwidth,height=!]{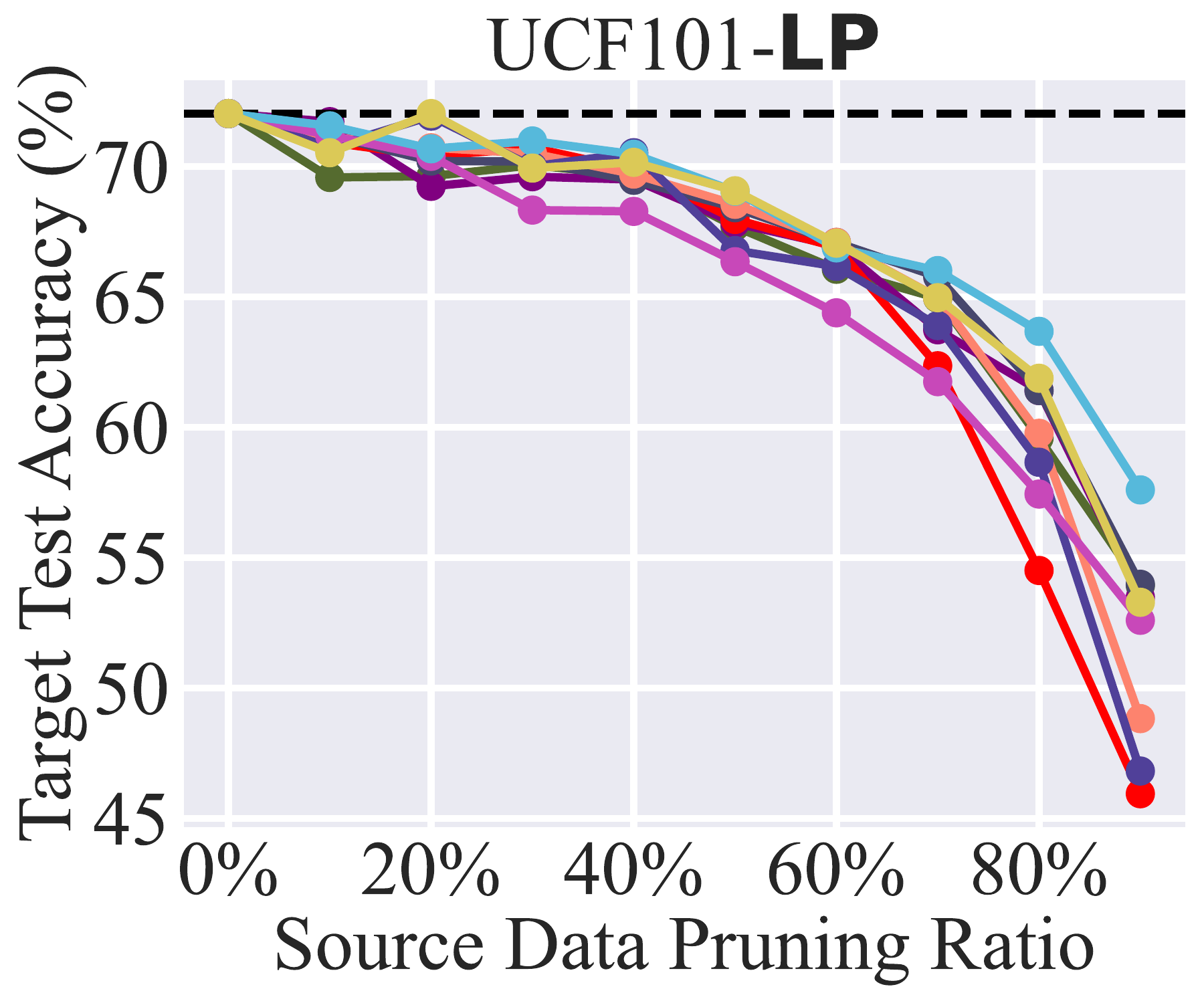} &
    \hspace*{-4mm}  \includegraphics[width=.23\textwidth,height=!]{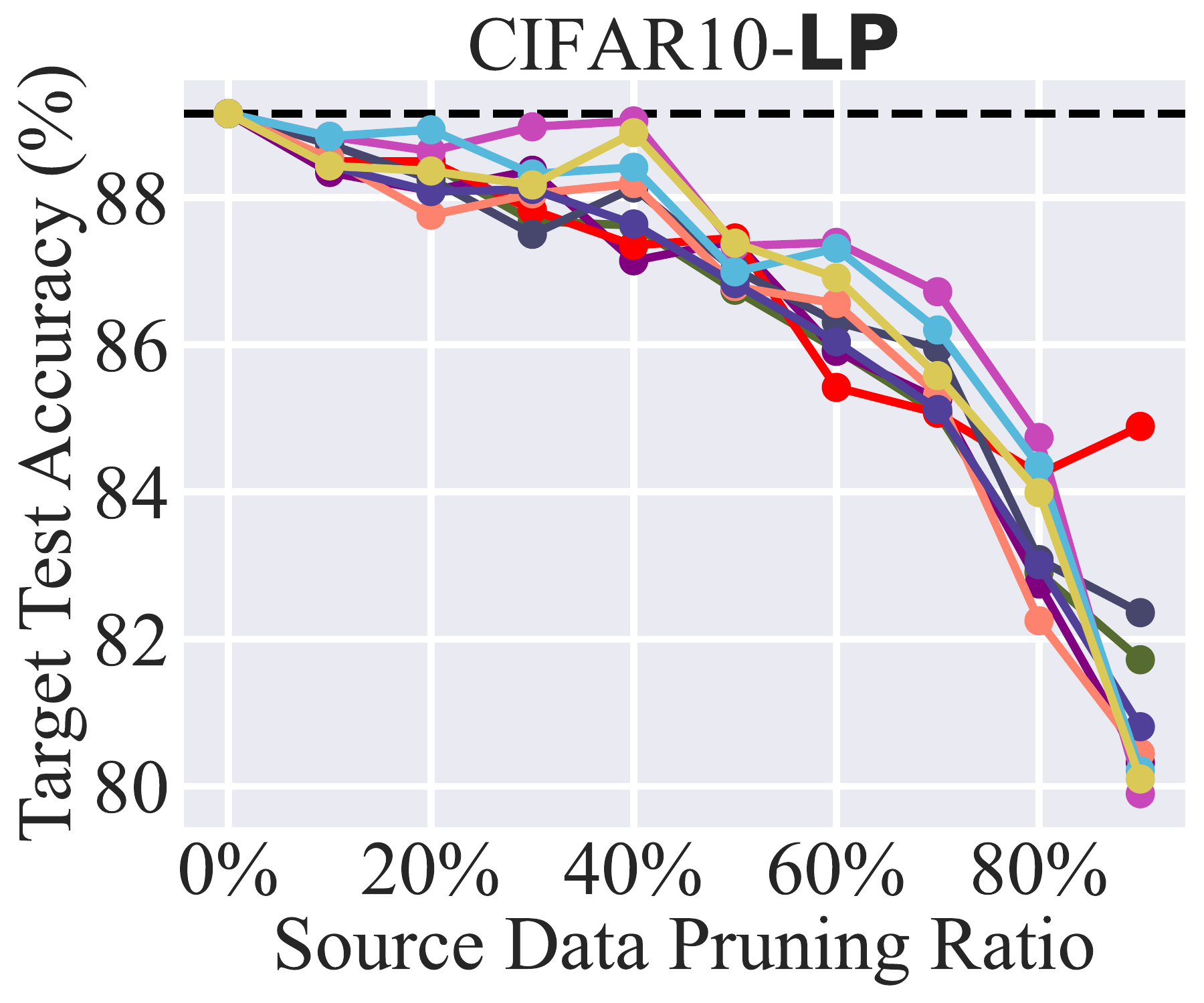} &
    \hspace*{-4mm} \includegraphics[width=.23\textwidth,height=!]{figures/indomain_motivation/ID_rn101_oxfordpets_LP.pdf} \\
    
    \end{tabular}}
    \caption{Extended performance comparison of different in-domain DP methods for transfer learning across all downstream tasks. Experiment settings are consistent with Fig.\,\ref{fig: main_lp}.}
    \label{fig: app_complete_main_figure}
    \vspace*{-5mm}
\end{figure}

\paragraph{Expanded performance evaluation of in-domain DP methods across all downstream datasets.}\textbf{Fig.\,\ref{fig: app_complete_main_figure}} expands on the performance comparisons in \textbf{Fig.\,\ref{fig: in_domain_motivation}}, providing a more thorough evaluation of various in-domain DP methods on all eight downstream datasets. The trends observed are consistent with those in \textbf{Fig.\,\ref{fig: in_domain_motivation}}: Random pruning shows a strong baseline method in DP for transfer learning compared to other state-of-the-art DP methods designed for non-transfer learning. This observation prompts us to explore more effective dataset pruning strategies for transfer learning. {\moderate} and {\gradnorm} are also demonstrating strong baselines, motivating us to choose them as the default DP baselines   in Section\,\ref{sec: exp}.

\begin{sidewaystable}[ph!]
    \centering
    \caption{Exact numbers and standard deviations for Fig. \ref{fig: main_lp}.}
        \resizebox{\textwidth}{!}{%
\begin{tabular}{cccccccccccccccccc}
\toprule[1pt]
\midrule
\multirow{2}{*}{Method}  & \multicolumn{17}{c}{Pruning Ratio}\\
                         & 5\%   & 10\%  & 15\%  & 20\%  & 25\%  & 30\%  & 35\%  & 40\%  & 45\%  & 50\%  & 55\%  & 60\%  & 65\%  & 70\%  & 75\%  & 80\%  & 85\%  \\ \midrule
\multicolumn{18}{c}{OxfordPets - LP} \\ \midrule
\random   & 89.81{$\pm 0.22$} & 89.32{$\pm 0.18$} & 89.92{$\pm 0.12$} & 88.99{$\pm 0.34$} & 89.26{$\pm 0.13$} & 89.26{$\pm 0.11$} & 88.66{$\pm 0.38$} & 89.23{$\pm 0.28$} & 88.23{$\pm 0.07$} & 87.76{$\pm 0.05$} & 87.08{$\pm 0.27$} & 86.92{$\pm 0.23$} & 86.10{$\pm 0.03$}  & 84.63{$\pm 0.31$} & 82.56{$\pm 0.09$} & 82.31{$\pm 0.2$} & 76.59{$\pm 0.35$} \\
\moderate & 89.67{$\pm 0.36$} & 89.59{$\pm 0.29$} & 89.86{$\pm 0.15$} & 88.91{$\pm 0.14$} & 88.96{$\pm 0.10$} & 88.63{$\pm 0.25$} & 88.93{$\pm 0.32$} & 88.96{$\pm 0.05$} & 87.79{$\pm 0.37$} & 87.24{$\pm 0.17$} & 88.01{$\pm 0.28$} & 86.73{$\pm 0.02$} & 86.59{$\pm 0.04$} & 85.17{$\pm 0.08$} & 83.84{$\pm 0.04$} & 82.34{$\pm 0.33$} & 78.71{$\pm 0.27$} \\
\gradnorm & 89.70{$\pm 0.20$}  & 89.78{$\pm 0.16$} & 89.78{$\pm 0.24$} & 89.04{$\pm 0.38$} & 89.40{$\pm 0.11$}  & 89.23{$\pm 0.07$} & 88.72{$\pm 0.04$} & 88.44{$\pm 0.39$} & 88.93{$\pm 0.28$} & 87.63{$\pm 0.20$} & 86.86{$\pm 0.29$} & 86.51{$\pm 0.17$} & 84.85{$\pm 0.02$} & 84.46{$\pm 0.14$} & 81.36{$\pm 0.33$} & 78.63{$\pm 0.13$} & 74.19{$\pm 0.08$} \\
LM                       & 89.72{$\pm 0.1$} & 89.34{$\pm 0.04$} & 90.27{$\pm 0.37$} & 90.11{$\pm 0.27$} & 89.97{$\pm 0.38$} & 90.02{$\pm 0.19$} & 89.72{$\pm 0.15$} & 89.62{$\pm 0.24$} & 89.46{$\pm 0.16$} & 89.45{$\pm 0.06$} & 89.92{$\pm 0.11$} & 90.02{$\pm 0.27$} & 89.89{$\pm 0.18$} & 90.19{$\pm 0.14$} & 90.16{$\pm 0.38$} & 90.11{$\pm 0.03$} & 88.85{$\pm 0.37$} \\
FM                       & 90.35{$\pm 0.32$} & 89.78{$\pm 0.07$} & 89.70{$\pm 0.24$}  & 89.45{$\pm 0.29$} & 90.02{$\pm 0.22$} & 89.83{$\pm 0.37$} & 89.89{$\pm 0.03$} & 89.72{$\pm 0.27$} & 89.56{$\pm 0.06$} & 89.72{$\pm 0.19$} & 89.94{$\pm 0.11$} & 90.08{$\pm 0.36$} & 90.62{$\pm 0.03$} & 89.78{$\pm 0.10$} & 90.32{$\pm 0.38$} & 89.64{$\pm 0.15$} & 90.16{$\pm 0.07$} \\ \midrule
\multicolumn{18}{c}{SUN397 - LP} \\ \midrule
\random                 & 51.25$\pm0.14$ & 51.23$\pm0.37$ & 51.23$\pm0.27$ & 51.16$\pm0.32$ & 51.06$\pm0.21$ & 50.88$\pm0.08$ & 51.11$\pm0.34$ & 51.32$\pm0.28$ & 50.80$\pm0.19$ & 50.31$\pm0.27$ & 50.90$\pm0.36$ & 50.33$\pm0.17$ & 50.41$\pm0.07$ & 49.88$\pm0.24$ & 49.11$\pm0.10$ & 49.45$\pm0.14$ & 48.63$\pm0.31$ \\
\moderate               & 51.34$\pm0.10$ & 51.17$\pm0.35$ & 51.36$\pm0.38$ & 51.73$\pm0.07$ & 51.45$\pm0.20$ & 51.48$\pm0.36$ & 51.04$\pm0.12$ & 51.16$\pm0.15$ & 50.68$\pm0.14$ & 50.70$\pm0.21$ & 50.87$\pm0.27$ & 50.06$\pm0.36$ & 50.34$\pm0.38$ & 49.83$\pm0.12$ & 50.16$\pm0.34$ & 49.51$\pm0.25$ & 49.00$\pm0.20$ \\
\gradnorm & 51.38$\pm0.30$ & 51.25$\pm0.24$ & 51.15$\pm0.33$ & 51.42$\pm0.14$ & 51.44$\pm0.38$ & 51.20$\pm0.24$ & 51.17$\pm0.13$ & 50.63$\pm0.32$ & 50.68$\pm0.29$ & 50.28$\pm0.36$ & 49.81$\pm0.38$ & 50.28$\pm0.10$ & 50.02$\pm0.28$ & 49.84$\pm0.12$ & 48.87$\pm0.21$ & 48.92$\pm0.11$ & 47.95$\pm0.27$ \\
LM                      & 51.70$\pm0.18$ & 51.78$\pm0.26$ & 51.76$\pm0.20$ & 51.67$\pm0.35$ & 51.52$\pm0.32$ & 51.55$\pm0.28$ & 51.87$\pm0.30$ & 51.40$\pm0.16$ & 51.46$\pm0.32$ & 51.62$\pm0.14$ & 51.48$\pm0.23$ & 51.39$\pm0.22$ & 51.65$\pm0.18$ & 50.95$\pm0.33$ & 49.62$\pm0.16$ & 49.53$\pm0.21$ & 48.93$\pm0.27$ \\
FM                      & 51.88$\pm0.26$ & 51.72$\pm0.31$ & 51.68$\pm0.18$ & 51.50$\pm0.14$ & 51.70$\pm0.37$ & 51.79$\pm0.30$ & 51.74$\pm0.24$ & 51.63$\pm0.38$ & 51.54$\pm0.33$ & 51.61$\pm0.28$ & 51.72$\pm0.27$ & 51.42$\pm0.32$ & 51.57$\pm0.23$ & 51.08$\pm0.22$ & 49.72$\pm0.18$ & 49.73$\pm0.21$ & 48.97$\pm0.35$ \\ \midrule
\multicolumn{18}{c}{Food101 - LP} \\ \midrule
\random                 & 62.22$\pm0.30$ & 62.64$\pm0.21$ & 62.32$\pm0.18$ & 62.53$\pm0.33$ & 62.23$\pm0.17$ & 62.65$\pm0.27$ & 61.33$\pm0.31$ & 61.42$\pm0.39$ & 60.91$\pm0.26$ & 60.50$\pm0.21$ & 60.19$\pm0.34$ & 59.20$\pm0.20$ & 58.15$\pm0.32$ & 56.74$\pm0.27$ & 54.70$\pm0.19$ & 54.64$\pm0.24$ & 51.12$\pm0.22$ \\
\moderate               & 62.47$\pm0.19$ & 62.34$\pm0.28$ & 62.49$\pm0.27$ & 62.22$\pm0.16$ & 62.45$\pm0.36$ & 62.00$\pm0.24$ & 61.76$\pm0.30$ & 61.63$\pm0.29$ & 60.75$\pm0.16$ & 60.13$\pm0.26$ & 59.42$\pm0.24$ & 58.78$\pm0.35$ & 58.80$\pm0.30$ & 57.19$\pm0.25$ & 55.73$\pm0.37$ & 54.43$\pm0.24$ & 51.79$\pm0.16$ \\
\gradnorm & 62.70$\pm0.18$ & 62.10$\pm0.24$ & 62.30$\pm0.33$ & 62.61$\pm0.22$ & 62.25$\pm0.27$ & 61.90$\pm0.24$ & 62.27$\pm0.31$ & 61.32$\pm0.34$ & 61.60$\pm0.22$ & 60.58$\pm0.24$ & 59.72$\pm0.25$ & 59.32$\pm0.33$ & 58.19$\pm0.27$ & 57.17$\pm0.21$ & 55.54$\pm0.19$ & 53.84$\pm0.30$ & 50.79$\pm0.28$ \\
LM                      & 63.17$\pm0.16$ & 63.39$\pm0.27$ & 63.60$\pm0.18$ & 63.55$\pm0.32$ & 63.30$\pm0.16$ & 63.50$\pm0.33$ & 63.97$\pm0.23$ & 63.08$\pm0.29$ & 63.34$\pm0.25$ & 63.47$\pm0.33$ & 63.09$\pm0.24$ & 63.05$\pm0.36$ & 62.78$\pm0.26$ & 62.14$\pm0.30$ & 61.51$\pm0.33$ & 61.15$\pm0.22$ & 58.78$\pm0.30$ \\
FM                      & 62.89$\pm0.28$ & 63.29$\pm0.24$ & 63.55$\pm0.16$ & 63.53$\pm0.37$ & 63.26$\pm0.30$ & 63.34$\pm0.27$ & 63.20$\pm0.35$ & 63.24$\pm0.22$ & 62.73$\pm0.32$ & 62.84$\pm0.24$ & 63.06$\pm0.23$ & 62.55$\pm0.16$ & 61.82$\pm0.35$ & 61.71$\pm0.32$ & 61.38$\pm0.24$ & 60.50$\pm0.36$ & 58.24$\pm0.32$ \\ \midrule
\multicolumn{18}{c}{DTD - LP} \\ \midrule
\random                 & 66.13$\pm0.20$ & 65.90$\pm0.17$ & 66.84$\pm0.21$ & 65.37$\pm0.23$ & 65.07$\pm0.30$ & 65.19$\pm0.19$ & 64.42$\pm0.22$ & 65.08$\pm0.18$ & 63.30$\pm0.25$ & 62.35$\pm0.32$ & 61.95$\pm0.29$ & 61.82$\pm0.24$ & 62.00$\pm0.28$ & 59.57$\pm0.31$ & 58.69$\pm0.34$ & 57.03$\pm0.33$ & 52.54$\pm0.26$ \\
\moderate               & 66.73$\pm0.18$ & 65.84$\pm0.30$ & 65.72$\pm0.16$ & 65.31$\pm0.33$ & 65.60$\pm0.22$ & 64.66$\pm0.27$ & 65.25$\pm0.16$ & 64.48$\pm0.35$ & 63.00$\pm0.22$ & 63.24$\pm0.36$ & 63.00$\pm0.33$ & 61.35$\pm0.24$ & 60.87$\pm0.25$ & 60.34$\pm0.16$ & 58.92$\pm0.34$ & 58.27$\pm0.28$ & 54.26$\pm0.24$ \\
\gradnorm & 66.67$\pm0.24$ & 65.90$\pm0.22$ & 65.96$\pm0.17$ & 65.66$\pm0.26$ & 65.13$\pm0.27$ & 64.01$\pm0.35$ & 64.42$\pm0.16$ & 64.66$\pm0.22$ & 63.30$\pm0.28$ & 62.65$\pm0.16$ & 61.47$\pm0.23$ & 61.41$\pm0.29$ & 60.40$\pm0.19$ & 59.75$\pm0.35$ & 57.27$\pm0.16$ & 55.08$\pm0.28$ & 51.95$\pm0.26$ \\
LM                      & 66.73$\pm0.17$ & 66.21$\pm0.21$ & 66.43$\pm0.22$ & 66.54$\pm0.25$ & 67.02$\pm0.33$ & 66.25$\pm0.28$ & 67.43$\pm0.22$ & 66.38$\pm0.24$ & 66.33$\pm0.16$ & 66.25$\pm0.21$ & 65.80$\pm0.22$ & 65.48$\pm0.16$ & 64.24$\pm0.32$ & 62.88$\pm0.35$ & 62.17$\pm0.26$ & 60.64$\pm0.28$ & 58.87$\pm0.32$ \\
FM                      & 66.19$\pm0.23$ & 67.38$\pm0.32$ & 66.67$\pm0.30$ & 66.22$\pm0.16$ & 66.31$\pm0.22$ & 66.13$\pm0.19$ & 66.55$\pm0.23$ & 66.72$\pm0.17$ & 66.61$\pm0.27$ & 66.77$\pm0.33$ & 66.78$\pm0.22$ & 64.95$\pm0.29$ & 63.48$\pm0.30$ & 63.30$\pm0.35$ & 62.07$\pm0.26$ & 60.34$\pm0.18$ & 60.05$\pm0.23$ \\ \midrule
\multicolumn{18}{c}{StanfordCars - LP} \\ \midrule
\random                 & 46.47$\pm0.23$ & 46.95$\pm0.20$ & 46.32$\pm0.21$ & 46.33$\pm0.19$ & 46.49$\pm0.32$ & 46.63$\pm0.28$ & 45.39$\pm0.30$ & 46.46$\pm0.18$ & 43.76$\pm0.25$ & 42.00$\pm0.28$ & 43.74$\pm0.33$ & 43.61$\pm0.22$ & 41.80$\pm0.35$ & 38.84$\pm0.29$ & 37.81$\pm0.30$ & 36.20$\pm0.31$ & 32.76$\pm0.28$ \\
\moderate               & 46.66$\pm0.16$ & 46.51$\pm0.24$ & 46.85$\pm0.30$ & 45.64$\pm0.16$ & 46.03$\pm0.21$ & 46.67$\pm0.22$ & 45.21$\pm0.17$ & 46.23$\pm0.32$ & 45.47$\pm0.29$ & 43.03$\pm0.33$ & 42.35$\pm0.30$ & 41.00$\pm0.28$ & 41.09$\pm0.24$ & 38.69$\pm0.26$ & 37.89$\pm0.33$ & 34.81$\pm0.22$ & 29.97$\pm0.32$ \\
\gradnorm & 46.70$\pm0.27$ & 46.77$\pm0.19$ & 46.62$\pm0.20$ & 46.33$\pm0.30$ & 45.55$\pm0.16$ & 45.59$\pm0.33$ & 45.94$\pm0.22$ & 46.02$\pm0.28$ & 44.40$\pm0.24$ & 40.92$\pm0.30$ & 39.93$\pm0.31$ & 38.73$\pm0.19$ & 37.71$\pm0.28$ & 35.56$\pm0.34$ & 32.51$\pm0.16$ & 31.23$\pm0.28$ & 28.06$\pm0.26$ \\
LM                      & 47.60$\pm0.28$ & 48.60$\pm0.24$ & 47.52$\pm0.30$ & 48.12$\pm0.33$ & 46.81$\pm0.20$ & 47.49$\pm0.22$ & 48.78$\pm0.27$ & 47.36$\pm0.23$ & 47.63$\pm0.26$ & 47.27$\pm0.29$ & 47.42$\pm0.30$ & 46.88$\pm0.24$ & 47.56$\pm0.26$ & 46.45$\pm0.33$ & 46.80$\pm0.18$ & 45.07$\pm0.25$ & 42.87$\pm0.32$ \\
FM                      & 47.69$\pm0.22$ & 47.98$\pm0.16$ & 48.69$\pm0.33$ & 47.21$\pm0.19$ & 47.23$\pm0.32$ & 48.17$\pm0.18$ & 47.80$\pm0.30$ & 47.41$\pm0.28$ & 47.63$\pm0.26$ & 48.46$\pm0.21$ & 47.08$\pm0.30$ & 47.37$\pm0.16$ & 46.93$\pm0.31$ & 46.64$\pm0.33$ & 46.47$\pm0.22$ & 44.45$\pm0.24$ & 42.10$\pm0.28$ \\ \midrule
\multicolumn{18}{c}{UCF101 - LP} \\ \midrule
\random                 & 71.96$\pm0.31$ & 70.53$\pm0.28$ & 71.53$\pm0.33$ & 72.03$\pm0.30$ & 70.37$\pm0.24$ & 69.94$\pm0.31$ & 70.84$\pm0.32$ & 70.16$\pm0.27$ & 69.07$\pm0.29$ & 69.07$\pm0.28$ & 67.27$\pm0.25$ & 67.06$\pm0.24$ & 65.98$\pm0.26$ & 64.97$\pm0.30$ & 63.84$\pm0.22$ & 61.88$\pm0.23$ & 58.60$\pm0.32$ \\
\moderate               & 72.72$\pm0.24$ & 71.56$\pm0.29$ & 71.58$\pm0.28$ & 70.68$\pm0.33$ & 70.13$\pm0.25$ & 70.98$\pm0.27$ & 70.18$\pm0.29$ & 70.47$\pm0.26$ & 70.31$\pm0.32$ & 69.05$\pm0.24$ & 67.78$\pm0.31$ & 66.88$\pm0.28$ & 66.03$\pm0.29$ & 66.01$\pm0.28$ & 64.63$\pm0.27$ & 63.68$\pm0.33$ & 60.38$\pm0.24$ \\
\gradnorm & 71.93$\pm0.22$ & 71.19$\pm0.25$ & 71.00$\pm0.33$ & 70.39$\pm0.29$ & 69.52$\pm0.24$ & 68.33$\pm0.30$ & 68.81$\pm0.32$ & 68.28$\pm0.22$ & 67.14$\pm0.26$ & 66.35$\pm0.31$ & 64.42$\pm0.33$ & 64.39$\pm0.30$ & 63.28$\pm0.31$ & 61.75$\pm0.33$ & 58.95$\pm0.26$ & 57.44$\pm0.33$ & 55.70$\pm0.28$ \\
LM                      & 73.32$\pm0.28$ & 73.54$\pm0.29$ & 72.77$\pm0.25$ & 72.82$\pm0.30$ & 72.69$\pm0.29$ & 72.58$\pm0.24$ & 72.35$\pm0.32$ & 72.83$\pm0.29$ & 72.51$\pm0.27$ & 71.82$\pm0.32$ & 71.76$\pm0.31$ & 71.48$\pm0.33$ & 70.98$\pm0.24$ & 69.44$\pm0.28$ & 68.75$\pm0.26$ & 66.64$\pm0.29$ & 64.95$\pm0.30$ \\
FM                      & 72.72$\pm0.32$ & 72.83$\pm0.27$ & 73.14$\pm0.30$ & 72.53$\pm0.32$ & 71.82$\pm0.29$ & 72.11$\pm0.26$ & 72.00$\pm0.31$ & 71.97$\pm0.30$ & 72.16$\pm0.28$ & 71.35$\pm0.33$ & 71.00$\pm0.24$ & 69.76$\pm0.31$ & 69.94$\pm0.33$ & 69.84$\pm0.30$ & 68.33$\pm0.28$ & 66.67$\pm0.29$ & 65.03$\pm0.33$ \\ \midrule
\multicolumn{18}{c}{CIFAR10 - LP} \\ \midrule
\random                 & 88.9$\pm0.24$ & 88.43$\pm0.28$ & 88.46$\pm0.27$ & 88.36$\pm0.29$ & 88.27$\pm0.31$ & 88.17$\pm0.33$ & 88.64$\pm0.26$ & 88.88$\pm0.27$ & 88.01$\pm0.30$ & 87.38$\pm0.24$ & 86.77$\pm0.28$ & 86.91$\pm0.31$ & 86.38$\pm0.24$ & 85.58$\pm0.32$ & 85.18$\pm0.27$ & 83.99$\pm0.26$ & 82.85$\pm0.33$ \\
\moderate               & 89.17$\pm0.30$ & 88.83$\pm0.32$ & 88.74$\pm0.26$ & 88.92$\pm0.33$ & 88.82$\pm0.27$ & 88.32$\pm0.28$ & 88.09$\pm0.29$ & 88.41$\pm0.24$ & 87.63$\pm0.26$ & 86.99$\pm0.27$ & 86.85$\pm0.29$ & 87.31$\pm0.32$ & 86.36$\pm0.33$ & 86.20$\pm0.26$ & 84.75$\pm0.30$ & 84.35$\pm0.26$ & 82.14$\pm0.31$ \\
\gradnorm & 89.35$\pm0.29$ & 88.83$\pm0.24$ & 88.69$\pm0.27$ & 88.63$\pm0.28$ & 88.91$\pm0.30$ & 88.96$\pm0.26$ & 89.04$\pm0.33$ & 89.04$\pm0.29$ & 88.27$\pm0.32$ & 87.34$\pm0.27$ & 87.75$\pm0.24$ & 87.39$\pm0.31$ & 86.86$\pm0.27$ & 86.72$\pm0.33$ & 85.61$\pm0.26$ & 84.74$\pm0.31$ & 82.72$\pm0.32$ \\
LM                      & 89.51$\pm0.31$ & 89.97$\pm0.33$ & 89.42$\pm0.24$ & 89.79$\pm0.27$ & 89.42$\pm0.26$ & 89.31$\pm0.32$ & 89.14$\pm0.30$ & 89.15$\pm0.31$ & 88.81$\pm0.33$ & 88.74$\pm0.30$ & 88.81$\pm0.27$ & 88.54$\pm0.26$ & 88.33$\pm0.32$ & 87.43$\pm0.29$ & 87.03$\pm0.30$ & 86.62$\pm0.27$ & 85.87$\pm0.31$ \\
FM                      & 89.53$\pm0.28$ & 89.32$\pm0.30$ & 89.32$\pm0.31$ & 89.47$\pm0.29$ & 89.49$\pm0.32$ & 89.25$\pm0.31$ & 89.23$\pm0.30$ & 89.05$\pm0.27$ & 89.41$\pm0.33$ & 88.89$\pm0.26$ & 88.48$\pm0.29$ & 88.13$\pm0.30$ & 87.79$\pm0.27$ & 87.39$\pm0.33$ & 86.50$\pm0.30$ & 85.95$\pm0.32$ & 84.84$\pm0.27$ \\ \midrule
\multicolumn{18}{c}{Flowers102 - LP} \\ \midrule
\random                 & 94.14$\pm0.24$ & 94.25$\pm0.28$ & 94.32$\pm0.27$ & 94.23$\pm0.29$ & 94.03$\pm0.31$ & 93.99$\pm0.33$ & 93.96$\pm0.26$ & 94.20$\pm0.27$ & 93.91$\pm0.30$ & 93.30$\pm0.24$ & 93.26$\pm0.28$ & 93.59$\pm0.31$ & 93.26$\pm0.24$ & 92.77$\pm0.32$ & 91.51$\pm0.27$ & 91.56$\pm0.26$ & 88.96$\pm0.33$ \\
\moderate               & 94.24$\pm0.30$ & 94.19$\pm0.32$ & 94.6$\pm0.26$ & 94.19$\pm0.33$ & 94.19$\pm0.27$ & 94.38$\pm0.28$ & 94.23$\pm0.29$ & 94.28$\pm0.24$ & 93.3$\pm0.26$ & 93.71$\pm0.27$ & 93.67$\pm0.29$ & 93.42$\pm0.32$ & 92.73$\pm0.33$ & 92.61$\pm0.26$ & 92$\pm0.30$ & 91.56$\pm0.26$ & 88.63$\pm0.31$ \\
\gradnorm & 94.03$\pm0.29$ & 94.44$\pm0.24$ & 94.11$\pm0.27$ & 94.36$\pm0.28$ & 94.68$\pm0.30$ & 94.03$\pm0.26$ & 94.4$\pm0.33$ & 94.2$\pm0.29$ & 94.28$\pm0.32$ & 93.4$\pm0.27$ & 93.59$\pm0.24$ & 93.34$\pm0.31$ & 93.02$\pm0.27$ & 92.45$\pm0.33$ & 91.03$\pm0.26$ & 90.66$\pm0.31$ & 89.28$\pm0.32$ \\
LM                      & 94.11$\pm0.31$ & 94.60$\pm0.33$ & 94.31$\pm0.26$ & 94.32$\pm0.30$ & 94.68$\pm0.26$ & 94.23$\pm0.32$ & 94.29$\pm0.27$ & 94.32$\pm0.33$ & 94.32$\pm0.26$ & 94.64$\pm0.31$ & 94.67$\pm0.30$ & 94.15$\pm0.24$ & 93.99$\pm0.32$ & 93.54$\pm0.31$ & 93.5$\pm0.26$ & 91.96$\pm0.29$ & 90.91$\pm0.30$ \\
FM                      & 94.36$\pm0.26$ & 94.59$\pm0.33$ & 94.40$\pm0.24$ & 94.71$\pm0.32$ & 94.60$\pm0.31$ & 94.28$\pm0.33$ & 94.34$\pm0.26$ & 94.21$\pm0.31$ & 94.60$\pm0.30$ & 94.44$\pm0.33$ & 94.32$\pm0.29$ & 93.91$\pm0.26$ & 93.42$\pm0.31$ & 93.34$\pm0.24$ & 93.50$\pm0.30$ & 92.05$\pm0.27$ & 90.22$\pm0.26$ \\
\midrule
\bottomrule[1pt]
\end{tabular}
}
    \label{tab: app_number_main_lp}
\end{sidewaystable}

\paragraph{The main results reported in Fig.\,\ref{fig: main_lp} are stable: analysis with detailed numerical results and standard deviations.}
In \textbf{Tab.\,\ref{tab: app_number_main_lp}}, we provide the exact numerical results used to generate \textbf{Fig.\,\ref{fig: main_lp}}. These numbers give a more granular view of the performance comparisons. The results, calculated over three independent trials, show that the magnitude of the standard deviations is quite small relative to the mean values. This indicates that the trends and conclusions drawn from \textbf{Fig.\,\ref{fig: main_lp}} are generally valid and not significantly affected by trial variations. To maintain readability and clarity, we choose not to include these minor standard deviations in \textbf{Fig.\,\ref{fig: main_lp}}.

\begin{figure}[htb]
    \centering
    \begin{tabular}{cc}
    \includegraphics[width=.3\textwidth,height=!]{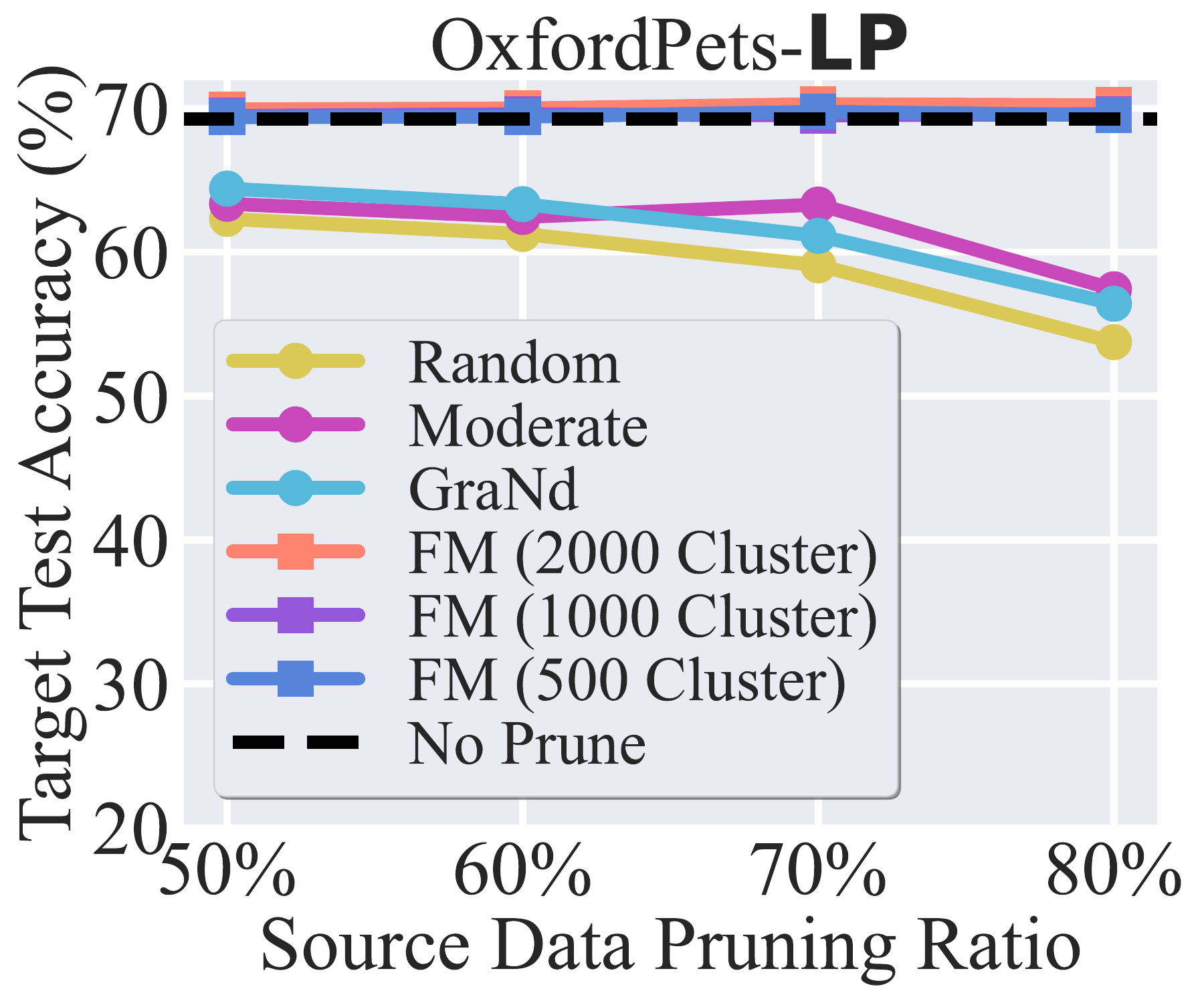} &
    \includegraphics[width=.3\textwidth,height=!]{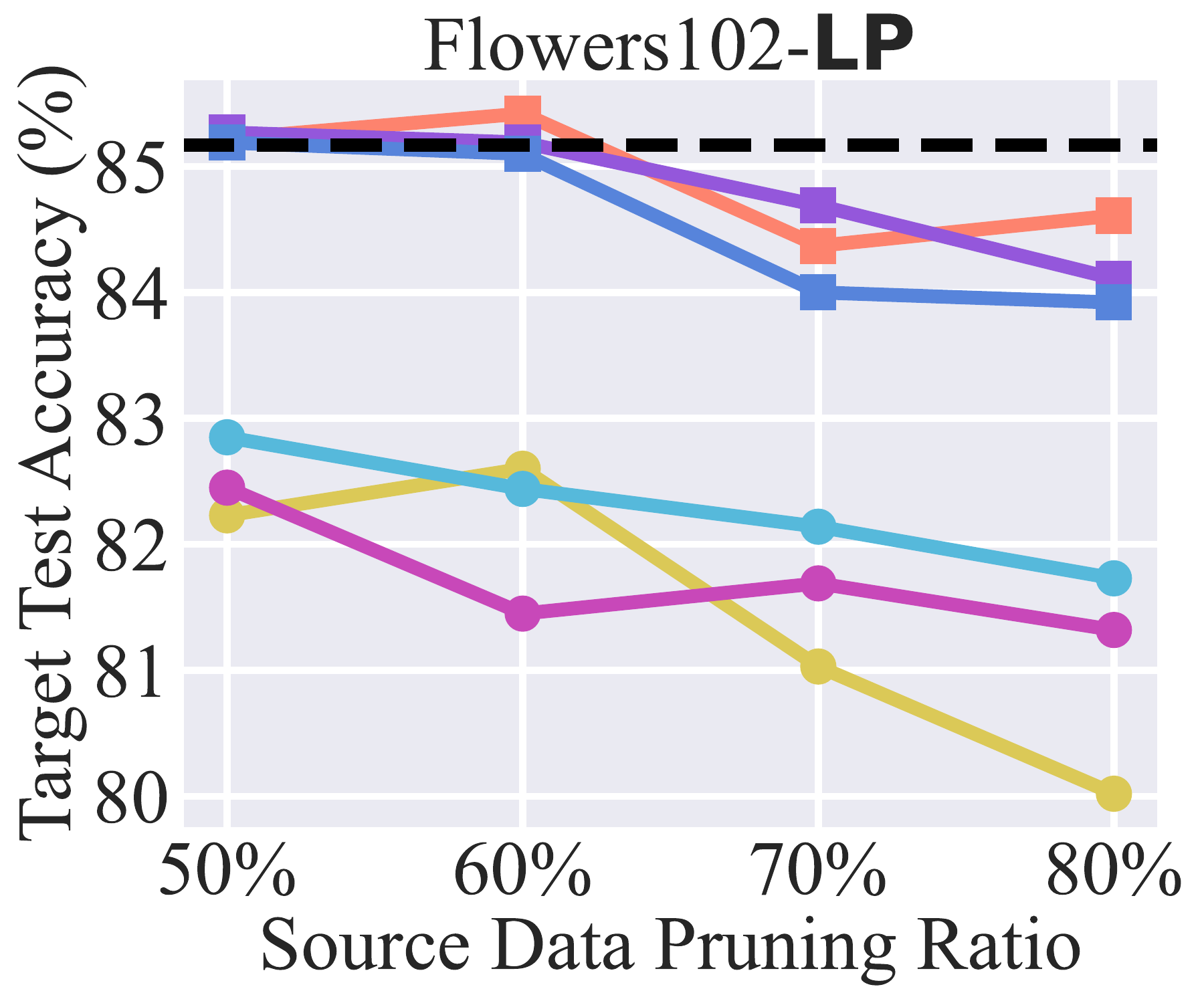} \\
    \end{tabular}
    \caption{Ablation study on the sensitivity of FM against the  choice of cluster numbers. For FM, the original dataset is pre-clustered into $K$ clusters, with $K=2000$ as the default setting in this study. Here, various $K$  values (500, 1000, and 2000) are studied in this experiment. Other settings are aligned with Tab.\,\ref{tab: main_ssl_lp}.}
    \label{fig: app_cluster_num_study}
    \vspace*{-1em}
\end{figure}

\paragraph{FM performance is robust to the choice of cluster number.} The results presented in \textbf{Fig.\,\ref{fig: app_cluster_num_study}} provide insights into the sensitivity of FM performance to variations of the cluster number ($K$). This extended study from Tab.\,\ref{tab: main_ssl_lp} shows that FM performance remains robust as the cluster number varies from  $500$ to $2000$, preserving the benefits of DP for transfer learning without much performance degradation. This implies that FM can provide reliable and stable results even when the hyperparameter settings are not optimal.

\begin{figure}[h]
    \centering
    \begin{tabular}{cc}
    \includegraphics[width=.3\textwidth,height=!]{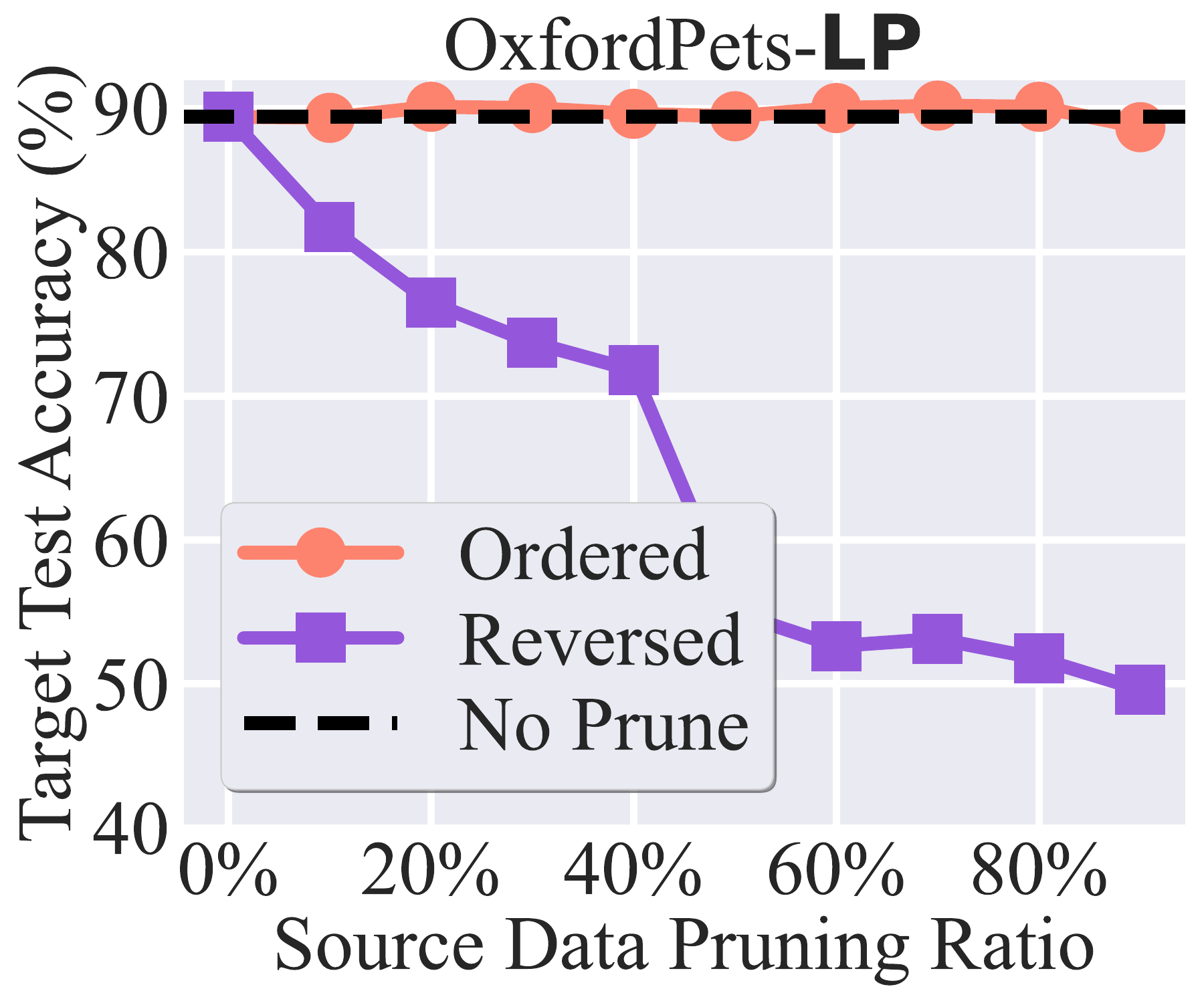} &
    \includegraphics[width=.3\textwidth,height=!]{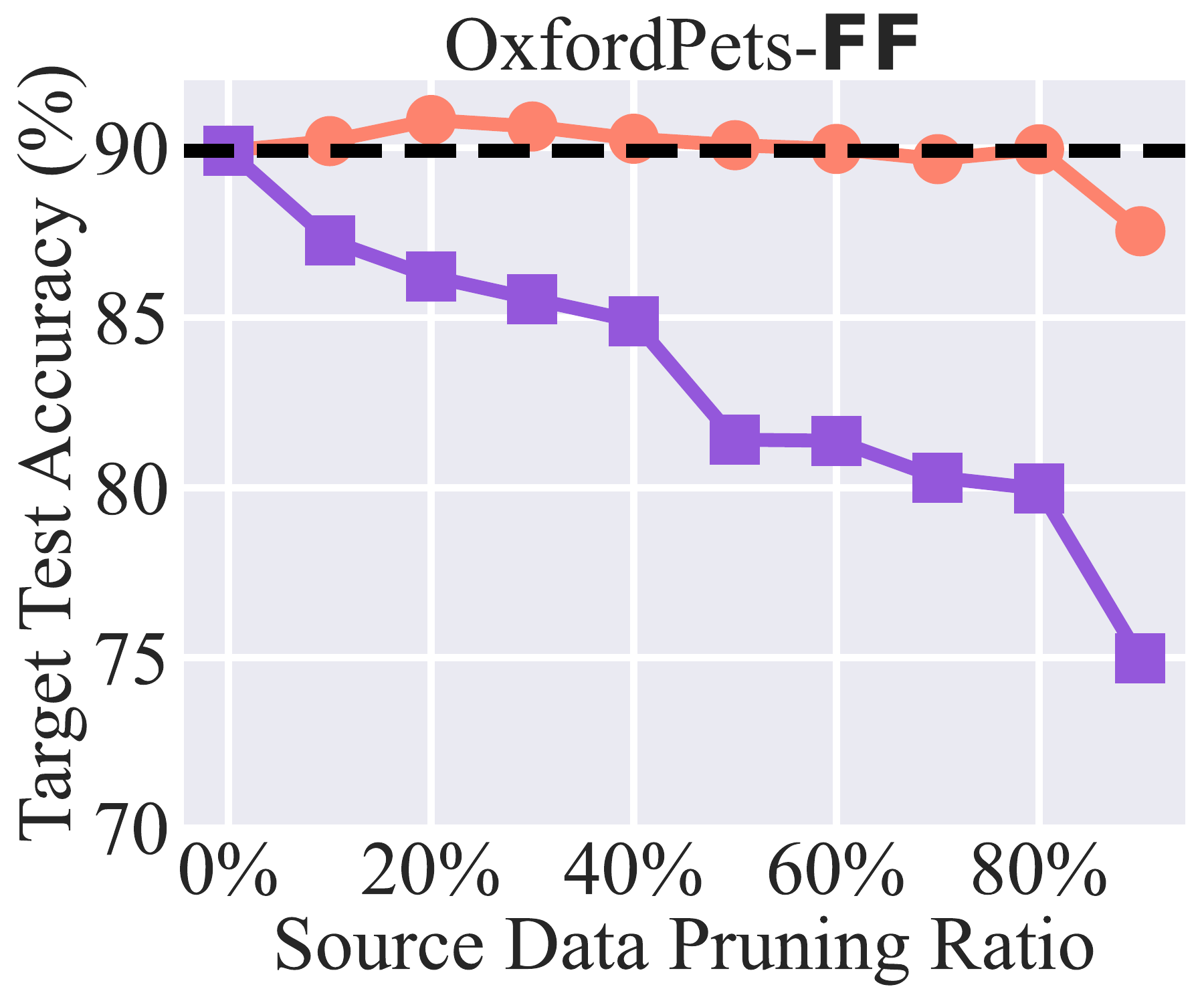} \\
    
    \end{tabular}
    \caption{Performance comparison on the downstream performance of LM with the ``Ordered'' and the ``Reversed'' pruning order. Here, the "Ordered" strategy retains source classes with the highest scores, while the "Reversed" order prunes these classes first. Other settings are aligned with Fig.\,\ref{fig: main_lp}.}
    \label{fig: app_reverse_order}
\end{figure}

\paragraph{LM identifies the most influential source classes.} To validate that the classes with the highest LM scores are the most influential, we present the pruning trajectory in \textbf{Fig.\,\ref{fig: app_reverse_order}}, where pruning is executed in the reverse order of the class scores, different from the proposed DP implementation. That is, classes with the smallest scores are retained, while the ones with the highest scores are pruned. Remarkably, even a slight pruning ratio (such as $10\%$) leads to a significant degradation in downstream performance. This evidence underscores the benefits of source classes with high LM scores in promoting the downstream performance.

\begin{figure}[htb]
    \centering
    \includegraphics[width=.4\linewidth]{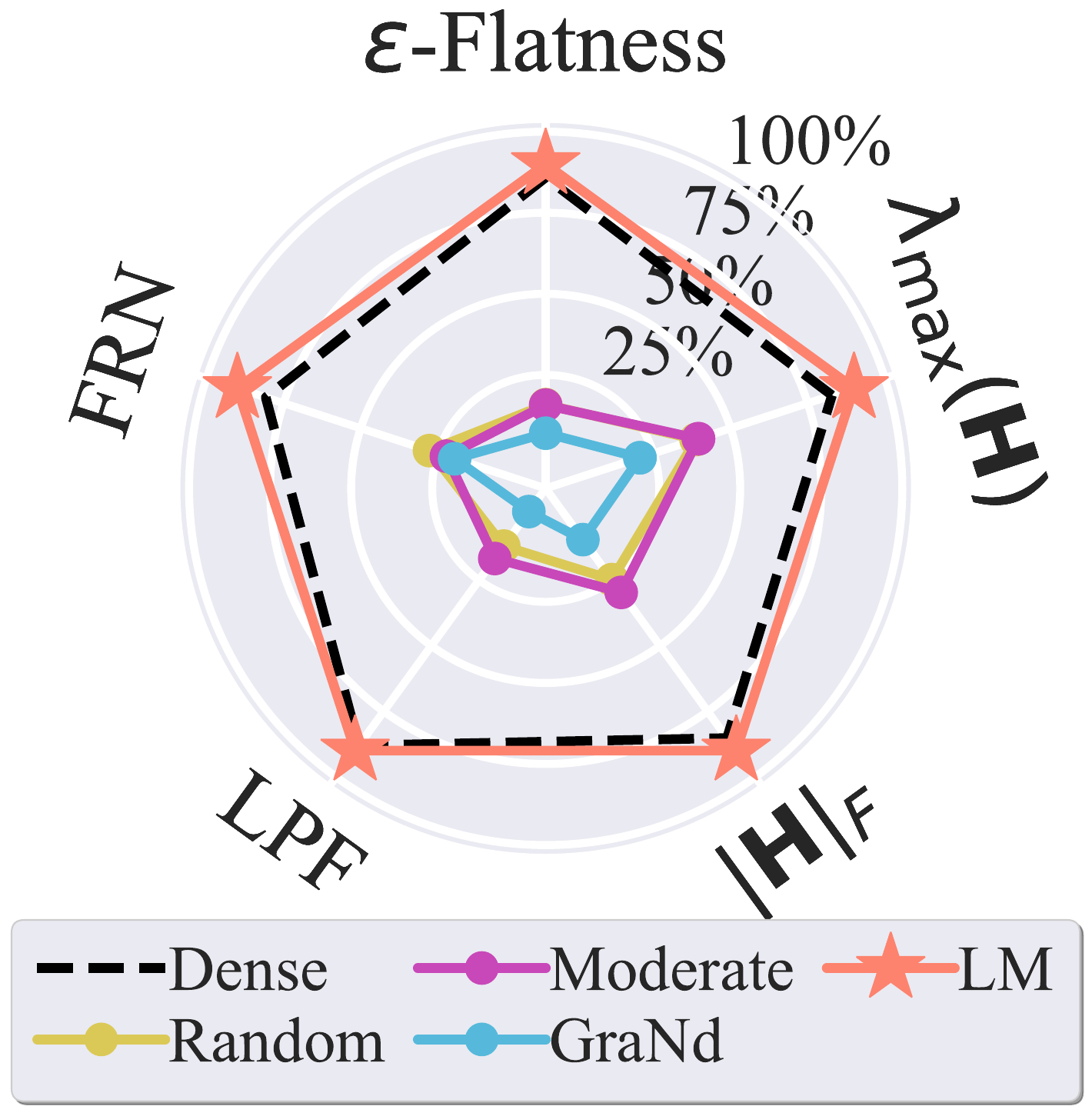}
    \caption{Flatness evaluations on the models pretrained with the pruned source data using different methods. The models are first pretrained and then finetuned using LP. The flatness are evaluated with respect to the downstream training loss and are quantified by the reversed sharpness evaluated through five widely acknowledged sharpness metrics. The results are normalized to $0\% \sim 100\%$, with $100\%$ denoting the highest (best) flatness across all the methods given one specific flatness metric.}
    \label{fig: app_sharpness}
    \vspace*{-1em}
\end{figure}

\begin{table}[t]
\caption{\textbf{Experiments with MoCov3 on ViT}. Other settings follow Tab.\,2 of the original submission. The performance surpassing the unpruned setting (pruning ratio $0\%$) is highlighted in \colorbox{LightCyan}{cyan}. The best result in each setting   is marked in \textbf{bold}. FM consistently outperforms other baselines and can find winning subsets with pruning ratios of more than 50\%.}
\centering
\resizebox{.5\linewidth}{!}{%
\begin{tabular}{c|ccccc}
\toprule[1pt]
\midrule
\multirow{2}{*}{\begin{tabular}[c]{@{}c@{}}Dataset\\ Pruning Ratio\end{tabular}} 
& \multicolumn{5}{c}{\texttt{OxfordPets}} \\ 
& 0\% & 50\%  & 60\%  & 70\%  & 80\% \\
\midrule
\random & \multirow{4}{*}{87.34} & 82.13 & 80.27 & 75.42 & 68.34 \\
\moderate & & 86.17 & 85.29 & 84.01 & 81.32 \\
\gradnorm & & 87.21 & 86.11 & 83.19 & 80.78 \\
FM (ours) &                        & \cellcolor{LightCyan}\textbf{87.68} & \cellcolor{LightCyan}\textbf{87.51} & \cellcolor{LightCyan}\textbf{87.39} & \textbf{84.14} \\
\midrule
\multirow{2}{*}{\begin{tabular}[c]{@{}c@{}}Dataset\\ Pruning Ratio\end{tabular}} 
& \multicolumn{5}{c}{\texttt{SUN397}} \\
& 0\% & 50\%  & 60\%  & 70\%  & 80\% \\
\midrule
\random & \multirow{4}{*}{60.36} & 59.22 & 58.45 & 56.73 & 54.29 \\
\moderate &  & 60.13 & 59.61 & 58.21 & 56.44\\
\gradnorm &  & \cellcolor{LightCyan}60.39 & 59.27 & 58.95 & 57.15 \\
FM (ours) &  & \cellcolor{LightCyan}\textbf{60.49} & \cellcolor{LightCyan}\textbf{60.55} & \cellcolor{LightCyan}\textbf{60.42} & \textbf{59.88} \\
\midrule
\multirow{2}{*}{\begin{tabular}[c]{@{}c@{}}Dataset\\ Pruning Ratio\end{tabular}} 
& \multicolumn{5}{c}{\texttt{Flowers102}} \\
& 0\% & 50\%  & 60\%  & 70\%  & 80\% \\
\midrule
\random & \multirow{4}{*}{93.96} & 92.41 & 91.65 & 90.17 & 88.41 \\
\moderate &  & 93.75 & 92.41 & 91.42 & 90.11 \\
\gradnorm &  & 93.88 & 93.21 & 91.77 & 90.45 \\
FM (ours) &  & \cellcolor{LightCyan}\textbf{94.11} & \cellcolor{LightCyan}\textbf{94.28} & \cellcolor{LightCyan}\textbf{93.97} & \textbf{91.42} \\
\midrule
\bottomrule[1pt]
\end{tabular}%
}
\label{tab: mocov3}
\end{table}

\paragraph{\revision{FM can be smoothly applied to MoCov3.}} We conducted experiments to illustrate the effectiveness of our proposed method (FM) when applied to the more recent SSL framework MoCov3\,\cite{fan2021multiscale} using the ViT structure. In line with our SSL experiment plan detailed in Tab.\,\ref{tab: main_ssl_lp} of the paper, \textbf{Tab.\,\ref{tab: mocov3}} tested FM on three downstream datasets, specifically OxfordPets, Flowers102, and SUN397, with a source data class pruning ratio ranging from $50\%$ to $80\%$. The results affirm that the principal conclusions drawn from MoCov2 remain consistent with MoCov3 on ViT. Our method, FM, successfully identifies data subsets within the source dataset that can be pruned at high ratios without compromising downstream performance (termed as “winning subsets”). For instance, in one particular case, the FM-based winning subsets achieve a pruning ratio of up to $70\%$ on OxfordPets.

\begin{figure}[t]
\centerline{
\begin{tabular}{cc}
    \hspace*{-2mm} \includegraphics[width=.3\textwidth,height=!]{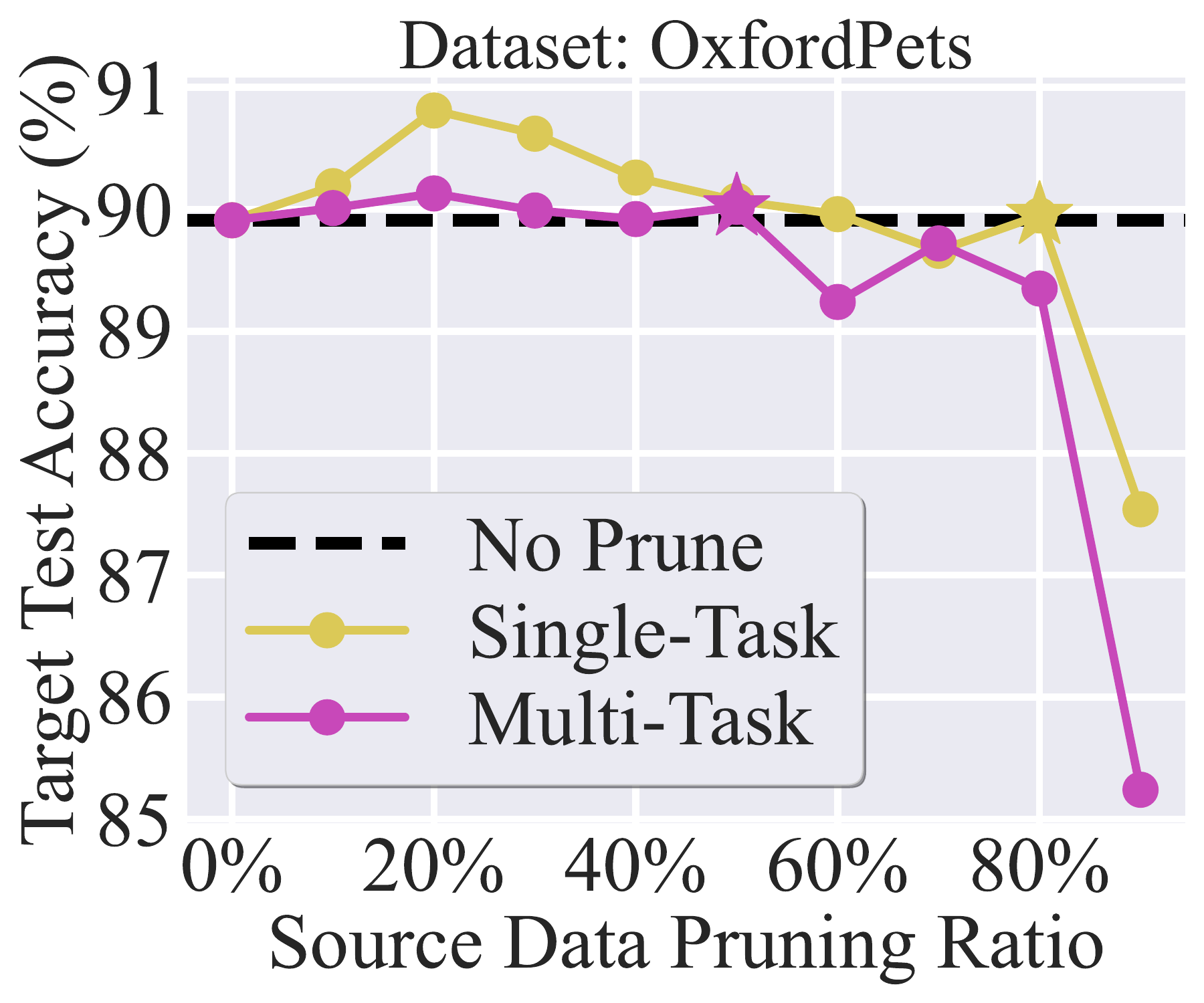} &
    \hspace*{-4mm}  \includegraphics[width=.3\textwidth,height=!]{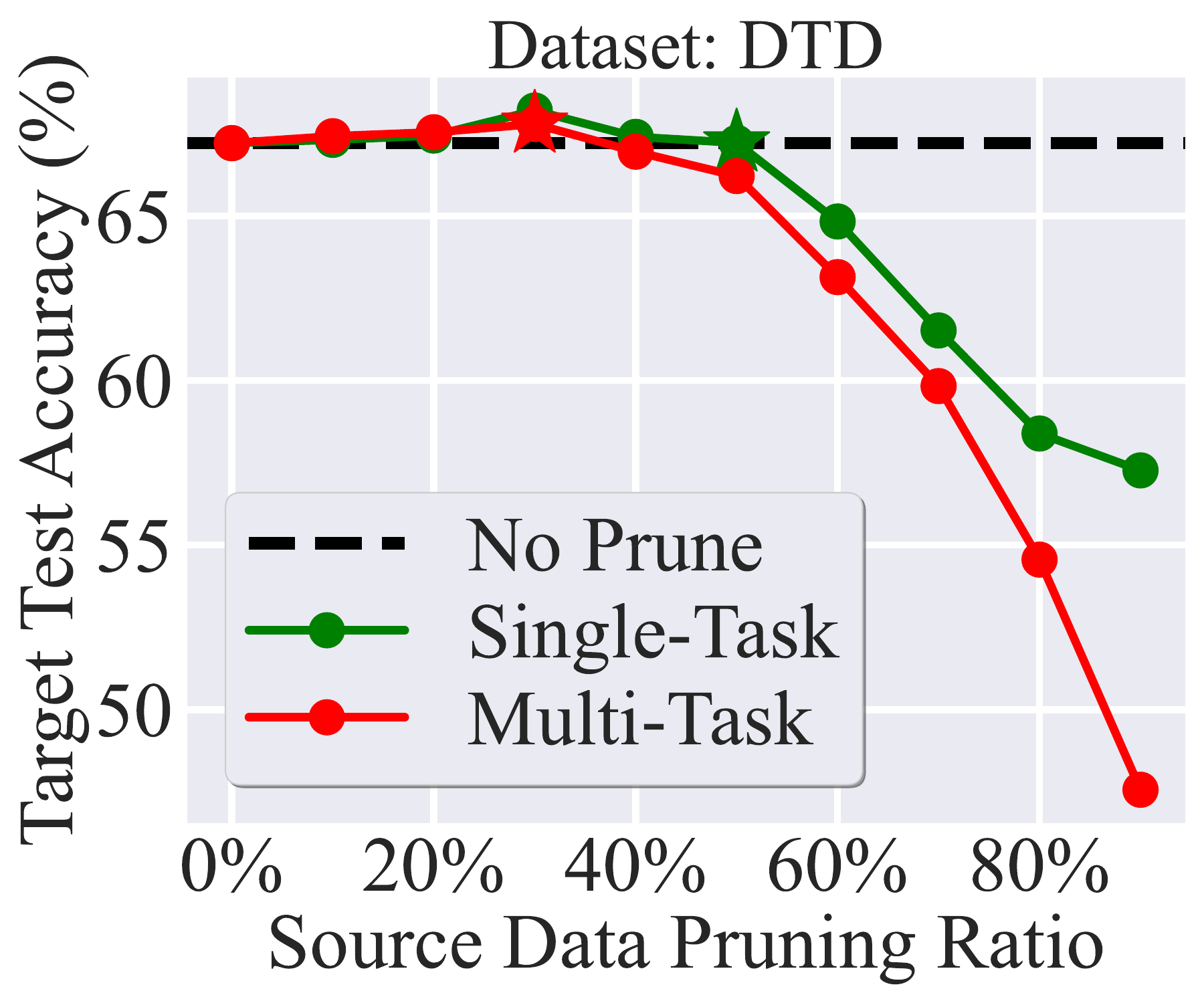} \\
    \hspace*{-2mm} \includegraphics[width=.3\textwidth,height=!]{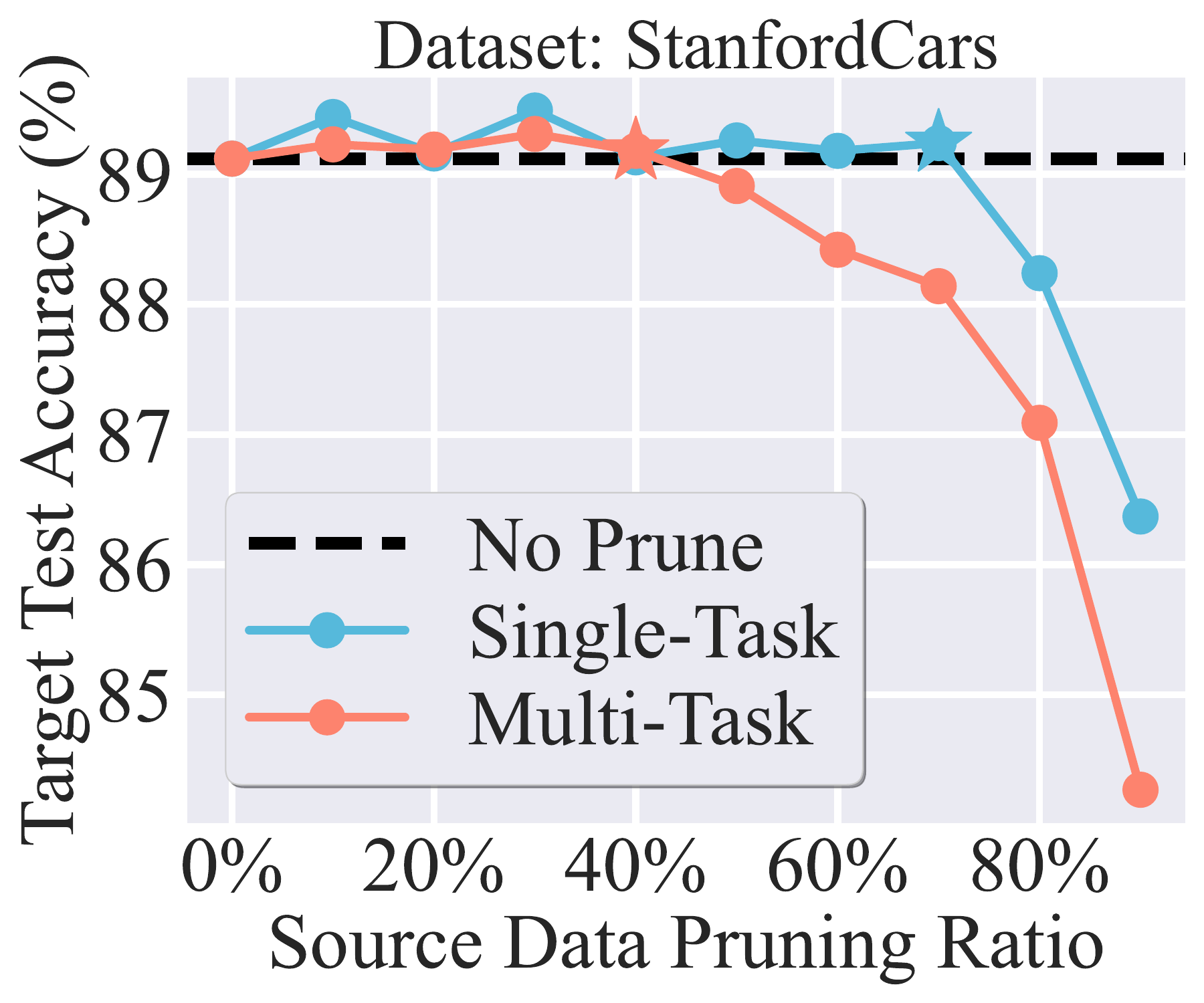} &
    \hspace*{-4mm}  \includegraphics[width=.3\textwidth,height=!]{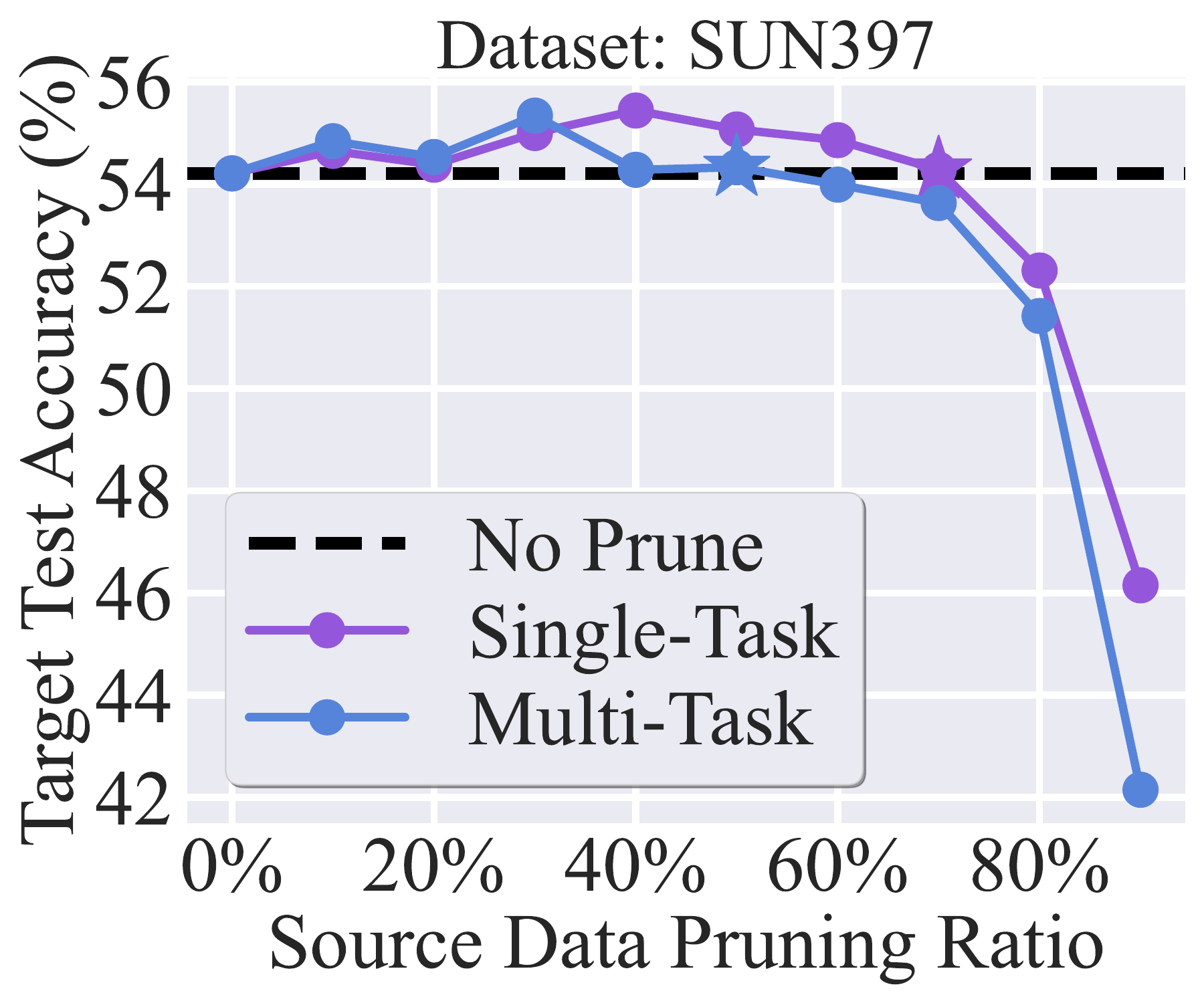} \\
\end{tabular}
}
\vspace*{-4mm}
\caption{\footnotesize \textbf{DP achieved by LM in the multi-task setting given 4 downstream tasks}. This expands Fig.\,\ref{fig: main_lp}, \textit{i.e.}, the single-task setting, where source data is pruned based on an individual task.
}
\vspace*{-2mm}
\label{fig: multi_task_study}
\end{figure}

\paragraph{\revision{Explore LM in the multi-task setting.}} A limitation of our method is its task-specific nature in data influence analysis. As our methods assess source data influence for specific downstream tasks, developing a universal DP solution for multiple tasks simultaneously is challenging.
As a preliminary study, we examine the performance of source dataset pruning for multiple downstream tasks simultaneously in \textbf{Fig.\,\ref{fig: multi_task_study}}. While LM can still identify winning subsets, the maximum pruning ratio diminishes as more tasks are considered. We will include this limitation in the Conclusion section.

\begin{table}[h]
\vspace*{-2mm}
\centering
\caption{\footnotesize Experiments on \texttt{CIFAR-10C}. LM-based source dataset pruning on ImageNet (given \texttt{CIFAR-10} as the downstream task) applies to transfer learning against \texttt{CIFAR-10C}. 5 out of the 19 corruption types are tested. 
}
\label{tab: cifar10c}
\resizebox{.5\columnwidth}{!}{%
\begin{tabular}{l|ccccc}
\toprule[1pt]
\midrule
\multicolumn{1}{c|}{\multirow{2}{*}{Dataset}} & \multicolumn{5}{c}{Pruning Ratio} \\
\multicolumn{1}{c|}{} & 0\% & 20\% & 40\% & 60\% & 80\% \\
\midrule
\multicolumn{1}{c|}{CIFAR10} & 96.83 & \cellcolor{LightCyan}96.88 & \cellcolor{LightCyan}\textbf{97.03} & 96.57 & 95.41 \\ \midrule
+ Gaussian Noise & 82.13 & \cellcolor{LightCyan}82.67 & \cellcolor{LightCyan}\textbf{82.89} & \cellcolor{LightCyan}82.60 & 81.19 \\
+ Defocus Blur & 84.73 & \cellcolor{LightCyan}85.22 & \cellcolor{LightCyan}\textbf{85.36} & \cellcolor{LightCyan}84.92 & 82.75 \\
+ Impulse Noise & 84.62 & \cellcolor{LightCyan}85.21 & \cellcolor{LightCyan}84.78 & \cellcolor{LightCyan}\textbf{85.93} & \cellcolor{LightCyan}85.11 \\
+ Shot Noise & 83.18 & \cellcolor{LightCyan}83.25 & \cellcolor{LightCyan}83.49 & \cellcolor{LightCyan}\textbf{83.76} & \cellcolor{LightCyan}83.24 \\
+ Speckle Noise & 83.11 & \cellcolor{LightCyan}\textbf{83.59} & \cellcolor{LightCyan}83.29 & \cellcolor{LightCyan}83.57 & 82.27 \\
\midrule
\bottomrule[1pt]
\end{tabular}%
}
\end{table}

\paragraph{\revision{LM helps remove data biases through DP for a downstream task.}} To investigate the scenarios with data biases, we conducted experiments on CIFAR-10C (the out-of-distribution scenario)\,\cite{hendrycks2019benchmarking}. We first pruned ImageNet given CIFAR-10 as a downstream task and evaluated the model on CIFAR-10C with different corruption types. \textbf{Tab.\,\ref{tab: cifar10c}} shows LM results with different pruning ratios for 5 strong perturbation types in CIFAR-10C. Impressively, LM can achieve winning subsets with pruning up to $80\%$, even better than on CIFAR-10, confirming our method’s effectiveness to filter biased data points to some degree. 

\begin{table}[h]
\vspace*{-3mm}
\caption{\footnotesize
Experiments on the few-shot transfer learning benchmark VTAB. Seven tasks in the \texttt{NATURAL} set are studied following the setting of Fig.\,\ref{fig: main_lp}. Each task contains 800 training and 200 testing samples.}
\centering
\fontsize{7pt}{7pt}\selectfont
\newcolumntype{C}{>{\centering\arraybackslash}X}
\setlength{\tabcolsep}{0pt}
\setlength{\extrarowheight}{5pt}
\renewcommand{\arraystretch}{0.75}
\begin{tabularx}{\linewidth}{p{10pt}p{1.35cm}!{\color{lightgray}\vline} CCCCCCC!{\color{lightgray}\vline}C}
\toprule
 &
 & \rotatebox{90}{\raisebox{0.5pt}{\tikz\fill[black] (0,0) circle (.5ex);} \texttt{Caltech101}}
 & \rotatebox{90}{\raisebox{0.5pt}{\tikz\fill[black] (0,0) circle (.5ex);} \texttt{CIFAR-100}}
 & \rotatebox{90}{\raisebox{0.5pt}{\tikz\fill[black] (0,0) circle (.5ex);} \texttt{DTD}}
 & \rotatebox{90}{\raisebox{0.5pt}{\tikz\fill[black] (0,0) circle (.5ex);} \texttt{Flowers102}}
 & \rotatebox{90}{\raisebox{0.5pt}{\tikz\fill[black] (0,0) circle (.5ex);} \texttt{OxfordPets}}
 & \rotatebox{90}{\raisebox{0.5pt}{\tikz\fill[black] (0,0) circle (.5ex);} \texttt{SUN397}}
 & \rotatebox{90}{\raisebox{0.5pt}{\tikz\fill[black] (0,0) circle (.5ex);} \texttt{SVHN}}
 & \rotatebox{90}{\raisebox{0.5pt}{\tikz\fill[natural] (0,0) circle (.5ex);} Mean}\\
\midrule

& No Prune & 80.23 & 46.39 & 62.48 & 90.39 & 88.42 & 33.32 & 87.24 & 69.78 \\

\arrayrulecolor{lightgray}\specialrule{.5pt}{0.6pt}{-0.5pt}\arrayrulecolor{black}
\multirow{4}{*}{\rotatebox{90}{\hspace*{-2pt}LM (Ours)}}
& Pruning 20\% 
& \cellcolor{LightCyan}\textbf{80.39} 
& \cellcolor{LightCyan}\textbf{46.65} 
& \cellcolor{LightCyan}62.88 
& \cellcolor{LightCyan}90.78 
& \cellcolor{LightCyan}88.57 
& \cellcolor{LightCyan}\textbf{34.22} 
& \cellcolor{LightCyan}\textbf{87.45} 
& \cellcolor{LightCyan}\textbf{70.13}\\
& Pruning 40\% 
& \cellcolor{LightCyan}80.34 
& \cellcolor{LightCyan}46.49 
& \cellcolor{LightCyan}\textbf{62.92} 
& \cellcolor{LightCyan}\textbf{90.88} 
& \cellcolor{LightCyan}\textbf{88.62} 
& \cellcolor{LightCyan}33.81 
& 87.21 
& \cellcolor{LightCyan}70.04 \\
& Pruning 60\% & 78.65 & 45.51 & \cellcolor{LightCyan}62.57 & 90.12 & 87.35 & \cellcolor{LightCyan}33.57 & 86.38 & 69.16 \\
& Pruning 80\% & 73.31 & 41.33 & 61.18 & 89.42 & 85.11 & 31.49 & 84.22 & 66.58 \\
\bottomrule
\end{tabularx}
\vspace*{-2mm}
\label{tab: vtab}
\end{table}

\paragraph{\revision{Experiments on the few-shot benchmark VTAB\,\cite{zhai2019large}.}} We extend our experiments to include the VTAB benchmark\,\cite{zhai2019large}. As shown in Tab.\,\ref{tab: vtab}, even in the few-shot setting, our major conclusions remain valid. LM can effectively pinpoint the most influential source subsets and eliminate the less valuable ones. For most tasks considered, winning subsets with pruning ratios up to $40\%$ are successfully found.

\begin{table}[h]
\centering
\vspace*{-5mm}
\caption{\footnotesize%
{Performance on surrogate model size.
Experiments follow the setting of Fig.\,\ref{fig: main_lp}:
RN-101 is first pretrained on the pruned source dataset ({ImageNet}) based on the surrogate model, and then finetuned on the downstream task {OxfordPets}. 
Under different surrogate models, the source class selection overlapping ratio with the used surrogate model RN-18 in the submission is reported under $50\%$ pruning ratio. 
}}
\label{tab: surrogate_model}
\resizebox{.85\columnwidth}{!}{%
\begin{tabular}{c|ccccccc|c}
\toprule[1pt]
\midrule
\begin{tabular}[c]{@{}c@{}} Surrogate Model \\ Architecture\end{tabular} & RN-20s & VGG-2 & RN-32s & VGG-4 & RN-44s & RN-56s & VGG-8 & \cellcolor{Gray}\begin{tabular}[c]{@{}c@{}} RN-18 \\ (Default)\end{tabular} \\
\midrule
Param. $\#$ (M) & 0.236 & 0.417 & 0.516 & 0.698 & 0.706 & 0.896 & 5.53 & \cellcolor{Gray}11.69 \\
\midrule
Source Acc. (\%) & 36.25 & 22.56 & 40.77 & 29.44 & 43.74 & 45.72 & 58.45 & \cellcolor{Gray}68.73 \\
\midrule
\begin{tabular}[c]{@{}c@{}}Largest Pruning Ratio\\ of Winning Subsets (\%)\end{tabular} & 60 & 50 & 80 & 70 & 80 & 80 & 80 & \cellcolor{Gray}80 \\
\midrule
\begin{tabular}[c]{@{}c@{}}Source Class \\ Selection Overlap (\%)\end{tabular} & 89.3 & 84.4 & 90.7 & 87.2 & 93.5 & 94.8 & 97.7 & \cellcolor{Gray}100 \\
\midrule
\bottomrule[1pt]
\end{tabular}%
}
\vspace*{-2mm}
\end{table}

\paragraph{Subsets obtained by LM improve the   flatness of loss landscape on downstream tasks.} \textbf{Fig.\,\ref{fig: app_sharpness}} evaluates the flatness of the loss landscape of the model pretrained on an $80\%$-pruned source dataset and subsequently finetuned on the downstream OxfordPets dataset \cite{parkhi2012cats}. Better flatness in the loss landscape is typically associated with better transferability. We quantify this flatness using the measure of \textit{reversed sharpness} calculated via five widely accepted metrics: $\epsilon$-sharpness \cite{keskar2016large}, low pass filter-based measure (LPF) \cite{bisla2022low}, the max eigenvalue of the Hessian ($\lambda_{\text{max}}(\mathbf{H})$) \cite{maddox2020rethinking}, the Frobenius norm of the Hessian ($|\textbf{H}|_F$) \cite{maddox2020rethinking}, and Fisher Rao Norm (FRN) \cite{liang2019fisher}. The results suggest that pretraining on the subset selected by LM leads the model towards a flatter region in the downstream loss landscape, potentially contributing to the superior transferability of LM when compared to the baseline methods.

\begin{figure}[htb]
\centerline{
\includegraphics[width=.8\linewidth]{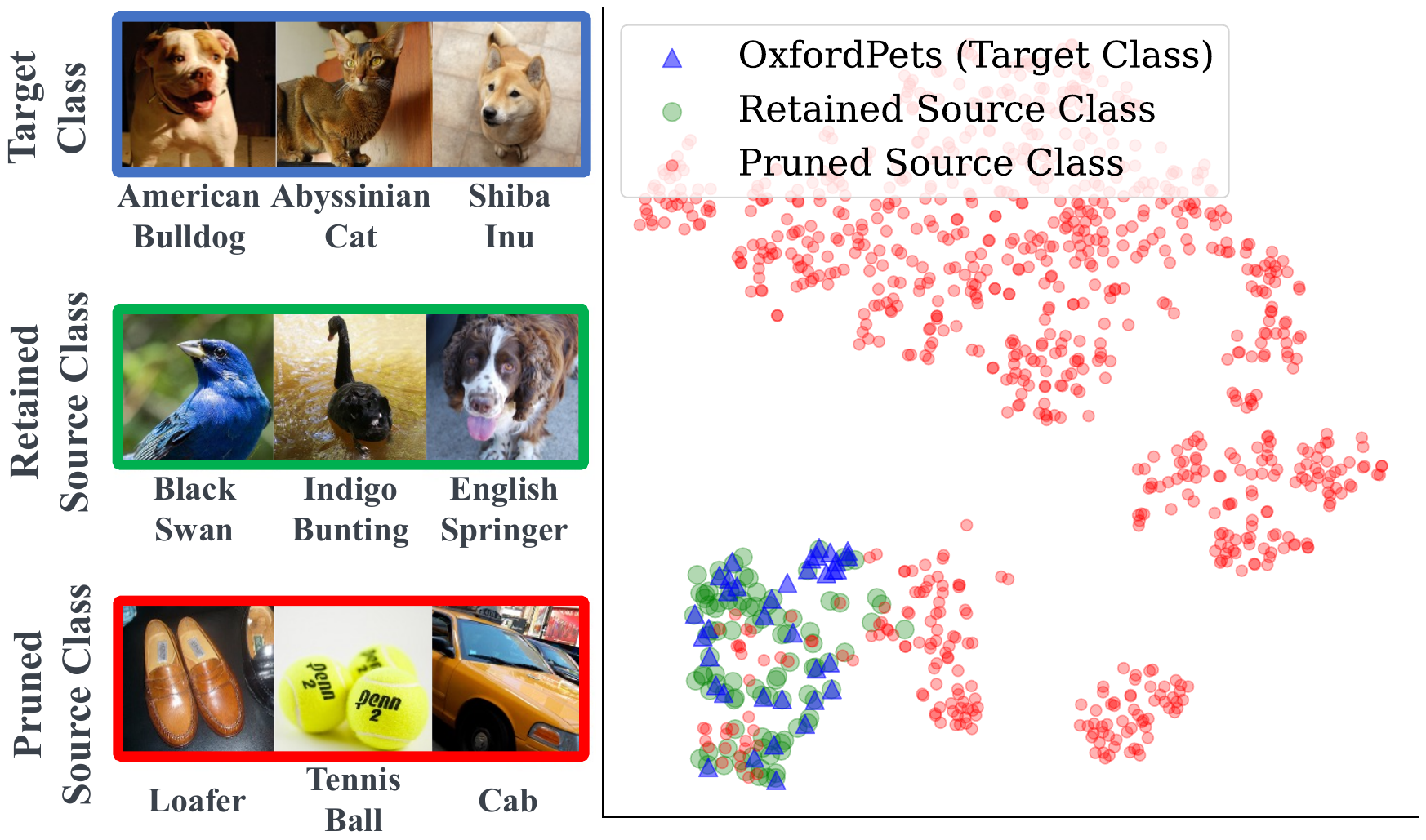}
}
\caption{\footnotesize{(Left) Interpretation merit of the data pruning strategy by LM. (Right) Feature distribution visualization using t-SNE for the source class selection by LM for OxfordPets with $90\%$ pruning ratio.
}}
\label{fig: app_feature_dist}
\end{figure}
\begin{figure}
  \centering
  \begin{subfigure}[b]{0.9\textwidth}
    \includegraphics[width=\linewidth]{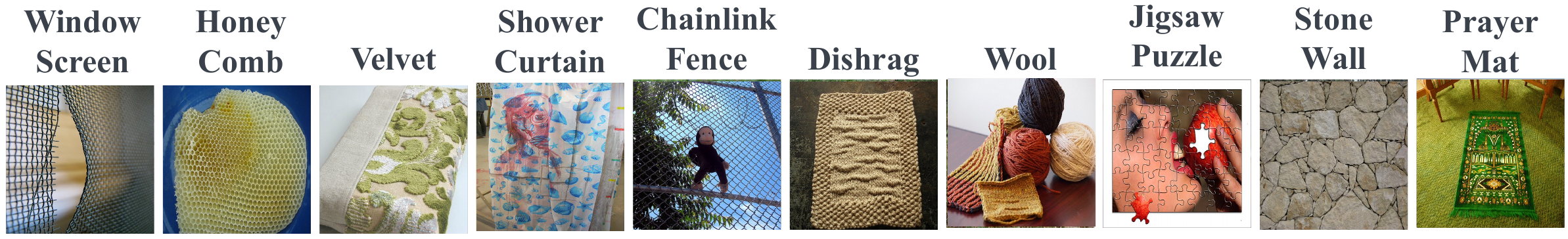}
    \caption{DTD.}
    \vspace{3.2mm}
    \label{fig: ebe_dtd}
  \end{subfigure}
  
  \begin{subfigure}[b]{0.9\textwidth}
    \includegraphics[width=\linewidth]{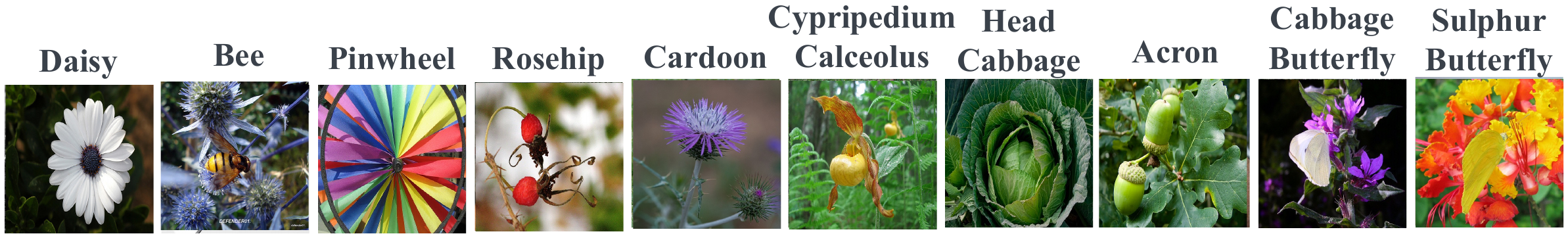}
    \caption{Flowers102.}
    \vspace{3.2mm}
    \label{fig: ebe_flowers}
  \end{subfigure}
  
  \begin{subfigure}[b]{0.9\textwidth}
    \includegraphics[width=\linewidth]{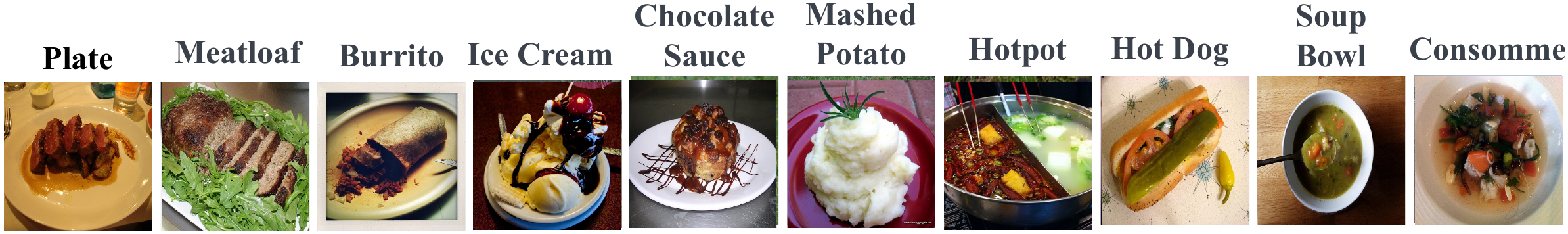}
    \caption{Food101.}
    \vspace{3.2mm}
    \label{fig: ebe_food}
  \end{subfigure}

    \begin{subfigure}[b]{0.9\textwidth}
    \includegraphics[width=\linewidth]{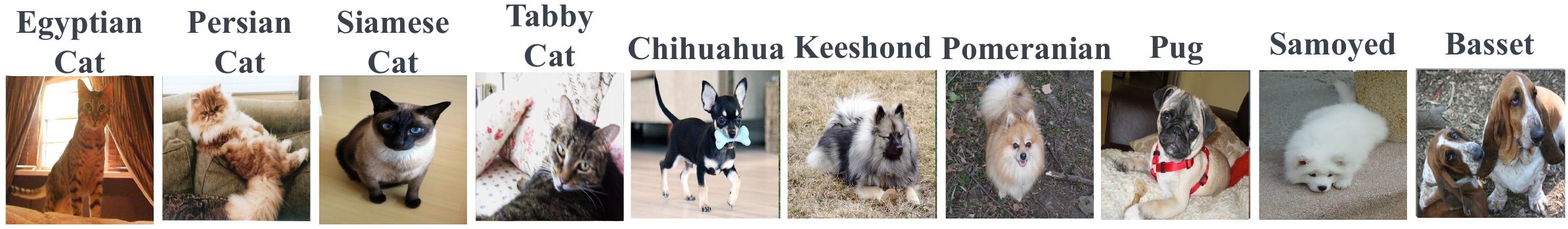}
    \caption{OxfordPets.}
    \vspace{3.2mm}
    \label{fig: ebe_pets}
  \end{subfigure}

    \begin{subfigure}[b]{0.9\textwidth}
    \includegraphics[width=\linewidth]{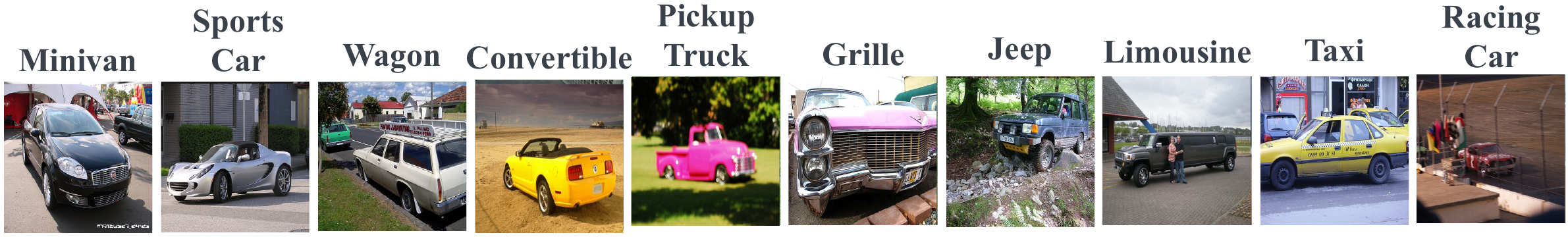}
    \caption{Stanfordcars.}
    \vspace{3.2mm}
    \label{fig: ebe_cars}
  \end{subfigure}

    \begin{subfigure}[b]{0.9\textwidth}
    \includegraphics[width=\linewidth]{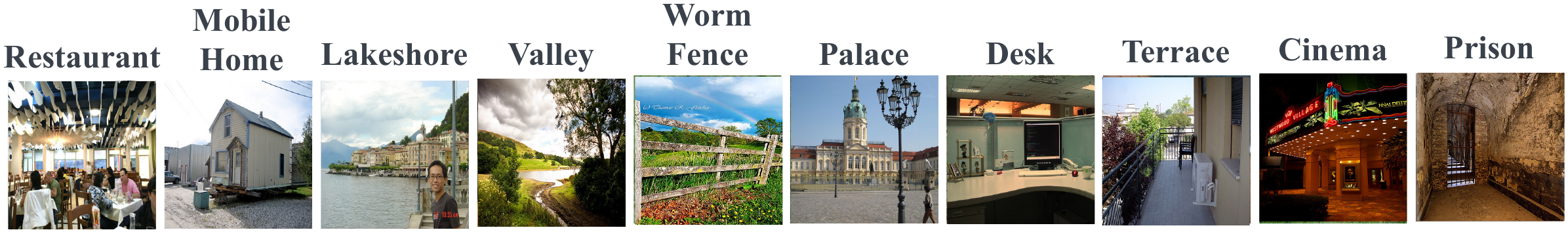}
    \caption{Sun397.}
    \vspace{3.2mm}
    \label{fig: ebe_sun}
  \end{subfigure}

    \begin{subfigure}[b]{0.9\textwidth}
    \includegraphics[width=\linewidth]{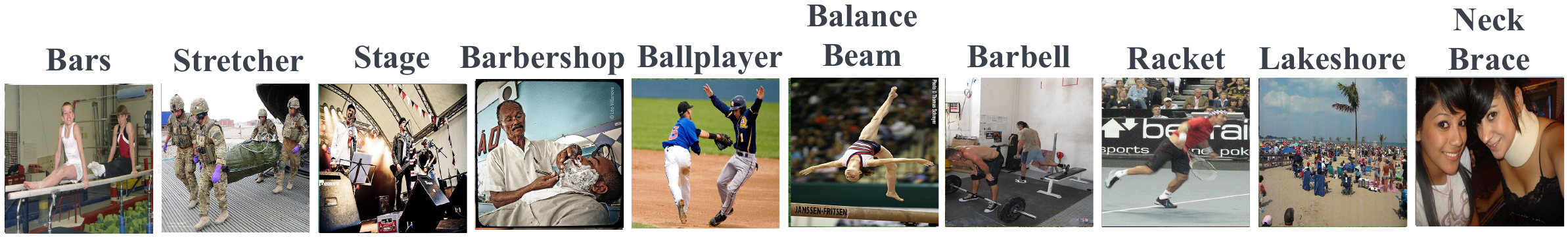}
    \caption{UCF101.}
    \vspace{3.2mm}
    \label{fig: ebe_ucf}
  \end{subfigure}

    \begin{subfigure}[b]{0.9\textwidth}
    \includegraphics[width=\linewidth]{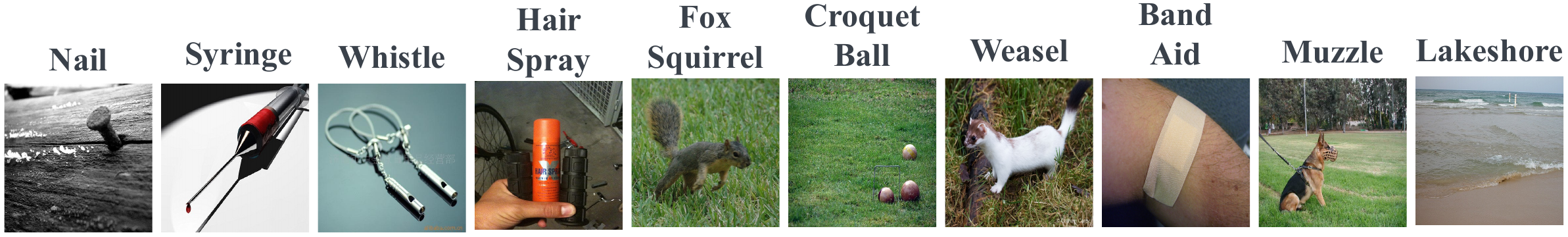}
    \caption{CIFAR-10.}
    \label{fig: ebe_cifar10}
  \end{subfigure}
  
  \caption{The source classes with top-10 scores selected by FM for the 8 downstream tasks studied in this work. For each class, the class label (name) as well as an representative image example is presented. }
  \label{fig: app_ebe_all}
\end{figure}

\paragraph{Feature distribution analysis.}
{\textbf{Fig.\,\ref{fig: app_feature_dist}} provides visual explanations of DP at both deep representation (feature) and data levels. Here we focus on the LM method with  a pruning ratio of $90\%$ given the downstream dataset OxfordPets. The other settings are consistent with Fig.\,\ref{fig: main_lp}. In \textbf{Fig.\,\ref{fig: app_feature_dist} (right)}, we  visualize the source ImageNet classes (including \textcolor{RedOrange}{pruned} ones and \textcolor{LimeGreen}{retrained} ones) and \textcolor{blue}{target} OxfordPets classes in terms of their class-wise feature centroids  in a 2D space achieved by  {t-SNE}\cite{van2008visualizing}. The class feature centroid is obtained by averaging the features of data points within a class, extracted by the pretrained source model  on the full ImageNet dataset. As we can see, all retained source classes are grouped together and are close to the target classes. This indicates that the source classes that the pruning process share the most resemblance with the target data. In contrast, the pruned source classes are more dispersed and located further away from the target data classes. Furthermore, \textbf{Fig. \ref{fig: app_feature_dist} (left)} exhibits image examples of  \textcolor{blue}{target}   classes as well as \textcolor{RedOrange}{pruned}  and \textcolor{LimeGreen}{retrained} source classes. We observe that image examples in  the retained source classes (\textit{e.g.}, relating to animals) are semantically closer to the target data  points (relating to pets) than the pruned ones. This highlights the ability of LM to effectively identify and retain the most relevant classes for the downstream tasks. We provide more examples for FM in Fig.\,\ref{fig: app_ebe_all}.
}

\paragraph{Examining the top selected classes by FM and their image examples.} In \textbf{Fig.\,\ref{fig: app_ebe_all}}, we showcase the top-10 source classes chosen by FM, as determined by the endowed scores. These selected classes closely correspond to the downstream datasets' subjects, demonstrating FM's effectiveness in identifying relevant classes for transfer learning. This finding also aligns with our observations   in Fig.\,\ref{fig: app_feature_dist}, showing that FM identifies  source classes resembling downstream data.

\end{document}